\documentclass{article}

\usepackage{arxiv}

\usepackage[utf8]{inputenc} 
\usepackage[T1]{fontenc}    
\usepackage{hyperref}       
\usepackage{url}            
\usepackage{booktabs}       
\usepackage{amsfonts}       
\usepackage{nicefrac}       
\usepackage{microtype}      
\usepackage{lipsum}		
\usepackage{graphicx}
\usepackage{natbib}
\usepackage{doi}
\usepackage{algorithm}
\usepackage[noend]{algpseudocode}
\usepackage{caption}
\usepackage{subcaption}

\usepackage{amsfonts,amsmath}       
\usepackage{subcaption}
\usepackage{epstopdf}
\usepackage{amsthm}
\usepackage{multicol}
\usepackage{wrapfig}
\usepackage{bbm}

\title{Information-Gathering in Latent Bandits}

\author{\href{https://orcid.org/0000-0002-7453-9186}{\includegraphics[scale=0.06]{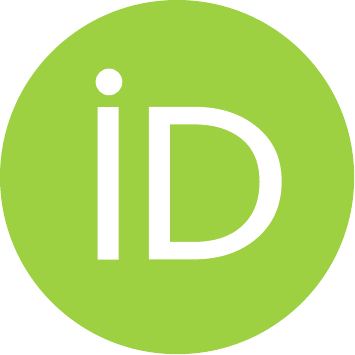}\hspace{1mm}Alexander Galozy}\thanks{corresponding author.} \\
	Center for Applied Intelligent Systems Research\\
	Halmstad University\\
	Kristian IV:s väg 3, 301 18 Halmstad \\
    \texttt{alexander.galozy@hh.se}\\
	\And
     \href{https://orcid.org/0000-0002-7796-5201}{\includegraphics[scale=0.06]{orcid.pdf}\hspace{1mm}S\l{}awomir Nowaczyk} \\
    Center for Applied Intelligent Systems Research\\
    Halmstad University\\
    Kristian IV:s väg 3, 301 18 Halmstad \\
    \texttt{slawomir.nowaczyk@hh.se} \\
}

\renewcommand{\shorttitle}{\textit{arXiv} Template}
\DeclareMathOperator{\EX}{\mathbb{E}}
\DeclareMathOperator{\diag}{diag}

\hypersetup{
pdftitle={A template for the arxiv style},
pdfsubject={q-bio.NC, q-bio.QM},
pdfauthor={David S.~Hippocampus, Elias D.~Striatum},
pdfkeywords={First keyword, Second keyword, More},
}

\newcommand{\argmax}{\mathop{\mathrm{argmax}}}
\newcommand{\argmin}{\mathop{\mathrm{argmin}}}
\begin{document}
\maketitle

\begin{abstract}
In the latent bandit problem, the learner has access to reward distributions and -- for the non-stationary variant -- transition models of the environment. The reward distributions are conditioned on the arm \textit{and} unknown latent states. The goal is to use the reward history to identify the latent state, allowing for the optimal choice of arms in the future. The latent bandit setting lends itself to many practical applications, such as recommender and decision support systems, where rich data allows the offline estimation of environment models with online learning remaining a critical component. Previous solutions in this setting always choose the \textit{highest reward arm} according to the agent's beliefs about the state, not explicitly considering the value of \textit{information-gathering arms}. Such information-gathering arms do not necessarily provide the highest reward, thus may never be chosen by an agent that chooses the highest reward arms at all times.
    
In this paper, we present a method for information-gathering in latent bandits. Given particular reward structures and transition matrices, we show that choosing the best arm given the agent's beliefs about the states incurs higher regret. Furthermore, we show that by choosing arms carefully, we obtain an improved estimation of the state distribution, and thus lower the cumulative regret through better arm choices in the future. We evaluate our method on both synthetic and real-world data sets, showing significant improvement in regret over state-of-the-art methods. 
\end{abstract}

\keywords{Latent Bandits \and Information Gathering \and Non-stationary \and Information Directed Sampling} 

\section{Introduction}

The multi-armed bandit problem (MAB) provides a principled way of modeling the exploitation-exploration dilemma found in sequential decision-making processes. Examples of the importance of the MAB problem can be found in applications such as clinical trials~\citep{Villar2015, Bastani2020}, finance~\citep{Shen2015, Huo2017}, routing networks~\cite{Boldrini2018, Kerkouche2018}, online advertising~\citep{Wen2017, Schwartz2017} and movie~\citep{Wang2019} or app recommendation~\citep{Baltrunas2015}.

Many of the applications mentioned above can be framed under the so-called \textit{latent bandit} problem, where a hidden underlying state governs the rewards. This setting assumes that the decision-maker has access to reward models, a natural assumption in any domain where large amounts of offline data are available~\citep{Hong2020}. An extension to the latent bandit setting considers the possibility of changing states studied under the \textit{non-stationary latent bandits} introduced by ~\cite{Hong2020}. It is quite common in practice that the latent state is subject to change, abrupt or gradual, as time progresses, e.g., a user might be interested in a particular movie genre for a shorter or longer period. In the non-stationary latent bandit setting, the agent has access to the state transition model in addition to the reward models.

Besides the rewards observed for particular arms, we are often presented with additional side observations, called the \textit{context}. These contexts allow the agent to uncover the latent state to some extent. The inherent challenge in real-world applications is that these observations are imperfect and only allow the agent to construct, over time, a probabilistic belief over the current state of the environment. This partial observability of the state often necessitates occasional \textit{information-gathering}, which may come with additional cost. Therefore, in the interest of maximizing payoff, the agent must carefully consider whether such information-gathering arms are justified in any given context. It is often a reasonable assumption that arms exist that provide more information about states than others.

In previous work, arms are chosen that maximize reward over the current state belief, choosing among the arms that have the highest reward in each state~\citep{Hong2020, Maillard14}. They do not explicitly consider how effectively these arms reduce state uncertainty. In the movie recommendation example, the agent may believe a user is in two states with similar probability, given the rewards received. The two states share similar reward distributions for the movie genre with high rewards, which the agent always chooses. Recommending movies from this shared genre is ineffective in uncovering the hidden state. To reduce state uncertainty, it may become imperative to consider movie recommendations other than ones with the highest expected reward. Moreover, as we will show, depending on the transition structure of the environment, knowing the current state more accurately can significantly impact future belief-states and thus arm selection and cumulative reward. 

We propose a more deliberate selection of arms that uncovers the latent state of the users effectively, allowing the agent to explicitly reason about incurring more regret now but avoid choosing sub-optimal arms in the future. For our approach, we follow the idea of \textit{information-gathering actions} in POMPDs research domain to allow the agent to balance its knowledge about the state space (where it is necessary to make better decisions) and maximize cumulative reward. Akin to \textit{Information-directed sampling}~\citep{Russo2014}, we exploit the knowledge of transition and reward models to estimate the usefulness of an arm for reducing state uncertainty.

The non-stationary latent bandit is a special case of the observable Markov decision process (POMPD) in full reinforcement learning~\citep{Hong2020}. In POMPDs, the agent's arms affect (usually) the dynamics of the environment, which is not the case in latent bandits. Nevertheless, the same difficulty in maximizing cumulative regret is apparent: significant state uncertainty may lead to more mistakes, increasing cumulative regret. Thus, policies that aim to maximize immediate reward, but neglect the effect arm selection on state uncertainty, continue selecting sub-optimal arms long into the future. 

We summarize our contributions as follows. We develop an algorithm based on posterior sampling that chooses arms via an information-directed criterion and trajectory roll-outs. This allows the agent to choose information-gathering arms deliberately to uncover the latent state more effectively, increasing cumulative reward. We evaluate our approach under controlled conditions using synthetic environments with different reward and transition model structures. These experiments provide a tool to analyze the effectiveness of our approach in different settings, where the use of information-gathering arms significantly improves cumulative regret (over contemporary methods). We test our algorithm on a large-scale real-world recommendation data set, showing the benefit of using information-gathering in improving cumulative regret. Additionally, we conduct experiments on synthetic and real-world data to determine when we benefit using our approach, particularly when the latent state cannot be effectively uncovered without information-gathering arms.

We structure the paper as follows. In section \ref{sec:motivation}, we illustrate on several simple examples the benefit of information-gathering arms in the latent bandit setting. In section \ref{sec:problem}, we formalize our setting in the non-stationary latent bandit problem. In section \ref{sec:algo_complete}, we present an algorithm based on posterior sampling that exploits the information-gathering arm for future reward. In section \ref{sec:exp}, we report the results of experiments using synthetic and real-world data. We review relevant related work in section \ref{sec:related} and conclude this paper in section \ref{sec:conclusion}. 

\section{Motivational example} \label{sec:motivation}

We start with defining a simple example to intuitively illustrate the benefits of information-gathering arms for faster state identification. Imagine a setting where two states determine the expected reward for a set of movies divided into three movie genres, where a movie from one of the genres needs to be recommended. Movies from the two genres have high but similar mean rewards in both states. Movies from the third genre have a high mean reward in one state but a much lower mean reward in the other. The agent assumes that this environment follows the stationary latent bandit setting. Since the reward model is available, the agent would know which movie is best to recommend if only the state was known. The difficulty lies in figuring out, from the received rewards, the latent state of the environment. For our example going forward the true latent state is $s_1$. 

\begin{figure}
 \begin{subfigure}[b]{0.45\textwidth}
     \centering
     \includegraphics[width=\textwidth]{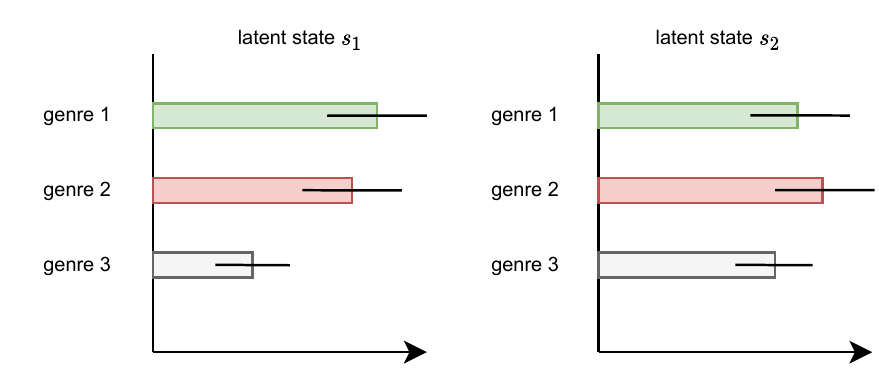}
     \caption{Rewards for three movies genres. Genre $1$ and genre $2$ are the highest reward arms for both latent states $s_1$ and $s_2$. Genre $3$ exhibits the highest difference in reward between the states.}
     \label{fig:examplea}
 \end{subfigure}
 \hfill
 \begin{subfigure}[b]{0.45\textwidth}
     \centering
     \includegraphics[width=\textwidth]{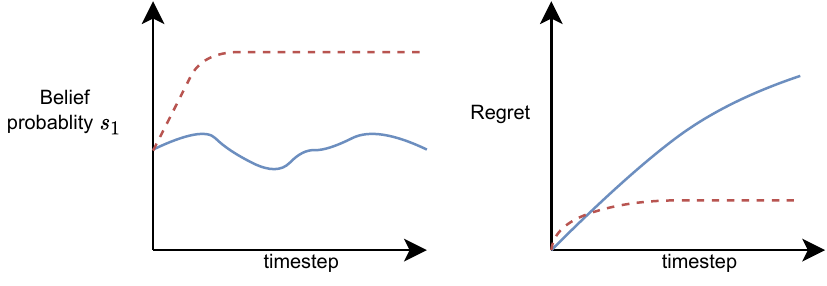}
     \caption{State belief using cumulative regret from movie genre $1$ and $2$ exclusively (blue, solid) and using movie 3 occasionally (red, dashed).}
     \label{fig:exampleb}
 \end{subfigure}
 \caption{Qualitative example scenario where the highest-reward arm is not ideal for quick identification of the state, leading to significant regret over longer periods of time. An information-gathering arm helps to significantly reduce regret through better (almost immediate) state identification.}
 \label{fig:example}
\end{figure}

Earlier work assumed a priori that it is reasonable only to consider movies that would score high in both states. The agent uses the feedback over a sequence of recommendations to determine what state is being served. While minimizing regret in the short term, such a movie recommendations strategy may take a long time to identify the latent state. Figure \ref{fig:examplea} shows the rewards distributions, where movies genres $1$ and $2$ have similar rewards, while genres $3$ has very different rewards between the two states. From the agent's perspective, any feedback received from movies in genre $1$ or $2$ could have come from either state.

While the cumulative rewards is expected to be generally high, the uncertainty in the belief-state of the agent may lead to continuously recommending from movie genres that are sub-optimal given the true latent state $s_1$. This results in high cumulative regret, as shown in figure \ref{fig:exampleb}. The blue line follows the strategy of only recommending movies from the highest reward genres $1$ and $2$. The agent samples the genres according to it's belief over latent states. Due to very similar reward distributions, the agent has a hard time identifying the true state, continuing to select movies from genre $2$ that are sub-optimal. The red line follows the strategy of initially recommending movies from genre $3$. This quickly identifies the state but comes with initially less reward. The reduced state uncertainty lets the agent choose movies from the best movie genre, maximizing cumulative reward.

\begin{figure}
 \begin{subfigure}[b]{0.45\textwidth}
     \centering
     \includegraphics[width=\textwidth]{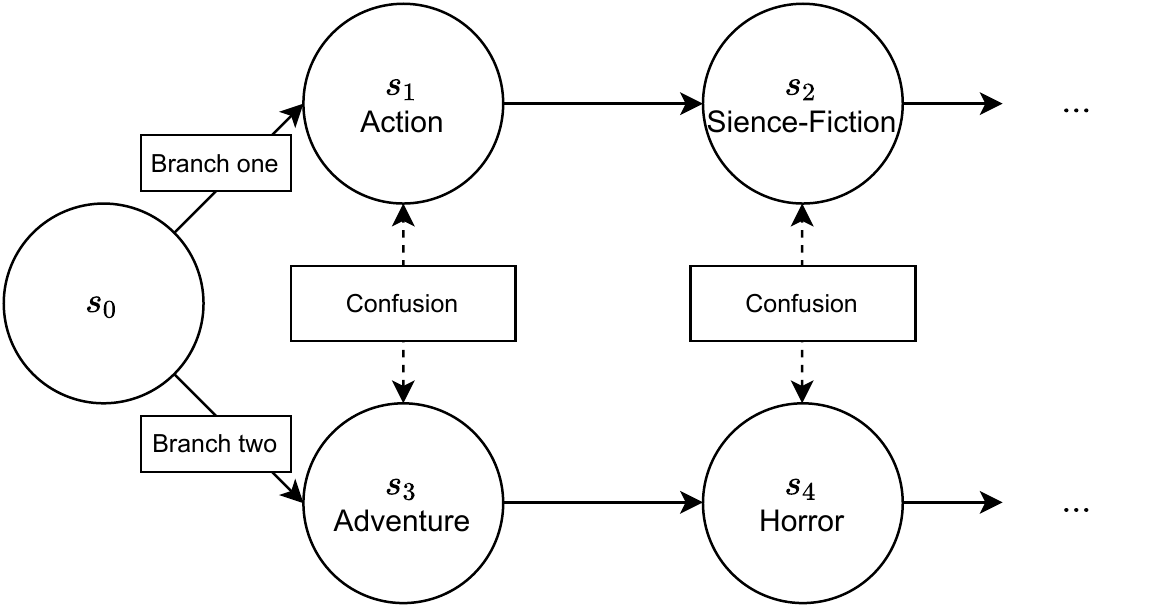}
     \caption{Example state transition structure. Rewards are similar between the two branches when recommending the best genre in each state.}
     \label{fig:examplenona}
 \end{subfigure}
 \hfill
 \begin{subfigure}[b]{0.45\textwidth}
     \centering
     \includegraphics[width=\textwidth]{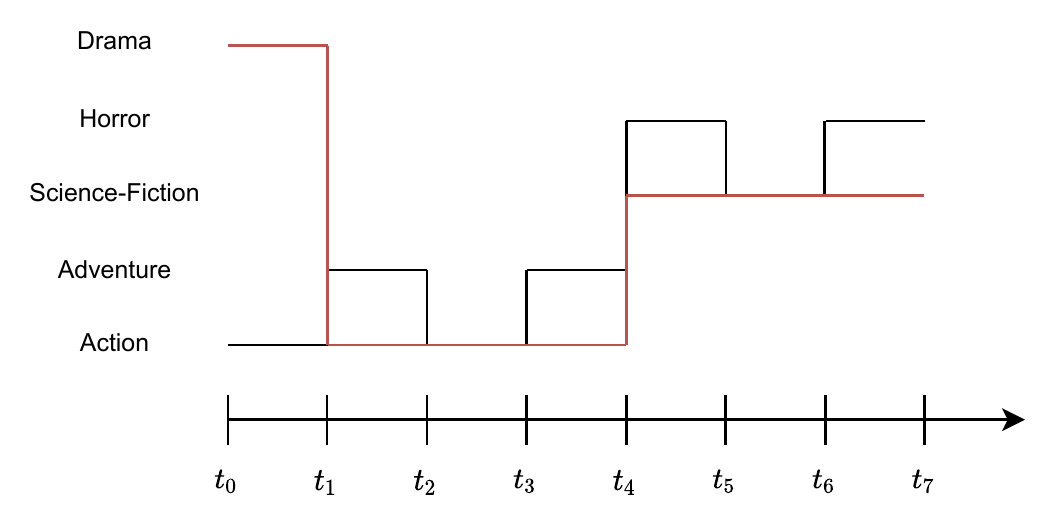}
     \caption{Genre selected by choosing always the best genre (black) according to the belief-state and information-gathering (red) via the Drama genre. The agent can discount one branch, thus it recommends the best movie genre more frequently in the future. The state changes to $s_2$ at $t_3$.}
     \label{fig:examplenonb}
 \end{subfigure}
 \caption{Non-stationary example where the state changes from $s_0$ into either branch one or branch two. Fast state identification helps concentrate the believe-state around a particular branch, resulting in better recommendations in the future.}
 \label{fig:long_term_example}
\end{figure}

Let us extend this example to a setting where the state changes over time. Quickly identifying the current state has additional benefits for state identification in the long term. Figure \ref{fig:long_term_example} shows a non-stationary example, where reward distributions between states are pairwise similar. Each state has one movie genre providing the highest reward if recommended. \textit{Action} and \textit{Adventure} genres have similar reward distributions in states $s_1$ and $s_3$, whereas \textit{Science-Fiction} and \textit{Horror} genres have similar reward distributions in $s_2$ and $s_4$.

State $s_0$ is the start state and switches to $s_1$ in branch one. Due to the change in rewards the agent receives, the belief over states changes from $s_0$ to $s_1$ and $s_3$. The states are confused, and the agent recommends both \textit{Action} and \textit{Adventure} movies. Before the agent can resolve its confusion, the user switches to $s_2$. The agent's belief remains split between the two branches, thus it needs to reconsider recommending movies from $s_2$ and $s_4$. On the other hand, if the agent can identify the branch early (through an information-gathering arm when the environment is still in $s_1$), it knows to recommend movies from $s_2$ exclusively. Figure \ref{fig:examplenonb} shows the recommendation behavior of the agent. We assume that a \textit{Drama} genre has a significantly different reward distribution between states $s_1$ and $s_3$. Selecting the \textit{Drama} genre in the beginning allows the agent to discard one of the state branches from consideration. While reward distributions of the best movie genres in $s_2$ and $s_4$ may lead to confusion, the agent has access to the transition matrix, knowing that the environment cannot be in state $s_4$ coming from state $s_1$. Discounting one branch effectively reduces the amount of genres the agent considers for recommendations, thus reducing the overall number of potential sub-optimal movie choices in the future.

\section{Problem Formulation}\label{sec:problem}

The notation used in this paper is as follows. The set of arms available for decision making is $\mathcal{A} = [K]$. The set of states is denoted as $\mathcal{S}$, where the number of states is much less than the number of arms, i.e.,  $|\mathcal{S}| \ll K$. The set of contexts is $\mathcal{X}$.  We use capitalized letters for all random variables.

The \textit (stationary) {latent bandit}~\citep{Maillard14} is an online learning problem with \textit{bandit feedback}, that is, only the reward of the chosen arm is revealed to the agent. The process goes as follows. At every time step $t = 1, 2 , \dots, n$:

\begin{enumerate}
    \item the agent receives context $X_t \in \mathcal{X}$.
    \item the agent chooses an arm $A_t \in \mathcal{A}$ according to its \textit{policy}, mapping \textit{history} $\mathcal{H}_t = (X_1, A_1, R_1,\dots, X_{t-1}, A_{t-1}, R_{t-1})$ and context $X_t$ to arms in $\mathcal{A}$. 
    \item the environment reveals the reward $R^{A_t} \in \mathbb{R}$ according to the joint conditional reward distribution $P(\cdot | A, X, S; \theta)$, parameterized by $\theta \in \Theta$, where $\Theta$ is the space of plausible reward models.
\end{enumerate}

The latent state $S_t \in \mathcal{S}$ is sampled according to a prior distribution $S_t \sim P_t(s)$, with the initial latent state $S_1 \sim P_1(s)$.
We define the mean reward of arm $A_t$ in context $X_t$ and latent state $S_t$ under the model parameters $\theta^*$, as $\mu(A_t, X_t, S_t;\theta^*) = \EX_{R\sim P(R | \cdot)}[R]$. Since the reward model is provided in the latent bandit setting, there are no strong assumptions placed on the form of $P(\cdot | A, X, S; \theta)$, which can be an arbitrarily complex function of $\theta$ and contexts $X$ can come from an arbitrary process. We only assume that the rewards for a particular $A_t$, $X_t$ and $S_t$ are Gaussian distributed, with mean $\mu(A_t, X_t, S_t;\theta^*)$ and variance proxy $\sigma^2$.

The performance of a bandit algorithm is measured through the regret an agent incurs by choosing a sub-optimal arm at time step $t$. Given the latent state $s^*$, we define $A_t^* = \argmax_{a \in \mathcal{A}} \mu(a, X_t, s^*;\theta^*)$ as the optimal arm as a function of context $X_t$ and parameters $\theta^*$. For the stationary latent bandit setting, we define the expected n-time step regret as

\begin{equation}
    \mathcal{R}(n, s^*) = \EX\Bigg[\sum_{t=1}^n\mu(A_t^*, X_t, s^*;\theta^*) - \mu(A_t, X_t, s^*;\theta^*)\Bigg],
\end{equation}
where the expectation is taken over the agent's randomness in the policy, as well as the randomness of $X_t$. In order to capture the algorithm's performance over a range of different initial latent states, we consider the Bayes regret. We compute the n-time step regret as an expectation over latent state randomness. The n-time step Bayes regret is defined as: 

\begin{equation}
    \mathcal{BR}(n) = \EX [\mathcal{R}(n; S^*) | \theta^*)].
\end{equation}

We also investigate settings where latent states can evolve. In the non-stationary latent bandit setting ~\citep{Hong2020} the initial latent state $S_t \in \mathcal{S}$ is sampled according to a prior distribution $S_1 \sim P_1(s)$ and evolves over time according to parameterized transition kernel $P(S|S_{t-1};\phi)$, with $\phi \in \mathbb{R}^{|\mathcal{S}| \times |\mathcal{S}|}$, that maps the current state to a distribution over next states. Contrary to the full reinforcement learning setting, the agent does not affect the environment dynamics through its arms. Only the current state determines the distribution over the next latent state. 

We define $A_t^* = \argmax_{A \in \mathcal{A}} \mu(a, X_t, S_t;\theta^*)$ as the optimal arm in latent state $S$ as a function of context $X$ and parameters $\theta$. For a fixed latent state sequence $s_{1:n} \in \mathcal{S}^n$, the expected n-time step regret is defined as:

\begin{equation}
    \mathcal{R}(n, \phi^*, S_{1:n}) = \EX\Bigg[\sum_{t=1}^n\mu(A_t^*, X_t, S_t;\theta^*) - \mu(A_t, X_t, S_t;\theta^*)\Bigg].
\end{equation}

Again, we consider the Bayes regret, computing the n-time step regret as an expectation over latent state randomness. The n-time step Bayes regret for the non-stationary setting is defined as:

\begin{equation}
    \mathcal{BR}(n;\theta^*;\phi^*) = \EX_{S_{1:n} \sim \phi^*} [\mathcal{R}(n, \phi^*, S_{1:n}) | \theta^*, \phi^*].
\end{equation}

We note that Bayes regret is considered a weaker benchmark than worst-case regret since we measure the performance as an average over latent states and latent state transitions. It does not adequately capture whether an algorithm performs significantly worse than another for a particular environment configuration. In practice, however, this is often a good metric to gauge the algorithm's performance since it is crucial to perform well for most users in a particular context.

\section{Algorithm}\label{sec:algo_complete}

In this section, we present our algorithm Active Greedy Exploration Model-Based Thompson Sampling (AGEmTS). We start by developing a strategy for the simple two-state stationary setting. Our strategy choose information-gathering arms to quickly uncover the latent state, thus improving future arm choices. We then describe the full AGEmTS algorithm, which uses the ideas developed in two-state setting to estimate the usefulness of information-gathering for reducing n-time step regret in more complex non-stationary multi-state settings.    

\subsection{A Strategy using Information-Gathering Arms in the Two-state Stationary Problem}

Due to its good practical performance, we choose \textit{model-based Thompson sampling} (mTS) algorithm from ~\cite{Hong2020} as a baseline. They have shown that mTS outperforms contextual LinTS and LinUCB algorithms that use change-point detection schemes as well as EXP4. They evaluated mTS on synthetic data and the MovieLens 1M data set, showing significantly better performance other state-of-the-art algorithms.

mTS is a posterior sampling technique and uses a posterior update rule shown in equation \ref{eq:post_update}. mTs chooses an arm according to its probability to be optimal given the current context $X_t$ and history $\mathcal{H}_t$, that is, $\mathbb{P}(A_t = a | \mathcal{H}_t, X_t) = \mathbb{P}(A^*_t = a | \mathcal{H}_t, X_t)$. mTS samples a state $B_t \in \mathcal{S}$ from its the posterior distribution over the belief-state and selecting the arm with the maximum mean reward given the sampled $B_t$:

\begin{equation}
    A_t = \argmax_{a \in \mathcal{A}} \mu(a, X_t, B_t;\theta^*)
\end{equation}

After the reward is received for the chosen arm, the posterior of the belief-state is formed using Bayes rule 

\begin{equation}\label{eq:post_update}
P_{t+1}(S_{t+1}) \propto \sum_{S_t \in \mathcal{S}} P_t(S_t)P(S_{t+1} | S_t;\phi^*) P(R_t | A_t, X_t, S_t;\theta^*).
\end{equation}

See algorithm \ref{alg:ps} for the pseudo-code of mTS. Note that this formulation of the algorithm requires a state transition model. For the stationary setting, we simply define the identity transition kernel $P(S|S_{t-1};\phi^*)$ with $\phi^* = I^{|\mathcal{S}| \times |\mathcal{S}|}$.

\begin{algorithm}
\caption{mTS~\citep{Hong2020}}
\label{alg:ps}
\begin{algorithmic}[1]
\State \textbf{Input:} 
\State \;\;\;\; Model parameters $\theta^*, \phi^*$, 
\State \;\;\;\; Prior over initial latent state $P_1(s)$
\For{$t=1,2,\dots,n$}
    \State Sample $B_t \sim P_t$
    \State Select the best arm according to $B_t$: $A^*_t = \argmax_{a \in \mathcal{A}}  \mu(A, X_t, B_t; \theta^*)$
    \State Observe $R_t$ and update state posterior using equation \ref{eq:post_update}
\EndFor
\end{algorithmic}
\end{algorithm}

As mentioned, mTS chooses the best arm given it's belief-state. Arms that might be candidates for information-gathering are not deliberately chosen. For the stationary two-state problem, we develop a simple strategy that achieves better regret than mTS and forms the principal basis for our algorithm. In a nutshell, the idea of this strategy is to estimate the regret for $n$ timesteps in the future under an assumption of using information-gathering arms.  

\subsubsection{Explore-commit}
We can achieve sub-linear regret with high probability in the two-state stationary problem straightforwardly using an explore-commit strategy. We frame the problem as a hypothesis testing. The goal, thus, is to sample enough information-gathering arms to assign the empirical reward mean of the information-gathering arm $a^e$ to either state $s_0$ or $s_1$. Let $R^{a^e}_{s_0,1}, \dots, R^{a^e}_{s_0,N}$; $R^{a^e}_{s_1,1}, \dots, R^{a^e}_{s_1,N}$ be reward realizations of two independent $\mathcal{N}(\mu_1, \sigma_1^2)$ and $\mathcal{N}(\mu_2, \sigma_2^2)$ for the information-gathering arm in state $s_0$ and $s_1$ respectively. We can simply compute the number $n^e$ of plays of information-gathering arms needed to detect a difference in means of at least $\mu_1 - \mu_2$, where $\mu_1 = \mu(a^e, s_0) $ and $\mu_2 = \mu(a^e, s_1) $. In other words, we would like to know the number of samples needed to detect a shift $\delta = \mu_2 - \mu_1$ from $\mu_1$ and, conversely, a shift $\delta$ from $\mu_2$.  

We focus on detecting a shift from $s_0$ without loss of generality, since the same computation holds for $s_1$. We define the null hypothesis that there is no statistically significant shift, i.e., $\delta = 0$, since the samples come from $s_0$. Alternatively, if a shift exists $\delta \neq 0$, the samples come from $s_1$. Since the reward distribution is a assumed to be normally distributed, we define the z score as

\begin{equation}
    z = \frac{\bar{R} - \mu_1}{\sigma_1/\sqrt{n_1}},
\end{equation}
where $\bar{R}$ is the sample average of the rewards received and $n_1$ is the sample size. For a confidence level $\alpha$ we can write the confidence interval as:

\begin{equation}
    \bar{R}  - z_{\alpha/2}\frac{\sigma_1}{\sqrt{n_1}} \leq \mu_1 \leq \bar{R}  + z_{\alpha/2}, \frac{\sigma_1}{\sqrt{n_1}},
\end{equation}
where $z_{\alpha/2}$ is the $z$ score and $\alpha/2$ is the tail-area of the standard normal distribution, respectively.

Since we do not a priori know the latent state, we want to make sure the power $1-\beta$ of the test is high, i.e., that we have enough samples to reduce the chance of type II errors, that is, committing to $s_0$ even though $s_1$ is the true state. 
Using the z score, we can compute the upper confidence limit (UCL), i.e., the maximum difference $\delta$ from $\mu_1$ before we reject the null hypothesis. The UCL is defined as
\begin{equation}
    UCL = \mu_1 + z_{1-\alpha/2}\frac{\sigma_1}{\sqrt{n_1}} = \mu_2 - z_{1-\beta}\frac{\sigma_1}{\sqrt{n_1}},
\end{equation}

where $1-\beta$ is the power of the hypothesis test. We can rearrange the UCL as

\begin{align}
    \begin{split}
    \delta & = z_{1-\alpha/2}\frac{\sigma_1}{\sqrt{n_1}} + z_{1-\beta}\frac{\sigma_1}{\sqrt{n_1}}\\
    & = (z_{1-\alpha/2} +  z_{1-\beta})\frac{\sigma_1}{\sqrt{n_1}}.
    \end{split}
\end{align}

Thus, the number of samples needed to detect the difference $\delta$ for a certain confidence level $\alpha$, and power $\beta$ are

\begin{equation}
    n_1 = \frac{(z_{1-\alpha/2} + z_{1-\beta})^2\sigma_1^2}{(\mu_1 - \mu_2)^2},
\end{equation}

where we substituted $\delta = (\mu_2 - \mu_1)^2$. $z_{1-\alpha/2}$ and $z_{1-\beta}$ are $z$ scores for $p$-values $\alpha$ and $\beta$ of the standard normal distribution, respectively. For example, for $\alpha = 0.95$ and $\beta = 0.8$, we have $z_{1-\alpha/2} \approx 1.96$ and $z_{1-\beta} \approx 0.84$. Note the same holds for the number of samples $n_2$ when rewards come from $s_1$, replacing $\sigma_1$ with $\sigma_2$. Finally, in order to get the number of information-gathering arms, we need to decide to either choose $n_1$ or $n_2$ number of information-gathering arms. We assume no bias towards either state, therefore we err on the safe side and choose the larger of the two sample sizes as $n^e = \max(n_1, n_2)$. Then, we compute the empirical reward mean $\bar{R}$ from the $N$ information-gathering arms and choose the state who's mean reward for $a^e$ is closest, that is

\begin{equation}
    s^* = \argmin_{s \in \mathcal{S}} |\bar{R} - \mu(a^e, s)|.
\end{equation}

In our synthetic experiments, we have for the mean and standard deviation of the information-gathering arm in states $s_0$ and $s_1$, $\mu^{a^e}_{s_0} = 1.7$, $\sigma^{a^e}_{s_0} = 0.05$ and $\mu^{a^e}_{s_1} = 1.5$, $\sigma^{a^e}_{s_1} = 0.05$ respectively. The best arms have mean rewards $\mu^*_{s_0} = \mu^*_{s_1} = 2.1$, and standard deviations $\sigma^*_{s_0} = \sigma^*_{s_1} = 0.5$, where $a^*_{s_0} \neq a^*_{s_1}$. The second best arm has $\mu_{s_0} = \mu_{s_1} = 2.05$. Thus, the best arm in state $s_0$ is the second best arm in $s_1$ and vice versa. The following reward calculations assume the latent state is $s_0$ and serve as an illustration. For the implementation, we compute a simple average over both states for estimated rewards. 

The explore-commit strategy initially chooses $n^e$ information-gathering arms and thus incurs in expectation with high probability $\EX [R^{a^e}] = n^e (R^*_{s_0} - R^{a^e}_{s_0})$ regret, assuming the true latent state of the environment is $s_0$. In order to decide whether our explore-commit strategy has any benefit over posterior sampling, we need to consider the expected regret incurred using posterior sampling and find the break-even point in regret between both strategies. The expected belief-state update can be computed by iteratively solving the differential equation

\begin{equation}\label{eq:diff_eq}
    \frac{\delta P_t(s_0)}{\delta t} = \frac{P_t(s_0)P(R_t | A_t, X_t, s_0;\theta^*)}{P_t(s_0)P(R_t | A_t, X_t, s_0;\theta^*) + (1-P_t(s_0))P(R_t | A_t, X_t, s_1;\theta^*)} - P_t(s_0).
\end{equation}

The differential equation can simply derived through application of the Bayes rule (equation \ref{eq:post_update}) for $|\mathcal{S}| = 2$. The expected regret using posterior sampling is simply

\begin{equation}
    R^{ps} = \EX \Bigg[\sum_{t=1}^n (R^*_{s_0} - R'_{s_0})\Big(1-\frac{\delta P_t(s_0)}{\delta t}\Big) \Bigg],
\end{equation}

where the expectation is taken over the agent's sampling of the arms based on its belief-state. $R'_{s_0}$ is the expected reward obtained in state $s_0$ when choosing $a^*_{s_1}$, the best arm in state $s_1$. For this simple setting, if $n^e \leq n$ and $R^{a^e} < R^{ps}$, we can use the explore-commit strategy. We can fall back to posterior sampling if the conditions are not met. The goal is to avoid the unnecessary choice of information-gathering arms if we expect that explore-commit will not have a higher cumulative reward than posterior-sampling after $n$ time steps.

\subsubsection{Explore than Posterior Sampling}
\begin{figure}
     \centering
     \begin{subfigure}[b]{0.45\textwidth}
         \centering
         \includegraphics[width=\textwidth]{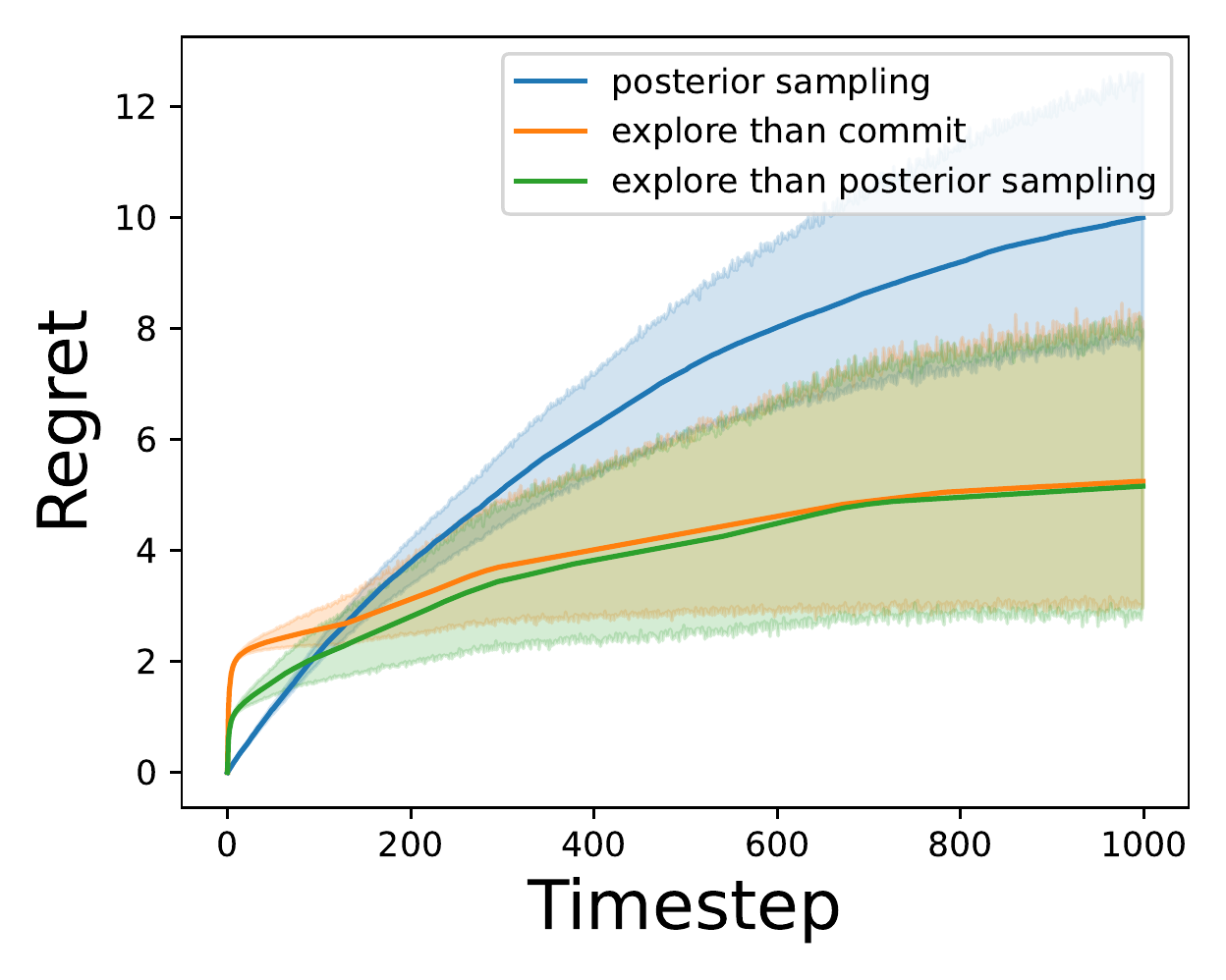}
         \caption{Comparison between posterior sampling, explore-commit and explore-than-posterior-sampling. $\alpha = 0.05$ and $\beta = 0.2$.  Shaded area shows 95 \% CI.}
         \label{fig:multistate_synth_examples_regretA}
     \end{subfigure}
     \hfill
     \begin{subfigure}[b]{0.45\textwidth}
         \centering
         \includegraphics[width=\textwidth]{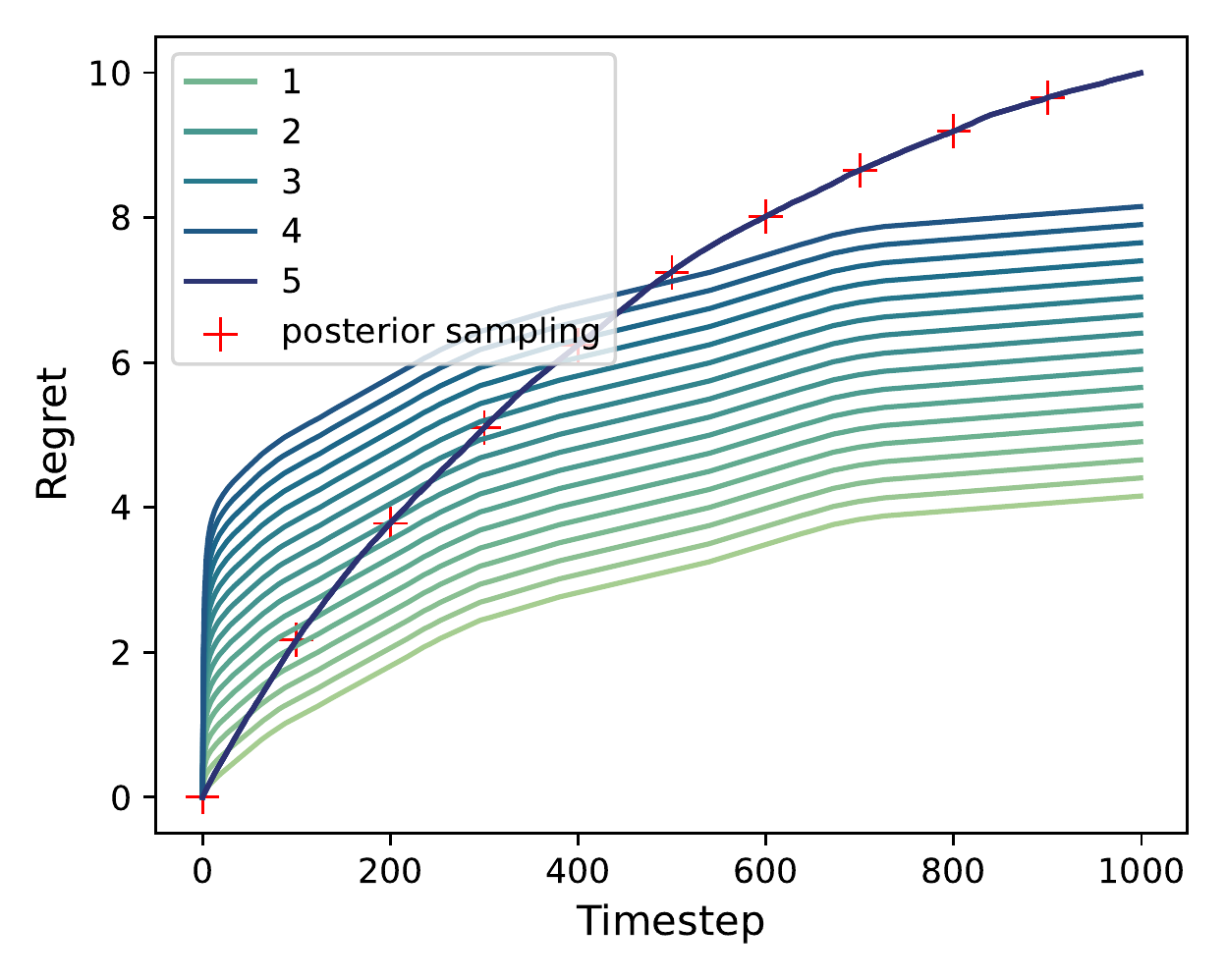}
         \caption{Regret using explore-than-posterior-sampling strategy. Mean reward difference between the best arms and information-gathering arm (color gradient. The legend shows different fixed values of $\Delta R$). The explore than posterior sampling strategy does not choose the information-gathering arm if the regret incurred is to high.}
         \label{fig:multistate_synth_examples_regretB}
     \end{subfigure}

     \caption{Regret for the two-state stationary example using the posterior sampling, explore-commit and explore-than-posterior-sampling strategies. (a) We observe a significant improvement in regret for the two information-gathering strategies. Explore-than-posterior-sampling incurs less regret than explore-commit in the early stages of interaction. (b) If the regret incurred through information-gathering arms is significant, explore-than-posterior-sampling falls back to simple posterior sampling (shown as red crosses).}
     \label{fig:simple_example}
\end{figure}

While simple and effective, we do not use an explore-commit strategy in our algorithm since it has the disadvantage of leading to linear regret, albeit with a low probability. Sampling $n^e$ information-gathering arms will only uncover the true hidden state with a certain confidence. Thus, we would like to continue to update our state beliefs over time. We call the following alternative strategy \textit{explore than posterior sampling}. We compute the expected regret using $\tau$ information-gathering arms followed by posterior sampling for $n-\tau$ time steps by minimizing the following objective:

\begin{equation}\label{eq:r_e}
     R_e =  \argmin_{\tau \in n} \sum_{t=1}^\tau (R^*_{s0} - R^{a^e}_{s0}) +  \EX \Bigg[\sum_{t=\tau+1}^n 1-\frac{\delta P_t}{\delta t}  (R^*_{s_0} - R'_{s_0})\Bigg],
\end{equation}

where the expectation is taken over the agent's sampling of the arm based on the belief-state. This strategy samples information-gathering arms enough to uncover the hidden state sufficiently and continues to use posterior sampling to update its belief over time. AGEmTS uses a similar strategy for the non-stationary multi-state setting as a \textit{reward estimator} (algorithm \ref{alg:2}). The regret of the two strategies on the two-state stationary example is shown in Figure \ref{fig:multistate_synth_examples_regretA}. We can see that explore-commit and explore than posterior sampling leads to improved n-step regret. Additionally, we avoid sampling information-gathering arms if the gained reward through better state identification does not compensate for the incurred regret through information-gathering, shown as the darkest blue line in Figure \ref{fig:multistate_synth_examples_regretB} overlapping with the red crosses.

Overall, the idea behind our proposed algorithm is that we occasionally but deliberately choose an information-gathering arm to reduce the uncertainty in the current state. Consequently, the agent chooses the highest reward arm according to its more accurate belief-state with higher frequency in future time steps. Information-gathering does not necessarily provide the highest reward, though, therefore we require the agent to estimate the benefit of information-gathering before doing it. 

\subsection{Active Greedy Exploration Model-Based Thompson Sampling (AGEmTS)}\label{sec:algo}

The strategies in the previous sections were developed for the stationary two-state latent bandit settings. Naturally, for the case where $|\mathcal{S}| > 2$, the strategies need to be adjusted to properly update the belief-state posterior and reward estimates. The obvious limitation comes from the fact that these algorithm are for the stationary setting, thus no mechanism exists to sample information-gathering arms again, after the first rounds. Furthermore, the differential equation \ref{eq:diff_eq} is only defined for two states and we need to compute expected believe state updates for the multi-state case. Before describing AGEmTS in detail, we mention relevant differences compared to mTS, which our algorithm is based on. 

We do not sample the state from the posterior over the belief-state, but always select the most likely state $B_t = \argmax_{s \in \mathcal{S}} P_t(s)$. This will render state selection, and thus arm selection, greedy. Greedy arm selection may result in choosing sub-optimal arms for long time periods. We compensate for this via information-gathering arms that can effectively reduce state uncertainty. Moreover, the systematic selection of the sub-optimal arm is limited by updating the belief-state posterior, eventually converging to the correct latent state. 

Choosing arms with significantly different reward distributions allow for a fast reduction in state uncertainty but may come with a less average reward compared to other arms. The higher the cost for information-gathering, the less favorable the information-gathering arm, since we can expect that the investment in regret will be harder to recuperate through better state identification. Thus, AGEmTS will either select the best arm according to the belief-state, like mTS, or attempt to reduce state uncertainty via information-gathering arms, depending on the current ``confusion'' and impact on cumulative reward over a time horizon.    

We start with the detailed description of the algorithm. As mentioned, instead of always selecting the arm with the highest reward given the belief state, the algorithm determines the level of ``confusion'' between states by computing the information entropy of the belief-state $P_t(s)$ as

\begin{equation}
    H(P_t) = - \sum_{s \in \mathcal{S}} P_t(s_t) \log_2P_t(s_t).
\end{equation}

If there exists a significant confusion between two or more states (i.e., the belief-state is not significantly concentrated around a particular state), we have $H(P_t) \geq 1$. For the case of $|\mathcal{S}| = 2$, the information entropy $H(P_t) = 1$ if both states are equally likely.
If $H(P_t) \geq 1$, the algorithm computes the benefit of an information-gathering arm in the following way. For normally distributed arm rewards, the $KL$-divergence $D_{KL}(P||Q)$ is defined as

\begin{equation}
    D_{KL} (\mathcal{N}(\mu_1, \sigma_1) \; || \; \mathcal{N}(\mu_2, \sigma_2)) = \log \frac{\sigma_2}{\sigma_1} + \frac{\sigma_1^2 + (\mu_1 - \mu_2)^2}{2\sigma_2^2} - \frac{1}{2},
\end{equation}

where $\mu_1$ and $\mu_2$ are the mean rewards and $\sigma_1$ and $\sigma_2$ are the reward standard deviations of the arms to be compared. The algorithm computes the mean $\bar{D}_{KL}$ for each arm in the arm set by comparing it against any other arm in the arm set $\mathcal{A}$. Furthermore, we compute the pairwise regret between arms over all states $s \in \mathcal{S}$ and contexts $x_i \in \mathcal{X}$ to estimate the cost. The two quantities are defined as

\begin{align}\label{eq:dkl}
    \bar{D}_{KL}(a_i) = \frac{1}{|\mathcal{X}|} \sum_{x_i \in \mathcal{X}}\frac{1}{|\mathcal{S}|} \sum_{s \in \mathcal{S}} \frac{1}{|\mathcal{A}|}\sum_{a_i \neq a_j}^{|\mathcal{A}|}
    D_{KL}\bigg(\mathcal{N}\big(\mu(a_i, x_i, s; \theta^*), \sigma^2_{a_i}\big), \mathcal{N}\big(\mu(a_j, x_i, s; \theta^*), \sigma^2_{a_j}\big)\bigg), \\
    \label{eq:dr}
    \overline{\Delta R}(a_i) = \frac{1}{|\mathcal{X}|} \sum_{x_i \in \mathcal{X}} \frac{1}{|\mathcal{S}|} \sum_{s \in \mathcal{S}} \frac{1}{|\mathcal{A}|}\sum_{\{a_i,a_j\} \in \mathcal{A}} \mu(a_i, x_i, s; \theta^*) - \mu(a_j, x_i, s; \theta^*).
\end{align}

The roll-out is performed for both posterior sampling and information-gathering. That is, we compute the reward as if the algorithm plays $a_t$ only or $a^e$ occasionally. We define two roll-out belief-states, $P_{ig}(s)$ and $P_{ps}(s)$, for using information-gathering and posterior sampling respectively. During the roll-out, AGEmTS might try to sample an information-gathering arm again, if $H\big(P_{ig}(s)\big) \geq 1$. We account for this by
simulating the choice of $a^e$ during the roll-out. We then update $P_{ig}(s)$ as follows:

\begin{equation}
     P_{ig}(s) \propto \sum_{s_i \in \mathcal{S}} P_{ig}(s) P(s|s_i;\phi^*) P^e_{roll}(s, s_i).
\end{equation}

Furthermore, since a simulated information-gathering arm $a^e$ has been sampled, we need to add the appropriate cost to $R^{ig}$. To account for estimation uncertainties in the roll-out, we opt to apply the worse-case cost to avoid overestimating the gained reward. Thus, we add the penalty $R^{ig}(s) = R^{ig}(s) - R_u$. We compute the upper bound on single step regret $R_u$ as:

\begin{equation}\label{eq:r_u}
    R_u = \argmax_{s \in \mathcal{S}} \big\{ \argmax_{a \in \mathcal{A}} \mu(a,\cdot, s, \theta^*) - \argmin_{a \in \mathcal{A}} \mu(a,\cdot, s, \theta^*) \big\}.
\end{equation}

Note that in the final algorithm, when simulating information-gathering in the roll-out phase, we additionally require that the minimum reward gain  $R^{ig} - R^{ps} > R_u$, to avoid oversampling of simulated information-gathering. In other words, AGEmTS ensures that the cumulative reward gain justifies the potential investment into the information-gathering arm.

If AGEmTS chooses not to gather information in the roll-out step, we update both posteriors as

\begin{align}
    P_{ig}(s) \propto \sum_{s_i \in \mathcal{S}} P_{ig}(s) P(s|s_i;\phi^*) P^*_{roll}(s, s_t),\\
    P_{ps}(s) \propto \sum_{s_i \in \mathcal{S}} P_{ps}(s) P(s|s_i;\phi^*) P^*_{roll}(s, s_i).
\end{align}

The algorithm then samples the next most likely states $B_{ig}$ and $B_{ps}$ according to $P_{ig}(s)$ and $P_{ps}(s)$, chooses greedily arms $a_{ig}$ and $a_{ps}$ and adds to the cumulative rewards $R^{ig} = R^{ig} + \mu(a_{ig})$.

Inspired by information-directed sampling~\citep{Russo2014}, we compute a metric akin to the information ratio. When gathering information, the algorithm chooses the arm that maximizes the ratio 

\begin{equation}\label{eq:inforatio}
    \Psi_{max} = \argmax_{a \in \mathcal{A}} \frac{\bar{D}_{KL}(a)}{\overline{\Delta R}^2(a)},
\end{equation}

where $\Psi_{max}$ encodes the intuition of the usefulness of an arm for information-gathering.
Let $a_e$ be the arm that maximizes $\Psi_{max}$. If $a^e$ is different to the arm $a_t$ chosen by maximizing over the belief-state, algorithm \ref{alg:2} computes the future reward of both $a_t$ and $a^e$ and chooses the arms that maximizes the reward over $t_{exp}$ time steps. $t_{exp}$ is defined as

\begin{equation}\label{eq:t_exp}
t_{exp}(s) = \sum_{s\in \mathcal{S}}\big(1^{|\mathcal{S}| \times 1}-\diag(P(\cdot | s))^{-1}\big) \cdot <P_t(\cdot | s), P^0(\cdot|s)>,
\end{equation}

where $P^0(s, s_t) \propto \sum_{s_i \in \mathcal{S}} P(s_i|s) - \mathbbm{1}\{s_i = s\}P(s_i|s)$ with $\mathbbm{1}$ being the indicator function. $P^0(s, s_t)$ is the distribution over possible next states. Given the current state $s_t$, we expect to \textit{have switched} to state $s'$ with probability $P_t^0(s', s_t)$. $t_{exp}$ is the average number time steps available to gather rewards before another switch is expected to occur. Thus, it allows the algorithm to gauge if the regret investment through information-gathering results in higher reward compared to mTS over period $t_{exp}$.  

The algorithm then runs a trajectory roll-out to compute the cumulative rewards $R^{ig}$ and $R^{ps}$ for all next states, given the current belief-state. $R^{ig}$ is the reward obtained when sampling information-gathering arms occasionally, while $R^{ps}$ is the reward obtain through a posterior sampling strategy (mTS). Simulating roll-outs with sampled state-trajectories and corresponding $a_t$ chosen by posterior sampling will return high-variance estimates of the reward. Therefore, we elect to use the \textit{weighted conditional reward distribution} for arm $a_t$. For all $t = 1, 2, \dots, t_{exp}$, we fix the next state $s_t$ and compute the probability of \textit{mean reward} $\mu(a, x_t, s_i)$ as an average over all arms in state $s_i$. We construct the matrix  $L$ from the probabilities that $\mu(a, x_t, s_i)$ came from state $s_i \in \mathcal{S}$ for all arms $a \in \mathcal{A}$ as

\begin{equation}
 L_{s_t} = \begin{bmatrix}
P\big(\mu(a_0, x_t, s_t)\;|\; a_0, x_t, s_0)\big)&\dots&  P\big(\mu(a_0, x_t, s_t)\;|\; a_0, x_t, s_{|\mathcal{S}|-1}\big)\\
\vdots &\ddots&\vdots\\
P\big(\mu(a_{|\mathcal{A}|-1}, x_t, s_t) \;|\; a_{|\mathcal{A}|-1}, x_t, s_0\big)&\dots& P\big(\mu(a_{|\mathcal{A}|-1}, x_t, s_t) \;|\; a_{|\mathcal{A}|-1}, x_t, s_{|\mathcal{S}|-1}\big)
\end{bmatrix}.
\end{equation}

$L_{s_t}$ is multiplied by the current belief-state $P_t$, generating a weighted average as

\begin{equation}\label{eq:prollmax}
    P^*_{roll}(s_i, s)= P_t(s) L_{s_i}^\top.
\end{equation}

$P^*_{roll}$ is used in computing the belief-state in the roll-out for arms chosen by posterior sampling. The roll-out conditional reward distribution for information-gathering arm $a^e$ is done in a similar fashion, but since the same arm $a^e$ is used for all states, it simplifies to

\begin{equation}\label{eq:prollexp}
    P^e_{roll}(s, s_i) = P_t(s) \odot \Big[P\big(\mu(a^e, x_t, s_i)\;|\; a^e, x_t, s_0)\big), \dots,   P\big(\mu(a^e, x_t, s_i)\;|\; a^e, x_t, s_{|\mathcal{S}|-1}\big)\Big],
\end{equation}

where $\odot$ denotes the component-wise multiplication. We note that the roll-out is done for all states $s \in \mathcal{S} \setminus {B_t}$, where $B_t$ is the current sampled state $B_t = \argmax_{s \in \mathcal{S}} P_t(s)$. Thus, we compute the \textit{average cumulative reward} for both $R^{ig}$ and $R^{ps}$ by averaging over the number of states $|S|-1$.

The two cumulative average rewards are compared and if the difference in favor of information-gathering exceeds the maximum investment $R_u$, the algorithm plays $a^e$ and $a_t$ otherwise. We note that it is not strictly necessary that the cumulative reward in favor of information-gathering exceeds a particular value. We have found it to be a reasonable choice to compensate for likely estimation errors throughout the roll-out and renders AGEmTS more conservative. Note that the same applies to the simulated information-gathering arm mentioned above.

Finally the posterior over the belief-state is updated according to
\begin{equation}\label{eq:posterior_update}
    P_{t+1}(s_{t+1}) \propto \sum_{s_t \in \mathcal{S}} P_t(s_t)P(s_{t+1} | s_t;\phi^*) P(R_t | A_t, X_t, s_t;\theta^*).
\end{equation}

The full algorithm is shown in Algorithm \ref{alg:1}, together with the reward estimator in Algorithm \ref{alg:2}.

\begin{algorithm}
\caption{AGEmTS}
\label{alg:1}
\begin{algorithmic}[1]
\State \textbf{Input:} 
\State \;\;\;\; Model parameters $\theta^*, \phi^*$, 
\State \;\;\;\; Prior over initial latent state $P_1(s)$
\State Compute upper bound on single step regret $R_u$ using equation \ref{eq:r_u}
\For{$t=1,2,\dots,n$}

    \State Select $B_t = \argmax_{s \in \mathcal{S}} P_t(s)$
    \State Select the best arm according to $B_t$: $a_t = \argmax_{a \in \mathcal{A}}  \mu(A, X_t, B_t; \theta^*)$
    \State $H(P_t) = -\sum_{s \in \mathcal{S}}P_t \cdot \log_2 P(s_t)$ \Comment{compute belief-state entropy}
    \If{$E(P_t) \geq 1$} 
    
        \State compute $\bar{D}_{KL}(a)$ using equation \ref{eq:dkl}
        \State compute $\overline{\Delta R}(a)$ using equation \ref{eq:dr}
        
        \State $a^e_t = \Psi_{max}$ \Comment{compute global best information-gathering arm}
        \If{$a_t \neq a^e_t$}
            \State compute future reward $R^{ig}$ and $R^{ps}$ using algorithm \ref{alg:2}
        \EndIf
        
        \State Compute reward difference $\delta_R = R^{ig} - R^{ps}$
        \If {$\delta_R > 0$  \textbf{and} $\delta_R > R_u$ } \Comment{reward gain exceeds upper bound on single step regret}
           \State $a_t \gets a^e_t$
        \EndIf
    \EndIf
\EndFor
\State Observe $R_t$ and update state posterior
\State $P_{t+1}(s_{t+1}) \propto \sum_{s_t \in \mathcal{S}} P_t(s_t)P(s_{t+1} | S_t;\phi^*) P(R_t | A_t, X_t, s_t;\theta^*)$
\end{algorithmic}
\end{algorithm}

\begin{algorithm}
\caption{Reward Estimator}
\label{alg:2}
\begin{algorithmic}[1]
\State \textbf{Input:} $a_t$, $a^e_t$, $P_t$, $R_u$, $B_t$

\For{$s \in \mathcal{S} \setminus {B_t}$}

    \State Compute expected time in next states $t_{exp}$ via equation \ref{eq:t_exp}
    \State $P_{ig}(s) \gets P_t$
    \State $P_{ps}(s) \gets P_t$
    \State $R^{ig}(s) \gets -R_u$
    \State $R^{ps}(s) \gets 0$

    \State  Compute $P_{roll}^*$ according to equation \ref{eq:prollmax}
    \State  Compute $P_{roll}^e$ according to equation \ref{eq:prollexp}
    
    \State $P_{ig}(s) \propto \sum_{s_t \in \mathcal{S}} P_{ig} P(s|s_t;\phi*) P_{roll}^e$ \Comment{Do one information-gathering arm and update posterior}
    
    \For{$t=1,2,\dots, t_{exp}$}

        \If{$H(P_{ig}(s)) \geq 1$  \textbf{and} $R^{ig}(s) - R^{ps}(s) > R_u$} \Comment{Explore again during roll-out}
        
            \State $P_{ig}(s) \propto \sum_{s_i \in \mathcal{S}} P_{ig} P(s|s_i;\phi^*) P_{roll}^e(s, s_i)$ 
            \State $R^{ig}(s) \gets -R_u$ \Comment{add penalty}
        \Else
          \State $P_{ig}(s) \propto \sum_{s_i \in \mathcal{S}} P_{ig} P(s|s_i;\phi^*) P_{roll}^*(s, s_i)$   
        \EndIf
        
        \State $P_{ps}(s) \propto \sum_{s \in \mathcal{S}} P_{ps}(s)P_{roll}^*(s, s_i)$

        \State $B_{ig} = \argmax_{s \in \mathcal{S}} P_{ig}(s)$ \Comment{get state according to explore belief-state}
        \State $B_{ps} = \argmax_{s \in \mathcal{S}} P_{ps}(s)$ \Comment{get state according to ps belief-state}
        \State $a_{ig} \gets \argmax_{a \in \mathcal{A}}  \mu(a, X_t, B_{ig}; \theta^*)$
        \State $a_{ps} \gets \argmax_{a \in \mathcal{A}}  \mu(a, X_t, B_{ps}; \theta^*)$
        
        \State $R^{ig}(s) \gets R^{ig}(s) +  \mu(a_{ig}, X_t, s; \theta^*)$
        \State $R^{ps}(s) \gets R^{ps}(s) +  \mu(a_{ps}, X_t, s; \theta^*)$
    \EndFor
\EndFor
\State $R^{ig} = \gets \frac{1}{|\mathcal{S}|-1} \sum_{s \in \mathcal{S}} R^{ig}(s)$
\State $R^{ps} = \gets \frac{1}{|\mathcal{S}|-1} \sum_{s \in \mathcal{S}} R^{ps}(s)$

\State \textbf{return} $R^{ig}$, $R^{ps}$
\end{algorithmic}
\end{algorithm}

\section{Experiments}\label{sec:exp}

We compare our algorithm to several baselines; as mentioned above, we start with mTS. Furthermore, we include an upper-confidence-bound algorithm \textbf{(CDUCB/CD-LinUCB)}~\citep{Auer2003b} and Thompson sampling algorithm \textbf{(CDTS/CD-LinTS)} ~\citep{agrawal2014thompson, Abeille2017}, both with a change-point detector. We include \textbf{mUCB}, which has been developed for the stationary latent bandit setting ~\citep{Hong2020b}. Moreover, we include \textbf{EXP4.S} with an enforced lower bound on the expert weights, achieving near-optimal regret in the piece-wise stationary bandit setting~\citep{Auer2003}. All the algorithms mentioned above have the conditional reward models as arms/experts and choose the best arm according to the respective conditional reward model.

The change-point detector for the CDUCB and CDTS algorithm computes the sum of rewards for each of the arms over a window $\tau$. If the difference between the sum of rewards of the last $\tau/2$ time steps in the window exceeds the sum of rewards in the first $\tau/2$ time steps by some threshold $b$, a change is detected. After the change is detected, the algorithms are reset. For the LinUCB and LinTS algorithms used in the real-world data experiments, we choose the change-point detector presented in ~\cite{Hong2020}. For timestep $t$ and window size $\tau$, the change-point detector computes weight vectors $\hat{W_t}$ via the least-squares solution of $X\hat{W}_t=R_t$ for arm $a$ using the features $X$ and rewards $R$ of the past $\tau/2$ time steps and $\hat{W_t}'$ for data of $\tau/2$ time steps before that. Given the empirical covariance matrix $\hat{\Sigma_t} = \sum_{t-\tau}^t X_{t,A_t}^\top X_{t,A_t}$, a change is detected when $||\hat{W_t} - \hat{W_t}'||_{\hat{\Sigma_t}} \geq b$. The weighted norm of a matrix $M$ is defined as $||M||_{\Sigma} = \sqrt{M^\top \Sigma M}$.

We run our experiments on synthetic data and on real-world data from MovieLens 1M. Furthermore, as a result of our experiments on the real-world data, we investigate where information-gathering arms can play a vital roll in improving regret.

\subsection{Synthetic Two-State Settings}

First, we investigate a simple stationary two-state problem. The stationary setting most clearly shows the significant regret incurred due to state confusion and serves as an illustrative example where information-gathering results in clear regret improvement. We follow up by a non-stationary two-state example that shows the need to correctly time the information-gathering arm to achieve the most benefit.

In the stationary setting, the latent state is randomly chosen and stays fixed for all time steps. The reward distribution between the state is chosen to be similar, such that it is difficult to distinguish them using the best arm of each state. Thus we choose mean and variance of the reward distributions so that $D_{KL}(R_{a^*_{s_1}} || R_{a^*_{s_2}}) = 0.005$. Note that, at the same time, we do not expect much regret since the rewards are similar.

\begin{figure}
     \centering
     \begin{subfigure}[b]{0.45\textwidth}
         \centering
         \includegraphics[width=\textwidth]{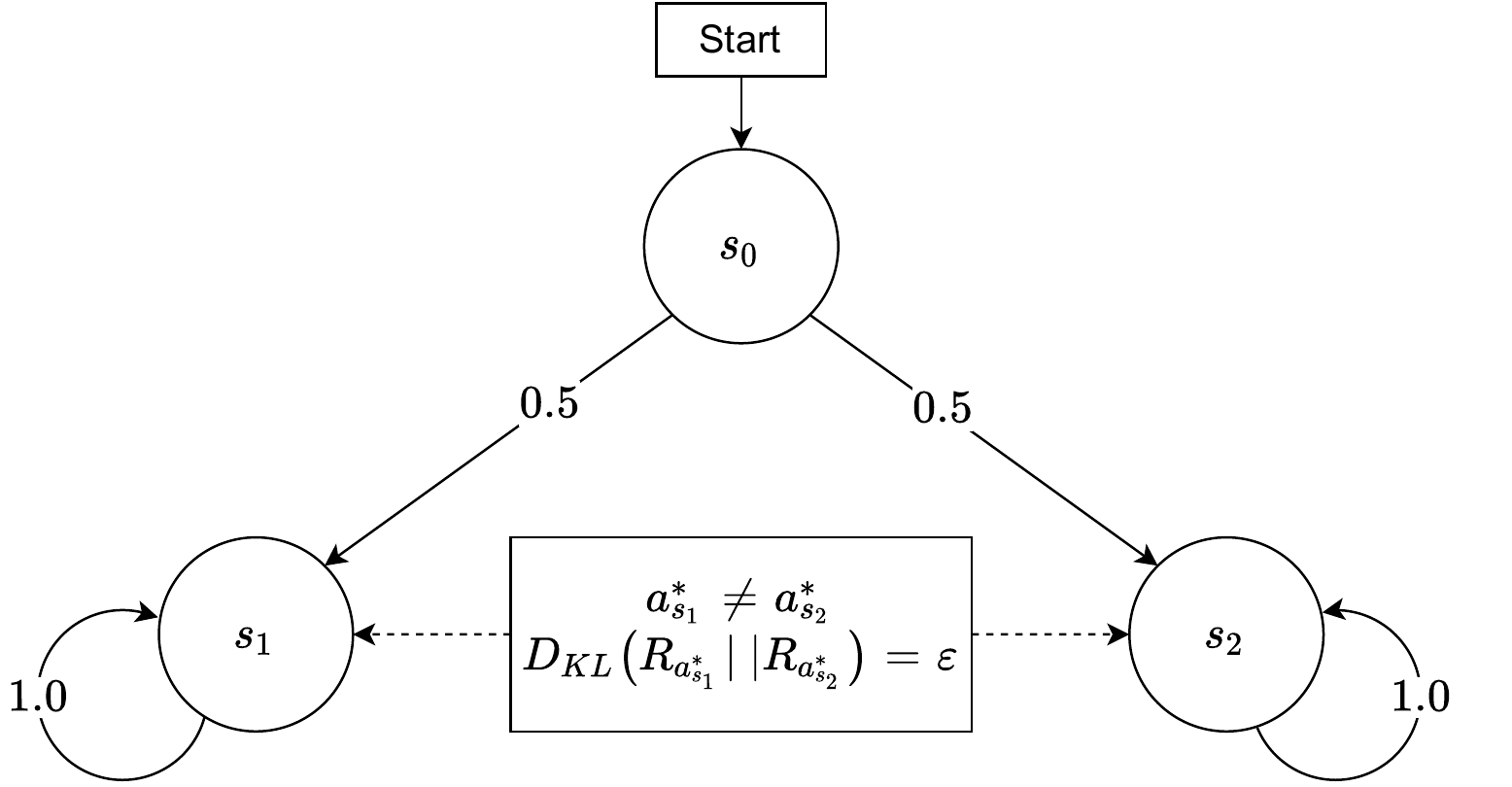}
         \caption{Two state stationary setting with $\epsilon = 0.005$. $S_0$ is the start state but switches immediately to $s_1$ or $s_2$, with equal probabilities.}
         \label{fig:twostatestationaryA}
     \end{subfigure}
     \hfill
     \begin{subfigure}[b]{0.45\textwidth}
         \centering
         \includegraphics[width=\textwidth]{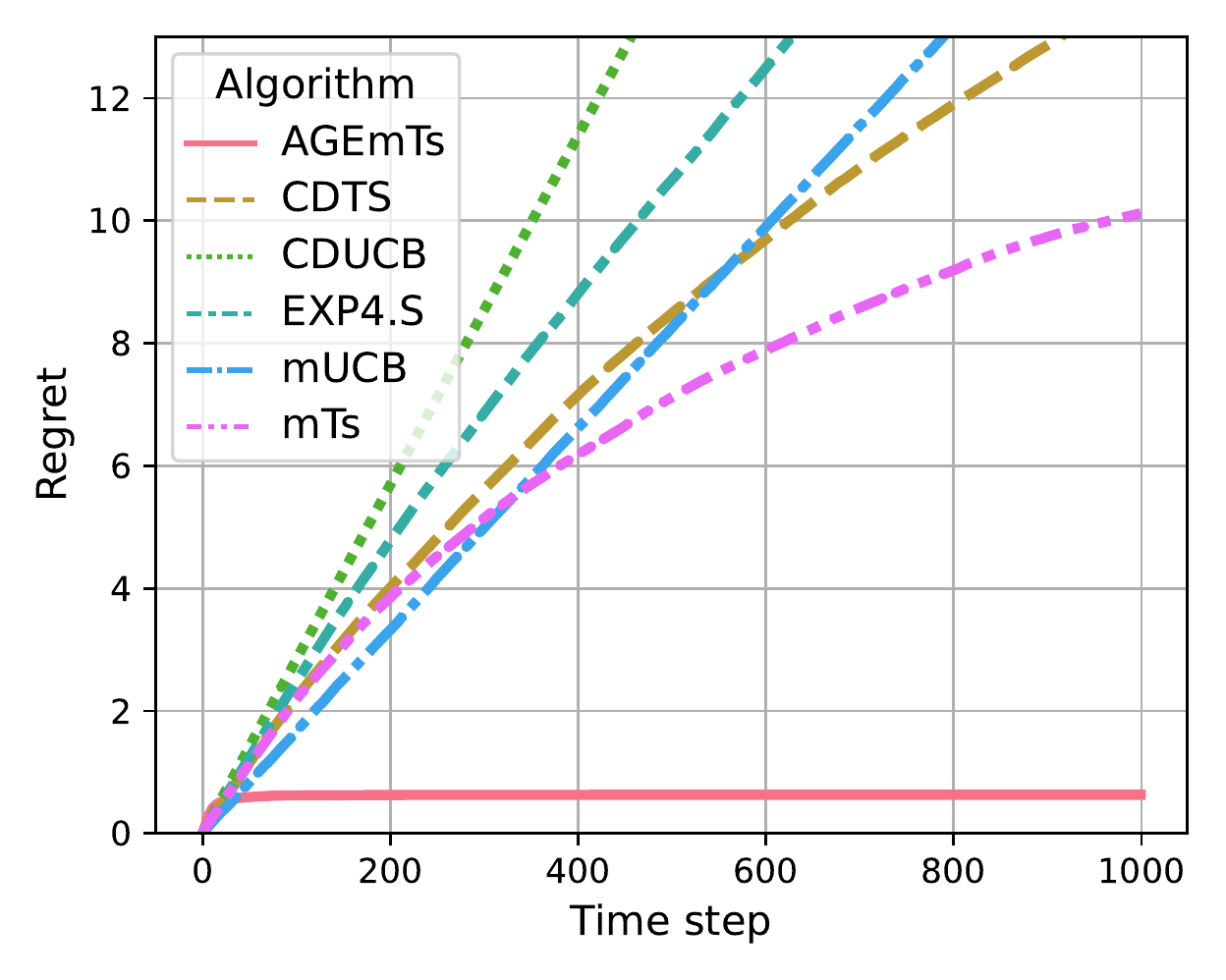}
         \caption{Regret curve for two-state stationary setting.}
         \label{fig:twostatestationaryB}
     \end{subfigure}
     \caption{Stationary two-state setting. The two states $s_1$ and $s_2$ are difficult to distinguish using the best arm in each state, but there exists an information-gathering arm that would not be chosen by state-of-the-art methods. A significant reward gain can be achieved over time.}
     \label{fig:stationary two state}
\end{figure}

For the means of the arm-rewards, we set
\begin{equation*}
     R(\cdot|a,s;\theta^*) = \begin{bmatrix}
    2.1 & 2.05\\
    2.05 & 2.1\\
    1.7 & 1.5 \\
    \end{bmatrix},
\end{equation*}

where rows are mean arm rewards and columns are states. We have for $\sigma = 0.5$ for all arms and states, except for the third arm (last row), where we set $\sigma=0.01$. The transition graph is shown in figure \ref{fig:twostatestationaryA}.

 Figure \ref{fig:twostatestationaryB} shows the regret in the stationary two-state setting. The information-gathering arm has a significantly lower reward. Thus it is never chosen by mTS, but has a KL-divergence of $50$ between the two states. AGEmTS uses this information-gathering arm early, as seen in the slightly higher regret in the first timesteps -- thus identifying the state quickly. It can then consistently choose the best arm compared to mTS. An agent using mTS will never choose the third arm, thus it takes a significantly longer time to identify the latent-state. Other state-of-the-art algorithms performed significantly worse than mTS and AGEmTS, even after parameter-tuning.  

Identifying the latent state early in this simple stationary setting results in a significant n-step regret advantage. If the latent state changed, we would expect more information-gathering arms to be required to maintain improved regret. To illustrate this, we investigate a simple non-stationary setting.   

\begin{figure}
     \centering
     \begin{subfigure}[b]{0.45\textwidth}
         \centering
         \includegraphics[width=\textwidth]{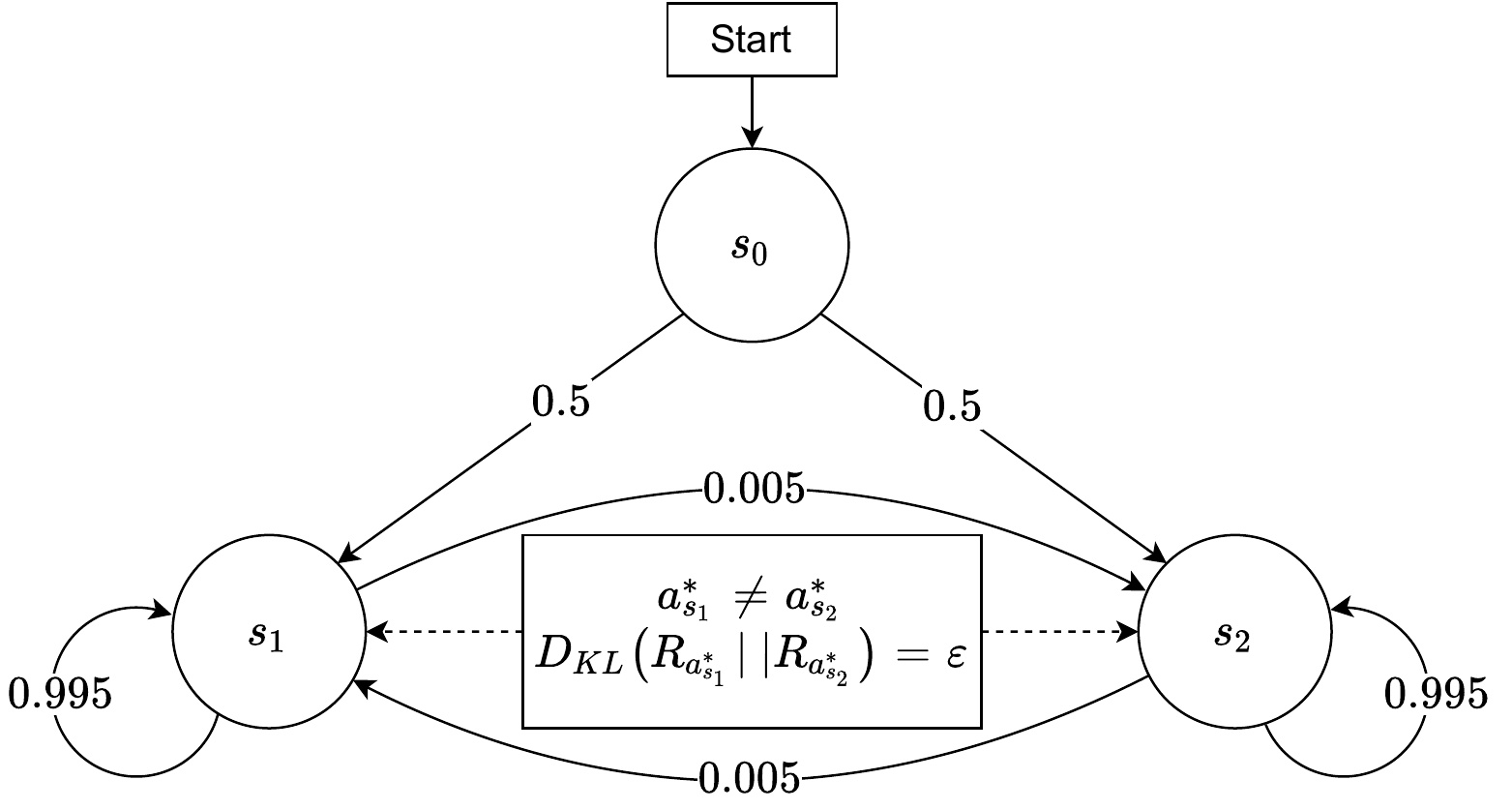}
         \caption{$\epsilon = 0.005$. }
         \label{fig:twostatenonstationaryA}
     \end{subfigure}
     \hfill
     \begin{subfigure}[b]{0.45\textwidth}
         \centering
         \includegraphics[width=\textwidth]{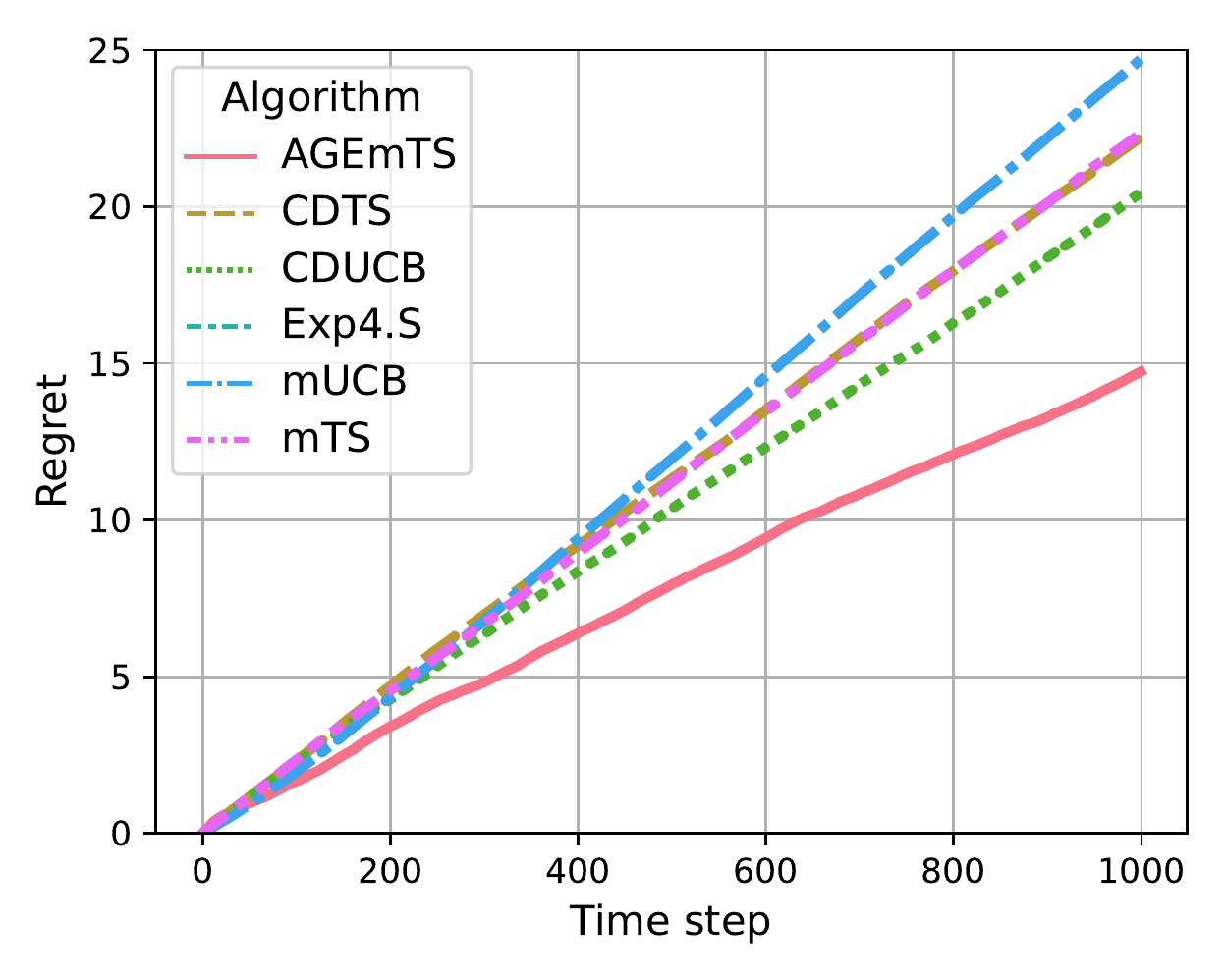}
         \caption{}
         \label{fig:twostatenonstationaryB}
     \end{subfigure}
     \label{fig:nonstationary_example}
     \caption{Non-stationary setting with change-points sampled according to the transition matrix. (a) Transition graph of the two-state stationary problem. $s_0$ is the start state but switches immediately to $s_1$ or $s_2$ with equal probabilities. (b) Regret curve for random transitions. The state changes approximately every 200 time steps.}
\end{figure}

The non-stationary setting has the same reward distributions between states, but the latent state may switch between states $s_1$ and $s_2$ with low probability, about every $200$ time steps. For the transition matrix we set $P(s'|s;\phi^*) = 0.995$ if $s'=s$ and $0.05$ otherwise. The transition graph is shown in figure \ref{fig:twostatenonstationaryA}.

Figure \ref{fig:twostatenonstationaryB} shows the regret for the non-stationary setting, where AGEmTS outperforms all other algorithms. Note that sub-linear regret is not achievable in the non-stationary setting, except if the number of state changes is sub-linear in time horizon $n$. The n-step regret advantage is not as pronounced as in the stationary setting. This is explained by the foreshortened segments in-between state changes. That is, the knowledge gathered through informative arms about the latent state loses its validity as soon as the state changes. It would require additional information-gathering, with the timing of such being particularly important to maximize the gain.         

\begin{figure}
    \centering
    \includegraphics[width=0.5\textwidth]{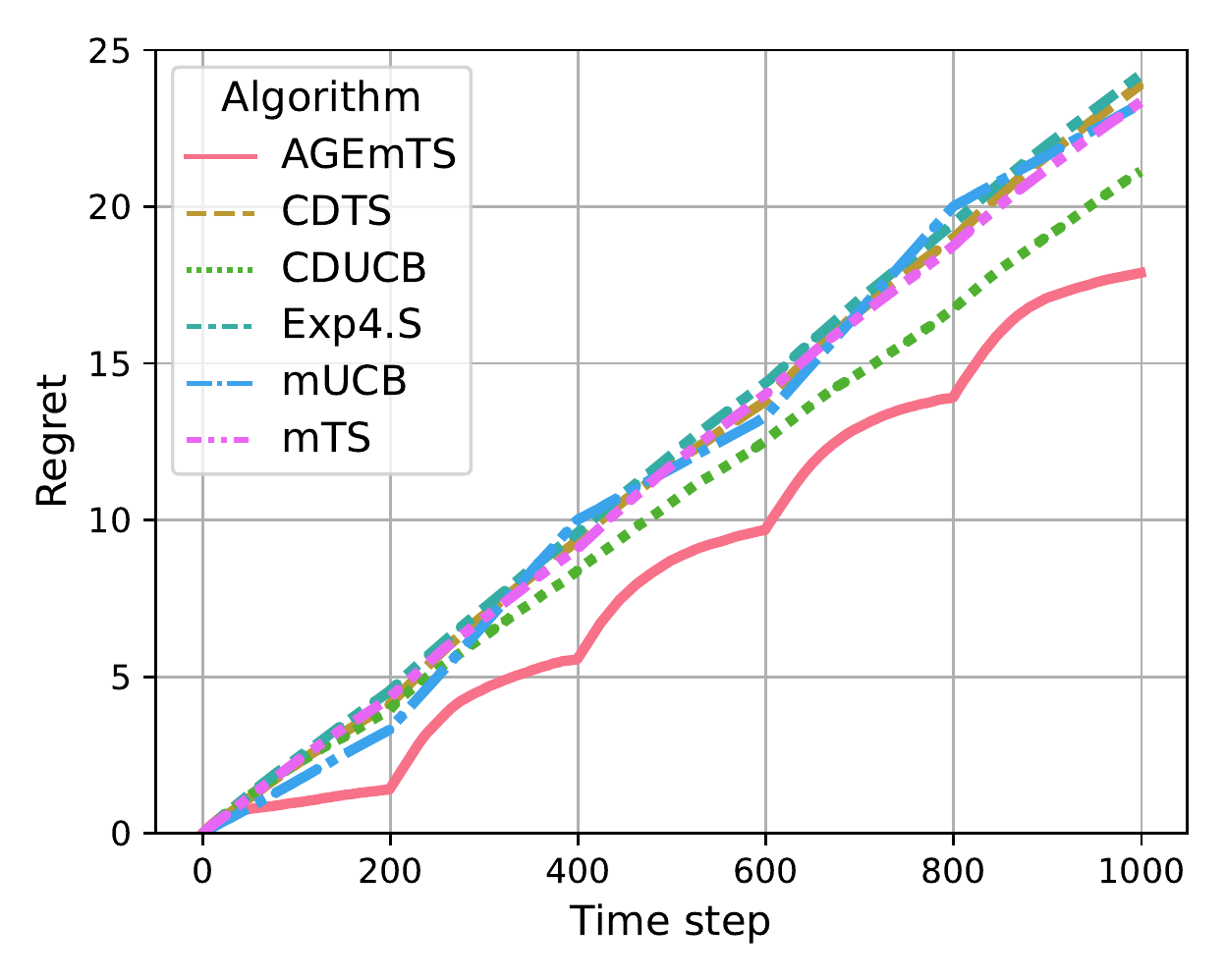}
    \caption{Non-stationary two-state setting. Regret curve for fixed transitions (every 200 time steps). Using the information-gathering arm, AGEmTS achieves sub-linear regret significantly faster than competing algorithms.}
    \label{fig:twostatenonstationaryC}
\end{figure}

To see the benefit of AGEmTS more clearly, and show the importance of correctly timing information-gathering arms, we investigate our algorithm's behavior for fixed change points. We set the interval between changes to a fixed $200$ time steps. The results are shown in figure \ref{fig:twostatenonstationaryC}. AGEmTS achieves sub-linear regret much faster than the other algorithms within each stationary interval.

We note that the highest state uncertainty at the start where the agent's belief-state is shared with an equal probability between state $s_1$ and $s_2$, which prompts the choice of an information-gathering arm very early, resulting in nearly zero additional regret in the first 200 time steps. As can be observed, after the first state change, the regret curve of AGEmTS is steeper than the first $200$ time steps, explained by the delay in information-gathering after a change occurred. This shows the importance of timing the information-gathering arm correctly to gain the most benefit in the stationary parts between changes. Our algorithm behaves conservatively, requiring significant state confusion and reward advantage before selecting an information-gathering arm. This behavior avoids incurring regret through information-gathering that might not be recuperable between too frequent state changes.

\subsection{Multi-State Non-stationary Settings}

\begin{figure}
     \centering
     \begin{subfigure}[b]{0.2\textwidth}
         \centering
         \includegraphics[width=\textwidth]{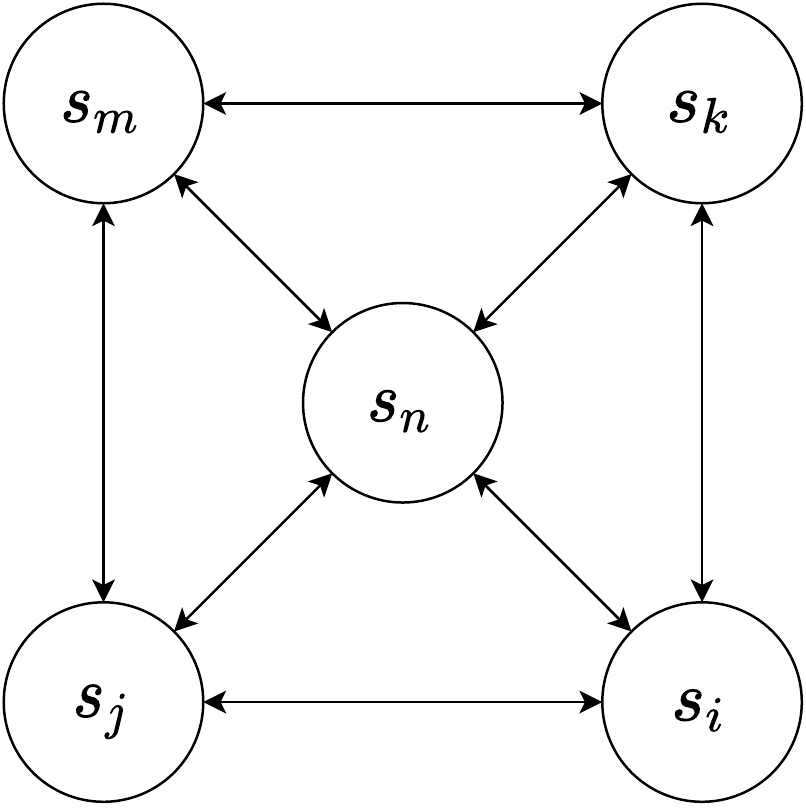}
         \caption{}
         \label{fig:simplefull}
     \end{subfigure}
     \hfill
     \begin{subfigure}[b]{0.2\textwidth}
         \centering
         \includegraphics[width=\textwidth]{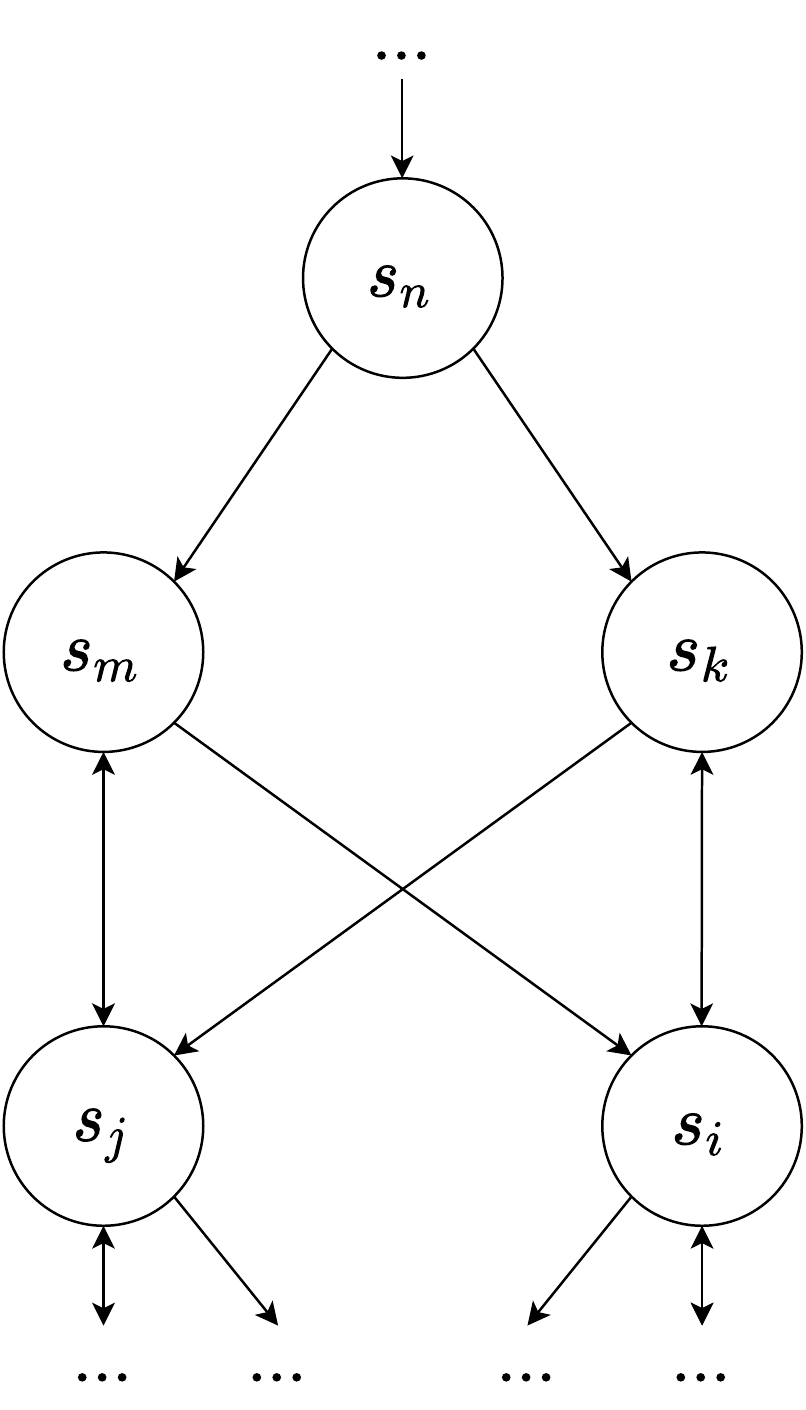}
         \caption{}
         \label{fig:simpleskip}
     \end{subfigure}
     \hfill
     \begin{subfigure}[b]{0.2\textwidth}
         \centering
         \includegraphics[width=\textwidth]{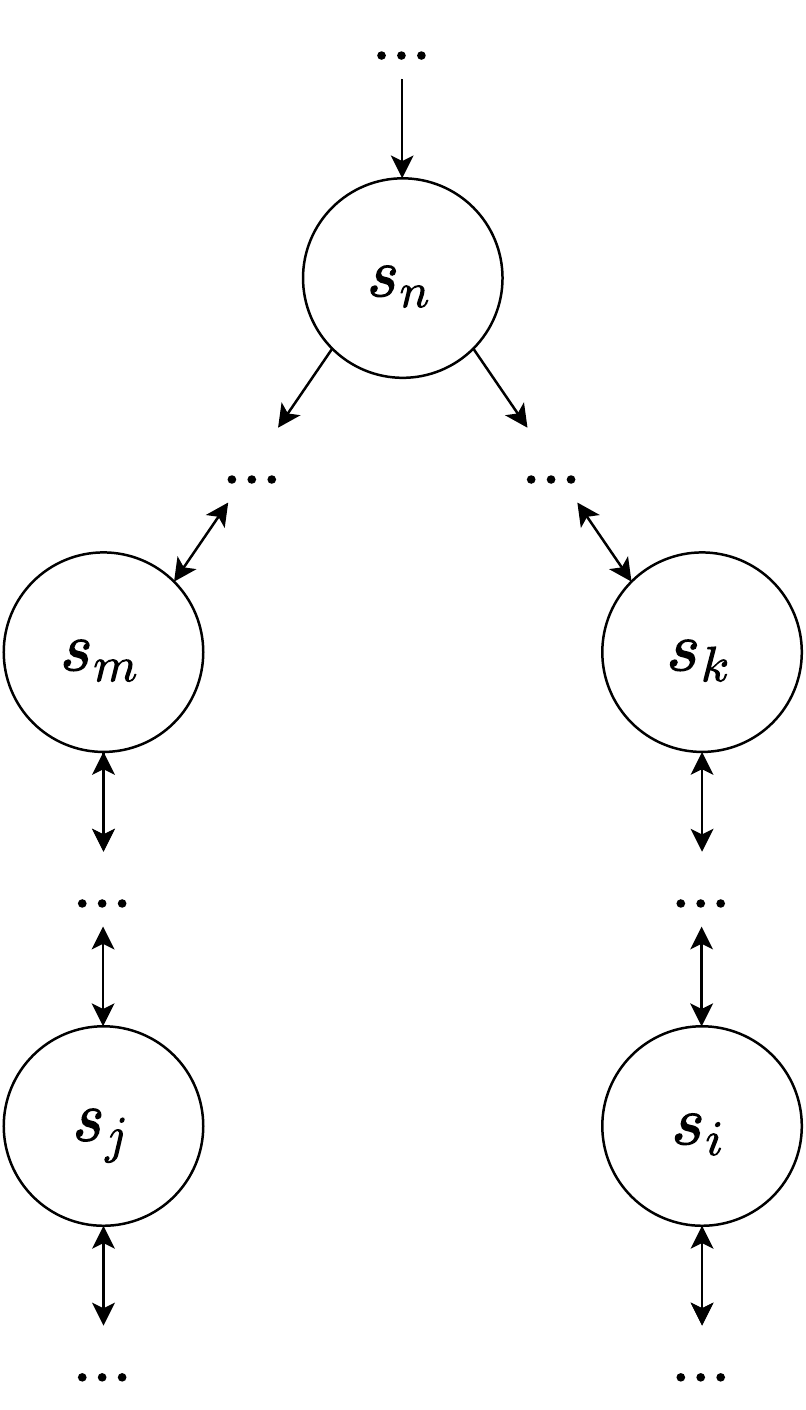}
         \caption{}
         \label{fig:simpletree}
     \end{subfigure}
        \caption{Investigated transition graphs. Irreducible (a) and reducible (b) and (c) Markov chains. (a) Fully connected graph, typically assumed in prior work, (b) tree graph with skip connections, as an intermediate between fully-connected and chain-structure and (c) chain-graph with two branches.}
        \label{fig:three graphs}
\end{figure}

In order to analyze state interactions in different settings, we generate several graph structures for the Markov chains. We investigate one fully connected irreducible Markov chain, a prior assumption for real-world applications used in prior works. Furthermore, we investigate two reducible Markov chains. The interactions between the graph structure and arm reward distributions determine how beneficial an information-gathering arm is in disentangling the agent's immediate and future believe-state. The schematic graph structures are shown in figure \ref{fig:three graphs}.

For the experiments, we set $|\mathcal{S}| = 5$ and $|\mathcal{A}| = 5$ with high similarity (low $D_{KL}$) between the best arms of states $s_1$ and $s_3$, as well as between best arms of states $s_2$ and $s_4$. Thus, it is quite difficult to distinguish between those states using the best arms \textit{only}. An information-gathering arm always has the lowest reward for all states but allows state differentiation with only a few samples. This reward structure is particularly interesting for the irreducible Markov chain in example \ref{fig:simpletree}, where sampling an information-gathering arm once can have significant benefit far into the future. The mean rewards for all structures is defined as

\begin{equation*}
     R(\cdot|A,S;\theta^*) = \begin{bmatrix}
    2.1 & 2.05 & 1.40 & 1.45 & 1.0\\
    2.05 & 2.1 & 1.45 & 1.40 & 0.95\\
    2.0 & 1.9 & 1.50 & 1.55 & 1.05\\
    2.05 & 2.1 & 1.55 & 1.50 & 1.1\\
    1.0 & 0.9 & 0.8 & 0.7 & 0.6\\
    \end{bmatrix},
\end{equation*}

\begin{figure}
     \centering
     \begin{subfigure}[b]{0.3\textwidth}
         \centering
         \includegraphics[width=\textwidth]{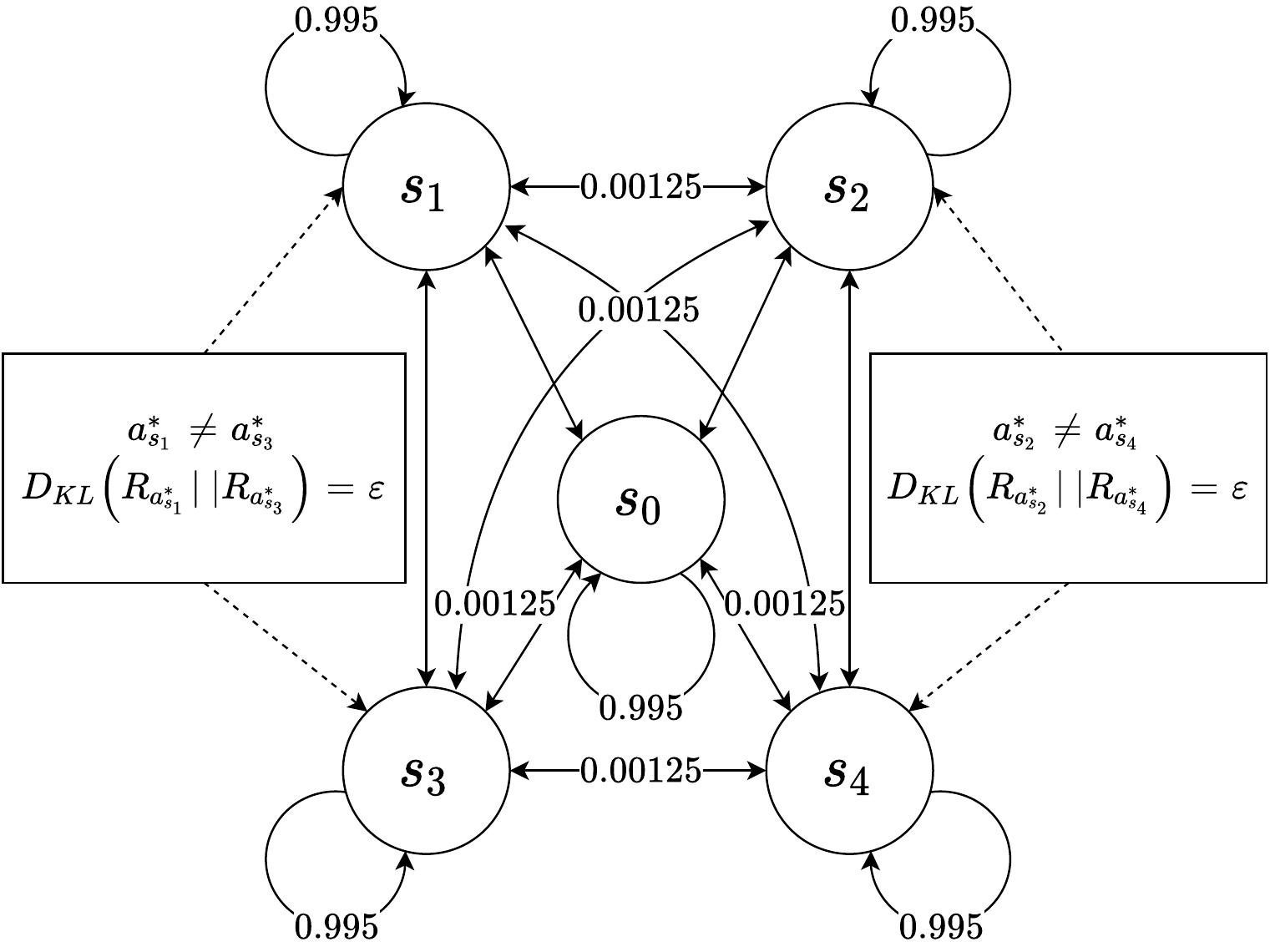}
         \caption{Fully connected graph.}
         \label{fig:multistate_synth_examples_regretA_b}
     \end{subfigure}
     \hfill
     \begin{subfigure}[b]{0.3\textwidth}
         \centering
         \includegraphics[width=\textwidth]{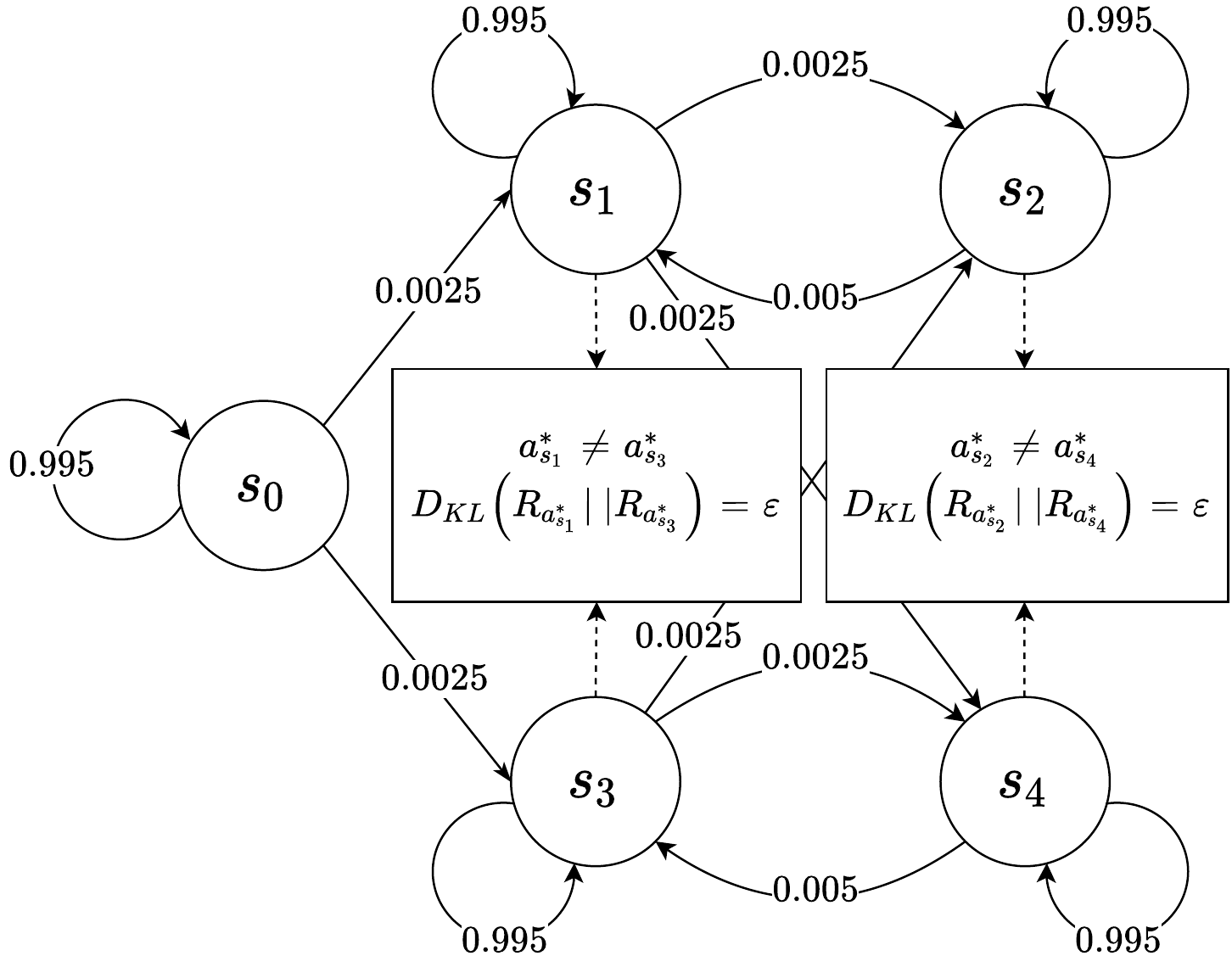}
         \caption{Skip connections between $s_1$ and $s_4$, as well as $s_3$ and $s_2$.}
         \label{fig:multistate_synth_examples_regretB_b}
     \end{subfigure}
     \hfill
     \begin{subfigure}[b]{0.3\textwidth}
         \centering
         \includegraphics[width=\textwidth]{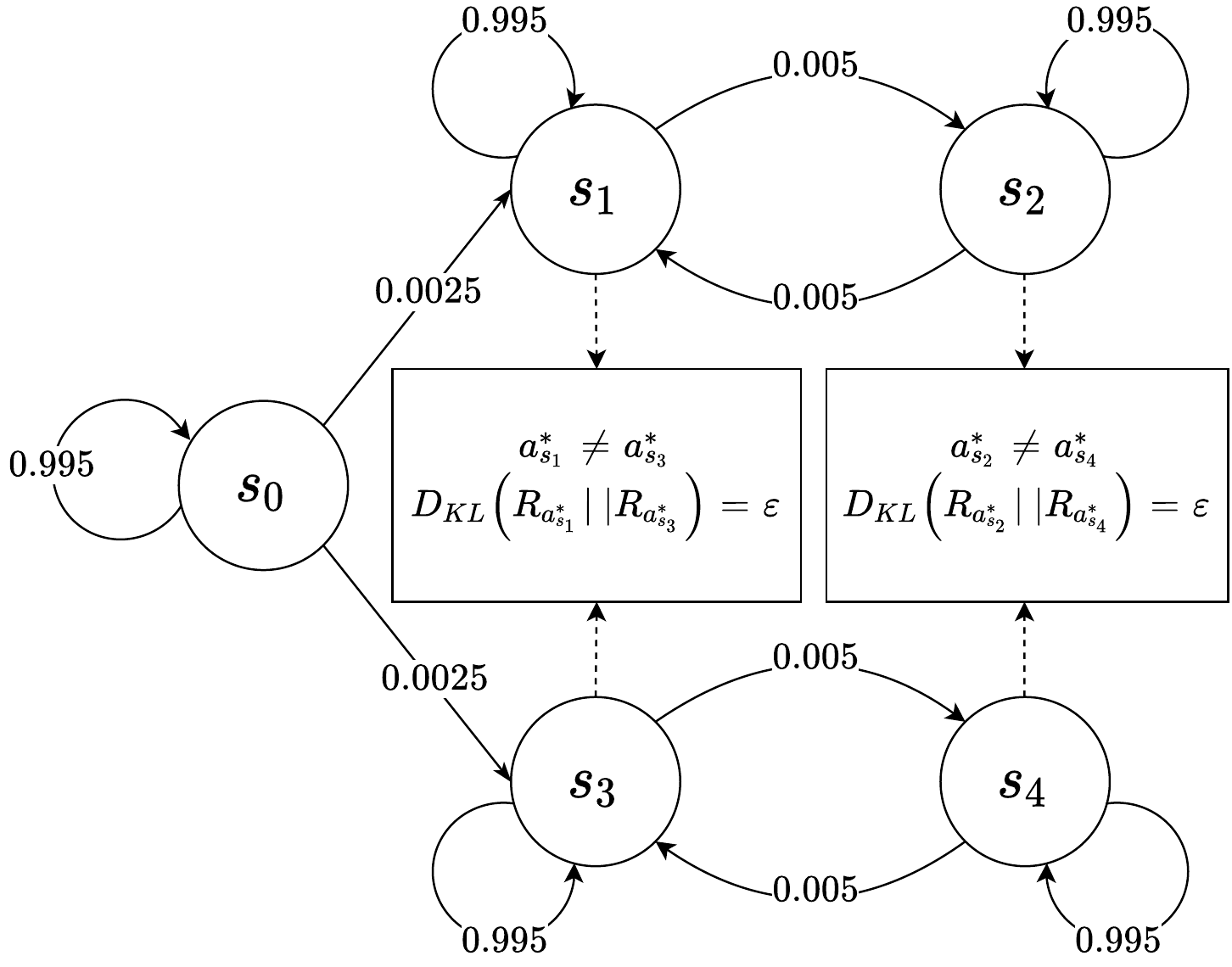}
         \caption{Graph with two branches.}
         \label{fig:multistate_synth_examples_regretC_b}
     \end{subfigure}
     \quad
     \begin{subfigure}[b]{0.3\textwidth}
         \centering
         \includegraphics[width=\textwidth]{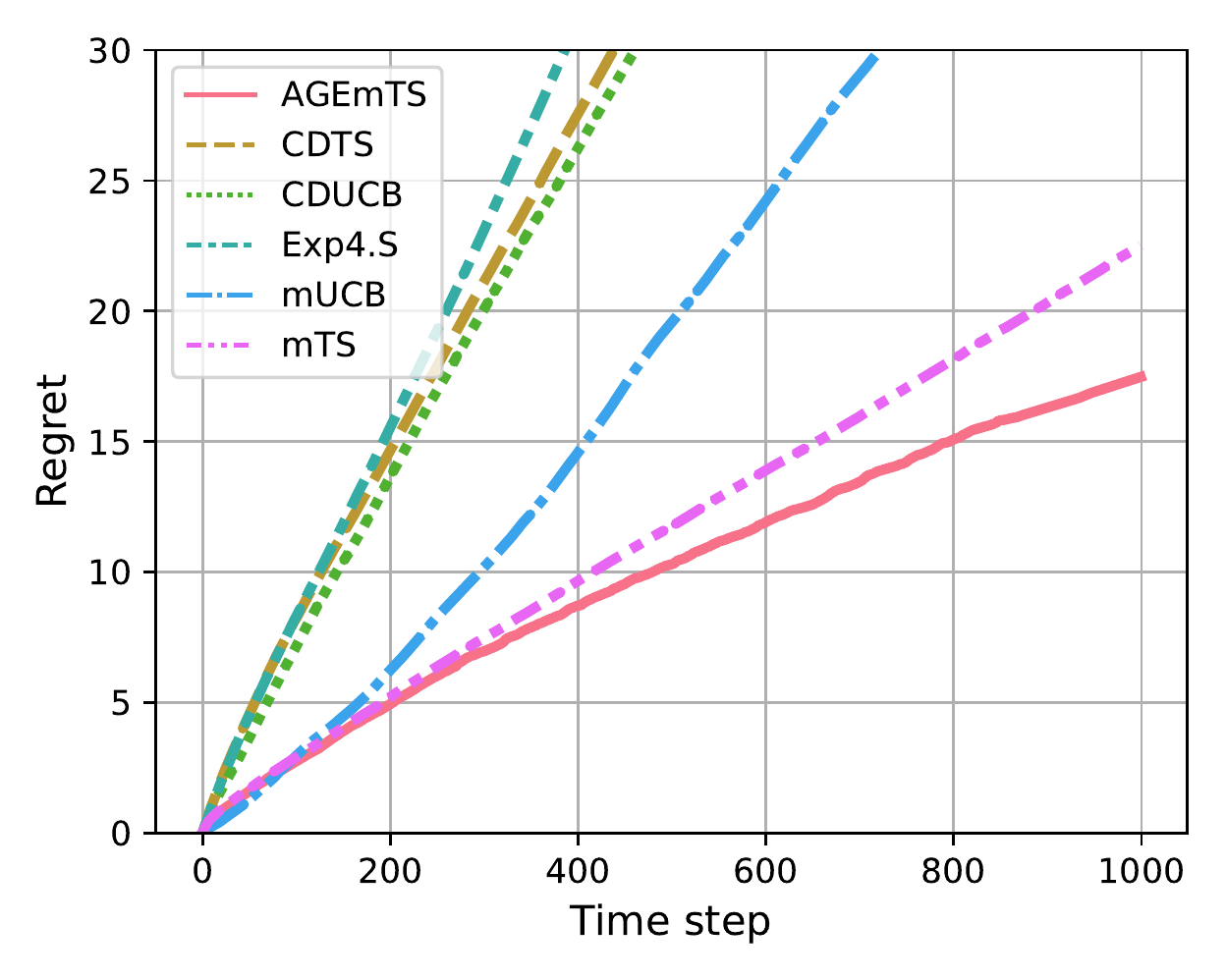}
         \caption{Fully connected graph.}
         \label{fig:multistate_synth_examples_regretD}
     \end{subfigure}
     \hfill
     \begin{subfigure}[b]{0.3\textwidth}
         \centering
         \includegraphics[width=\textwidth]{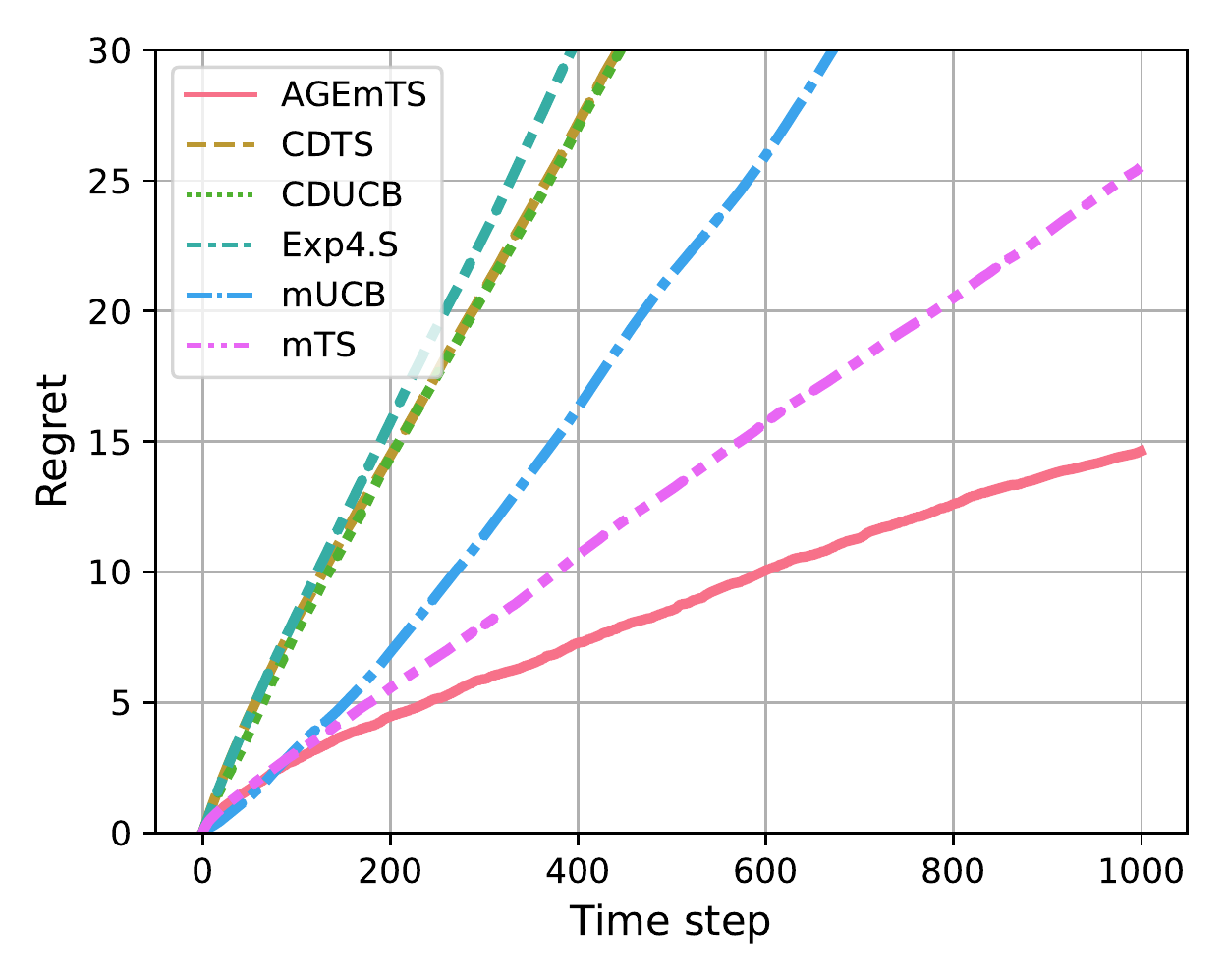}
         \caption{Chain graph with skip connections.}
         \label{fig:multistate_synth_examples_regretE}
     \end{subfigure}
     \hfill
     \begin{subfigure}[b]{0.3\textwidth}
         \centering
         \includegraphics[width=\textwidth]{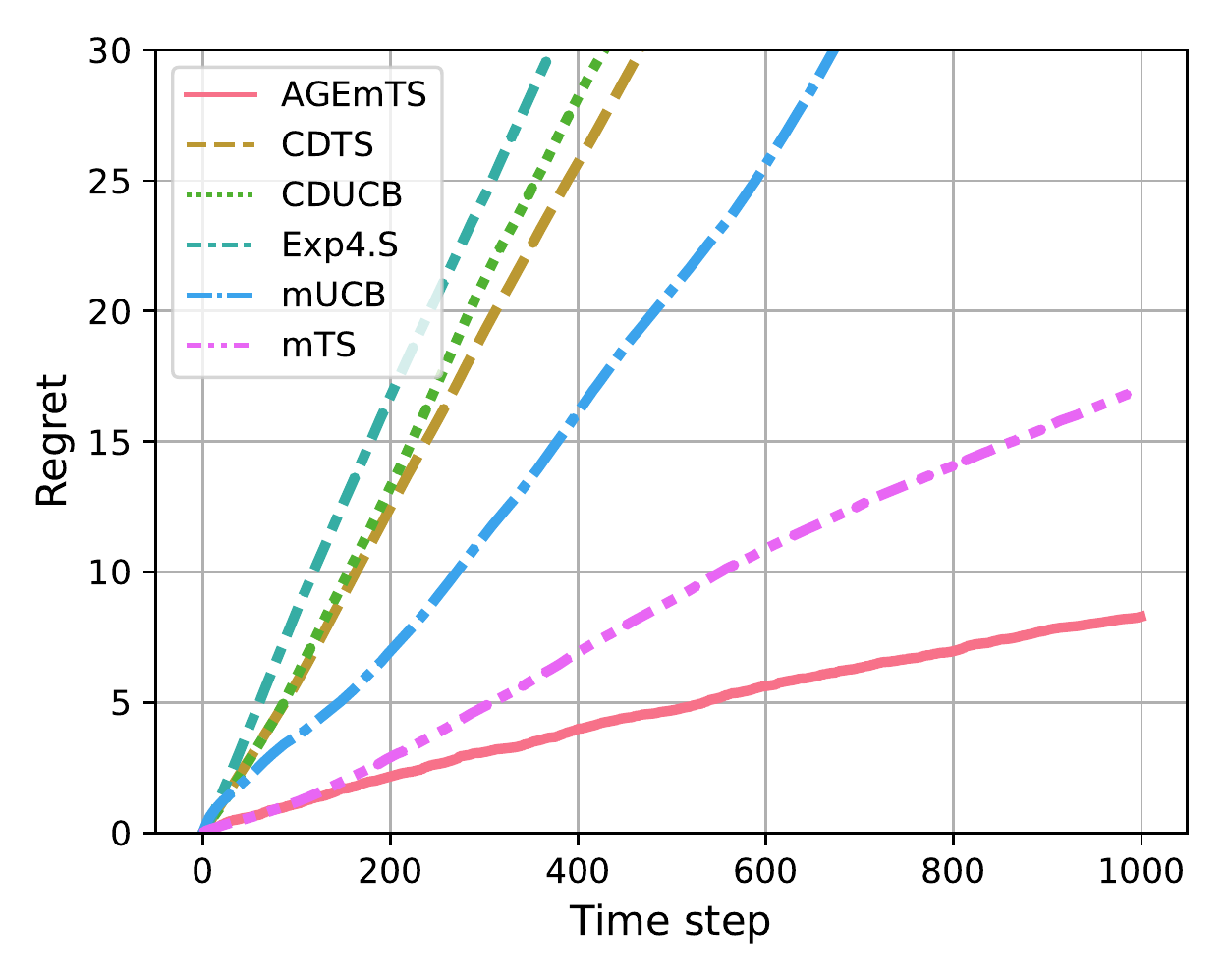}
         \caption{Chain graph with two branches.}
         \label{fig:multistate_synth_examples_regretF}
     \end{subfigure}
     \caption{Influence of transition matrix on the algorithms. Top row shows transition structures, while bottom row shows average regret over 100 runs. Information-gathering arms effectively reduce the subset of arms to consider in future states, particularly for graphs (b) and (c), where AGEmTS immediately discards one branch from consideration. 
     }
     \label{fig:multistate_synth_examples_regret}
\end{figure}

where rows are arms, and columns are states. The transition graphs and cumulative regret for multi-state non-stationary setting are shown in Figure \ref{fig:multistate_synth_examples_regret}. The optimal arms are different between the $s_1, s_3$ and $s_2, s_4$ branches, as seen from the mean rewards.  

Using the highest reward arms according to the belief-state alone, the agent has difficulty identifying the current state, leading to a close to uniform belief-state for mTS. For the reducible Markov chains (figure \ref{fig:multistate_synth_examples_regretB_b} and \ref{fig:multistate_synth_examples_regretC_b}), this leads to significantly worse regret compared to our algorithm, which ``cleverly'' selects information-gathering arms. mTS will continue to sample arms that result in low reward in one branch but in higher reward in the other. On the other hand, AGEmTS can discount sampling arms from one of the branches by identifying the latent state. This is particularly apparent in figure \ref{fig:multistate_synth_examples_regretF}, where AGEmTS regret curve is significantly less steep compared to mTS in the first 600 time steps. Eventually, mTS will identify the current state sufficiently well and focus on one of the branches resulting in both algorithms achieving similar regret over time after about 600 time steps. 

We see a less significant difference in regret for the fully-connect graph (figure \ref{fig:multistate_synth_examples_regretA_b}), yet AGEmTS still outperforms all other algorithms. In the fully connected structures, long-term effects of state uncertainty are not as pronounced as in the chain type graphs, where branches may reduce the number of arms to consider. Similar to the stationary experiments, the benefit of information-gathering in fully connected structures is limited to the length of the stationary segments. 

The benefit of information-gathering is not limited to uniform transition probabilities between states. To illustrate this, we ran several additional experiments. For each transition graph, we computed the regret as an average of over $100$ different instantiations of the transition matrix. States transition to themselves with a probability of $99.5\%$ (about every 200 time steps) as before, but the probabilities of transition to other states are non-uniform. That is, we uniformly sample a transition probability $p_{s_i}$, such that $\sum_{\{s_i \neq s_j\} \in \mathcal{S}} p_{s_i} = 0.05$.

\begin{figure}
     \centering
     \begin{subfigure}[b]{0.3\textwidth}
         \centering
         \includegraphics[width=\textwidth]{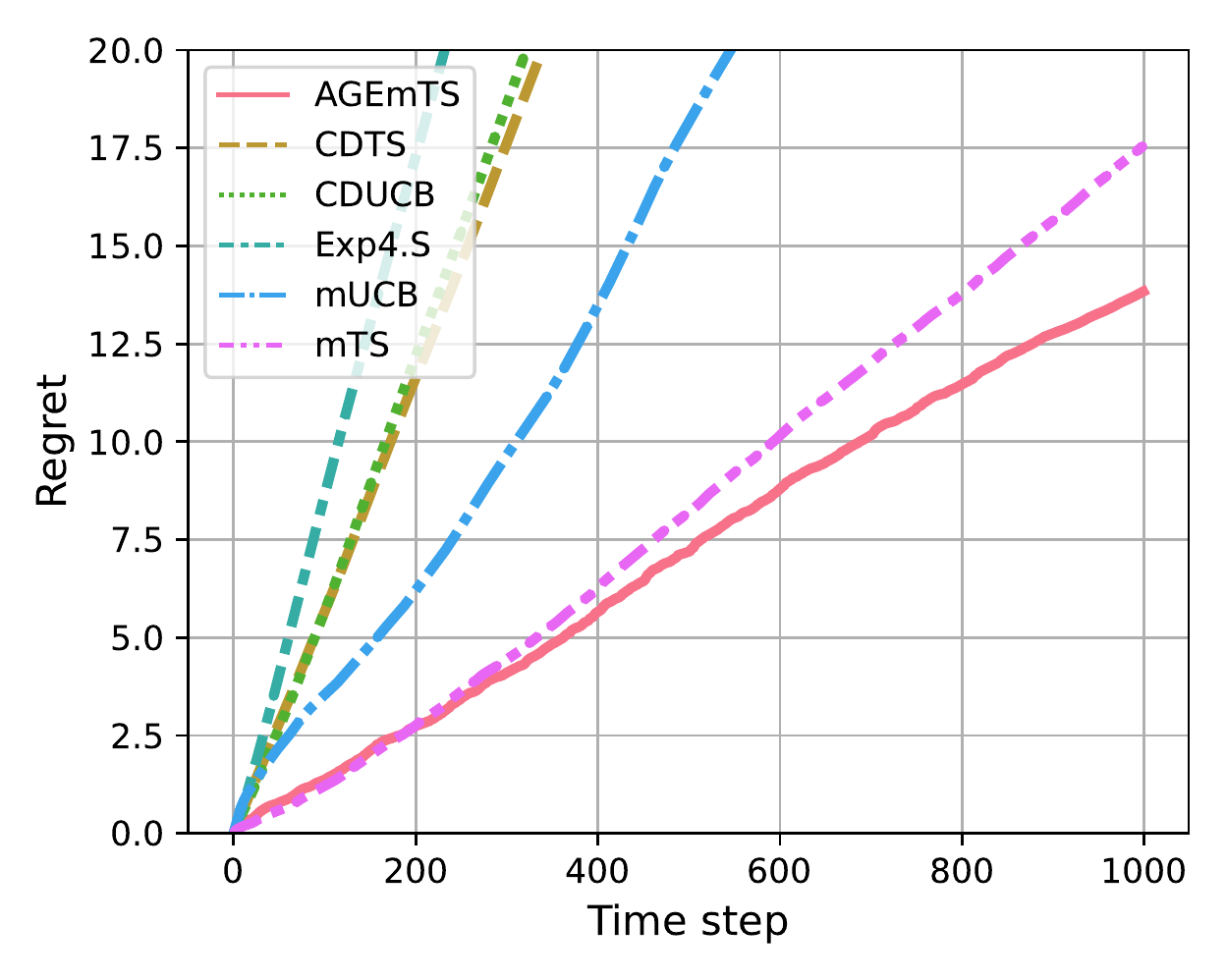}
         \caption{Fully connected graph.}
         \label{fig:y equals x}
     \end{subfigure}
     \hfill
     \begin{subfigure}[b]{0.3\textwidth}
         \centering
         \includegraphics[width=\textwidth]{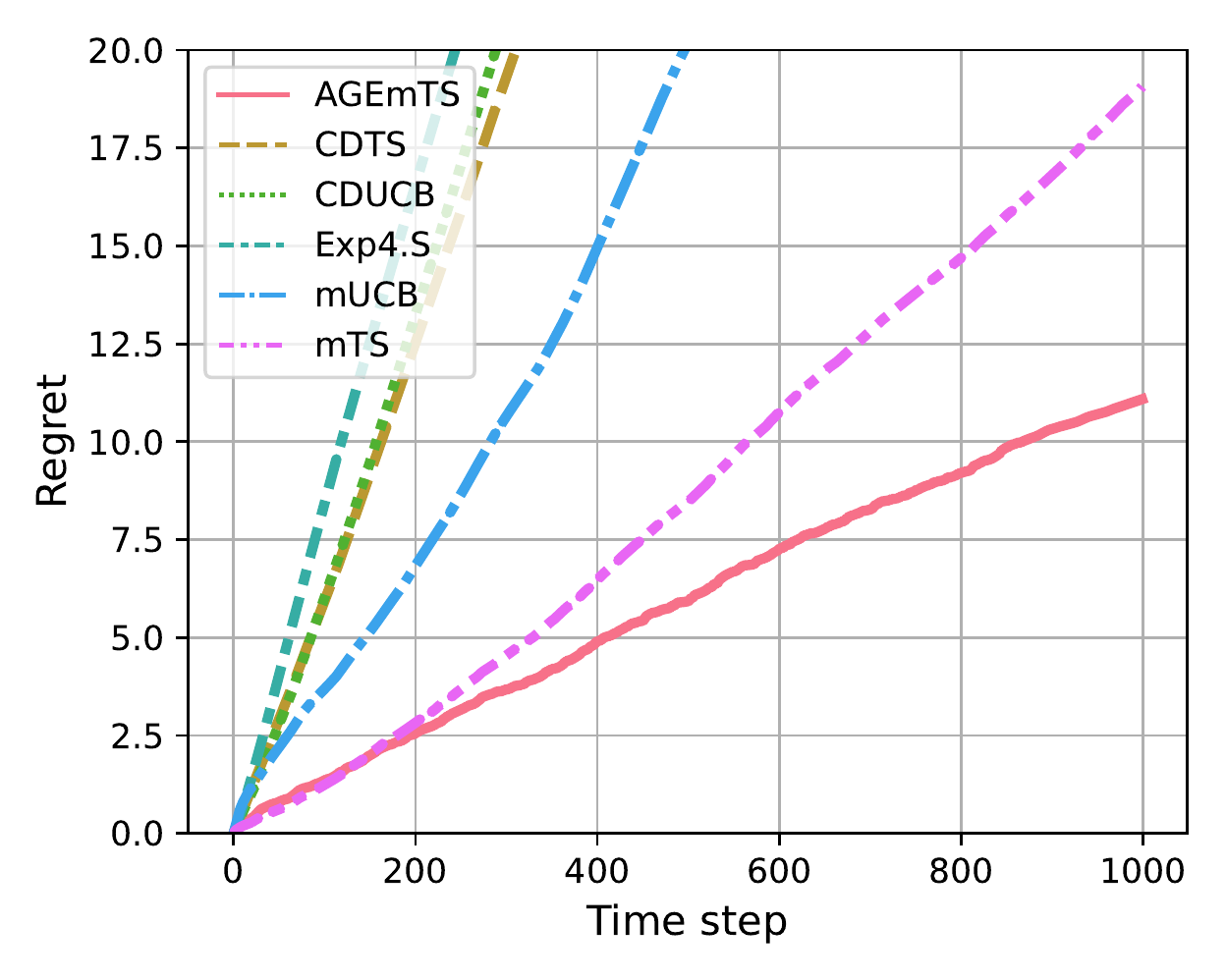}
         \caption{Chain graph with skip connections.}
         \label{fig:three sin x}
     \end{subfigure}
     \hfill
     \begin{subfigure}[b]{0.3\textwidth}
         \centering
         \includegraphics[width=\textwidth]{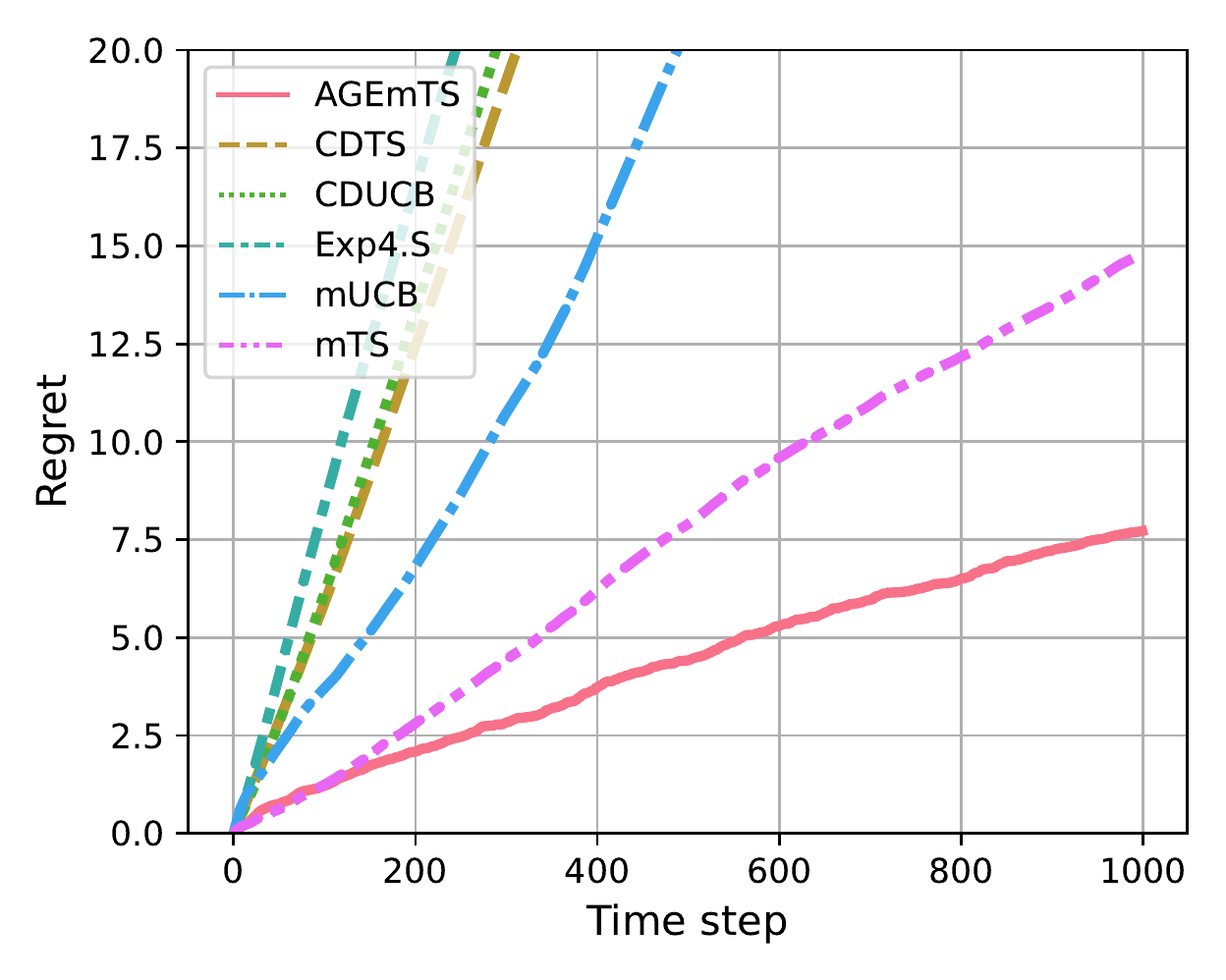}
         \caption{Chain graph with two branches.}
         \label{fig:five over x}
     \end{subfigure}
        \caption{Average regret (over 100 runs) for non-uniform transition probabilities. Regret patterns are similar to those seen in Figure \ref{fig:multistate_synth_examples_regret}.}
        \label{fig:avgoverinstances}
\end{figure}

Figure \ref{fig:avgoverinstances} shows the regret curves for mTS and for our algorithm. We observe a similar pattern as in the uniform transitions above, where the reward gain strongly depends on the interactions between the reward distribution and the transition matrix.

\subsection{Real-world Data Experiments on MovieLens 1M}\label{sec:movielens1m}

We follow a similar experimental setup for the MovieLens 1M data set as described in~\cite{Hong2020}, except that we also investigate different transition matrices. MovieLens is a popular collaborative filtering data set used to analyze and develop recommendation engines. The data set constitutes 6040 users that rated 3706 movies, totaling 3883 movies. Each movie is categorized into one of $301$ genres. We removed users from the data set who rated less than $200$ movies and removed movies with less than $200$ ratings. The final data set constitutes $1589$ users and $1132$ movies. The true reward distributions are provided to the algorithms. Missing ratings are imputed using probabilistic matrix factorization~\citep{Salakhutdinov2008} with the parameters: $\lambda_\alpha=0.001$, $\lambda_\beta=0.001$ and the size of the latent space of $10$. The gradient optimizer is run with a learning rate $2 \times 10^{-4}$. We reserved $10$\% of the data set for validation while training the factor matrices $U$ and $V$. We choose a latent size of $10$ since higher values yield no statistically significant difference in terms of validation error. We cluster the user vectors $U$ using k-Means clustering. Following \cite{Wu2018}, the reward distribution is sampled from a ``super-user''. The super-user constitutes a random sample of $|S|$ users, one from each cluster.

We define several non-stationary latent bandit instances with a fixed number of arms $|\mathcal{A}| = 20$ and fixed number of states $|\mathcal{S}|=5$. For the fully connected graph, the transition matrix is defined as $P(s'|s;\phi^*) = 1-p$ if $s'=s$ . $p$ is assigned to other transitions from $s$ according to a sum of $p_{s_i}$ sampled uniformly such that $\sum_{{s_i \neq s_j} \in \mathcal{S}} \cdot p_{s_i} = p$. The same procedure is done for the graphs with skip connections and two branches, but the weight for missing edges is always set to $0$. For the experiments, we set $p=0.05$. Thus, state changes occur every $200$ time steps on average. We run each bandit instance for $1000$ time steps.

A run of the latent bandit instance obeys the following protocol. First, a super-user $i_1, \dots, i_{|\mathcal{S}|}$ is sampled, such that similarity in reward distributions between states is ensured according to figure \ref{fig:multistate_synth_examples_regret}(a-c). Specifically, user $i_1$ ($s_1$) and $i_3$ ($s_3$), as well as, $i_2$ ($s_2$) and $i_4$ ($s_4$) are nearest neighbors, according to $||U_i - U_j||^2_2$. The super-user stays fixed for all time steps. The initial latent state is set to $s_0$. For each time step, the next latent state $S_{t+1}$ is sampled according to the current latent state $S_t$ and the transition matrix. The movie set is sampled uniformly at random from $\mathcal{A}$. The agent is provided with the sampled movies, specifically, the context $X_t \in \mathbb{R}^{|A|\times 10}$ is provided to the agent. The context contains the training set vectors in $V$ of the sampled movies. The mean reward is computed as the dot product between the user vector $U_{i_s}$, and the chosen movie vector $V_j$. The reward is then drawn from a normal distribution with fixed standard deviation $\sigma=0.25$, $R \sim \mathcal{N}(U^\top_{i_s}V_j, 0.25^2)$. $U$ and $V$ are not provided to the agent.

\begin{figure}
     \centering
     \begin{subfigure}[b]{0.3\textwidth}
         \centering
         \includegraphics[width=\textwidth]{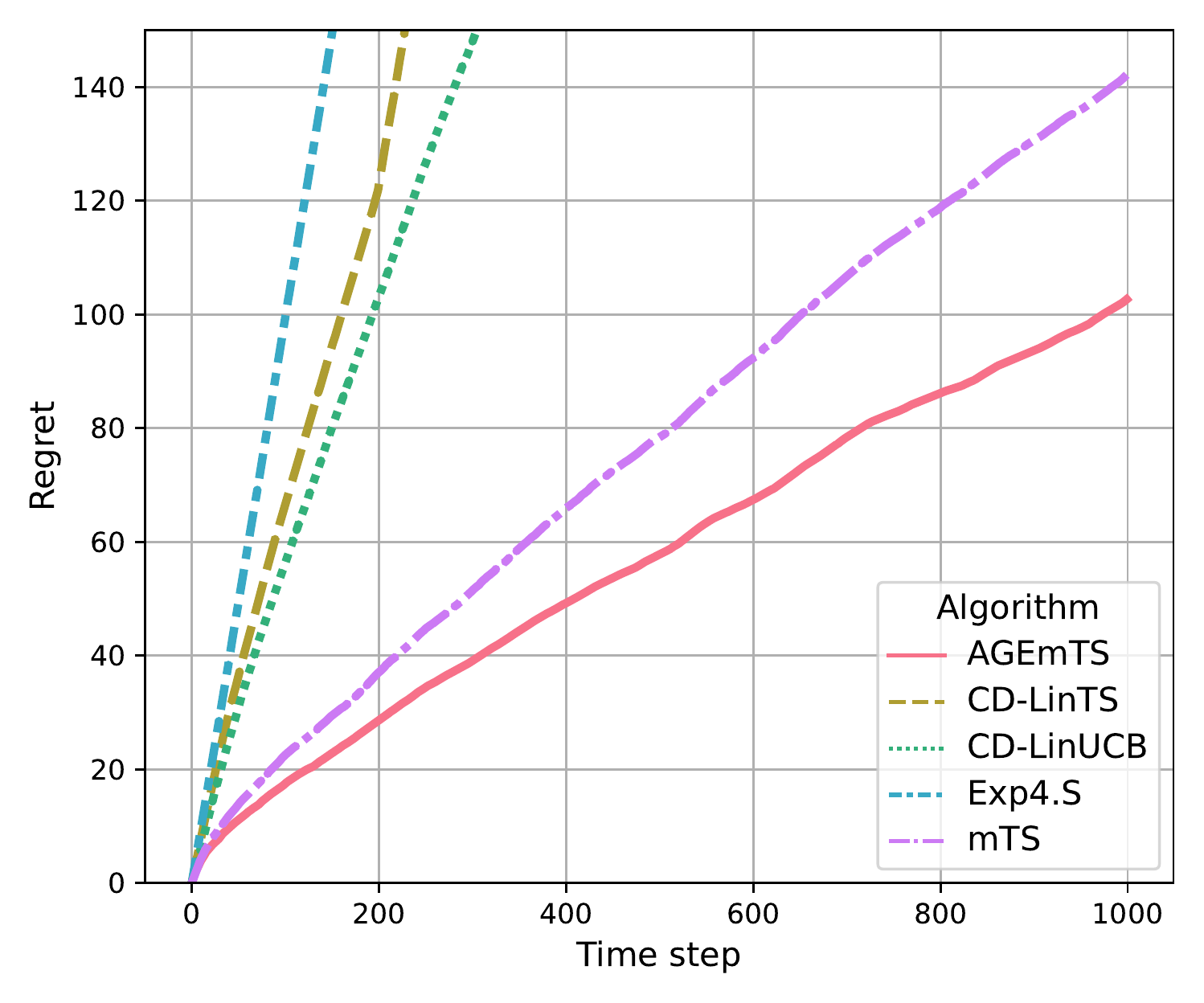}
         \caption{Fully connected graph.}
         \label{fig:y equals x}
     \end{subfigure}
     \hfill
     \begin{subfigure}[b]{0.3\textwidth}
         \centering
         \includegraphics[width=\textwidth]{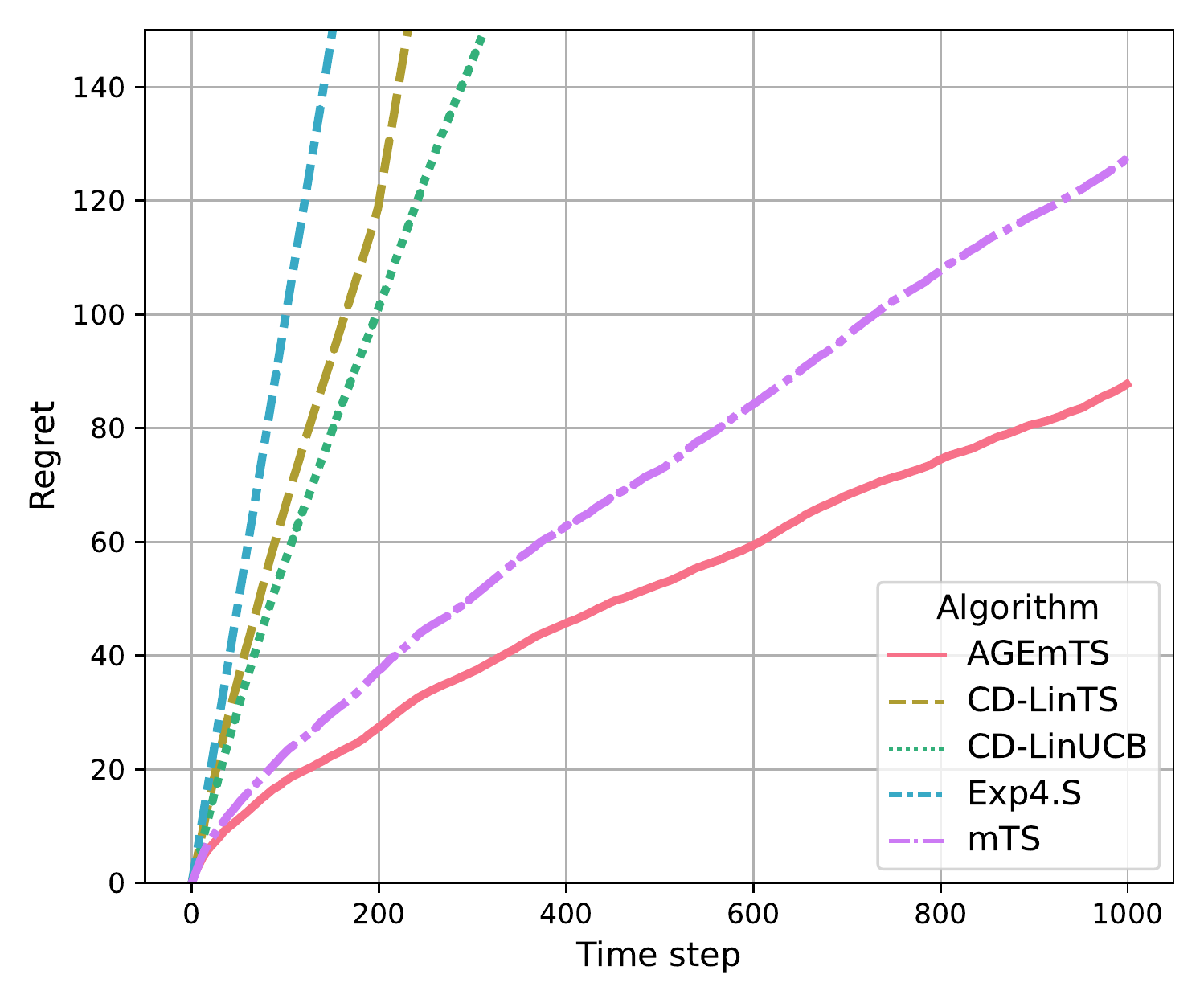}
         \caption{Chain graph with skip connections.}
         \label{fig:three sin x}
     \end{subfigure}
     \hfill
     \begin{subfigure}[b]{0.3\textwidth}
         \centering
         \includegraphics[width=\textwidth]{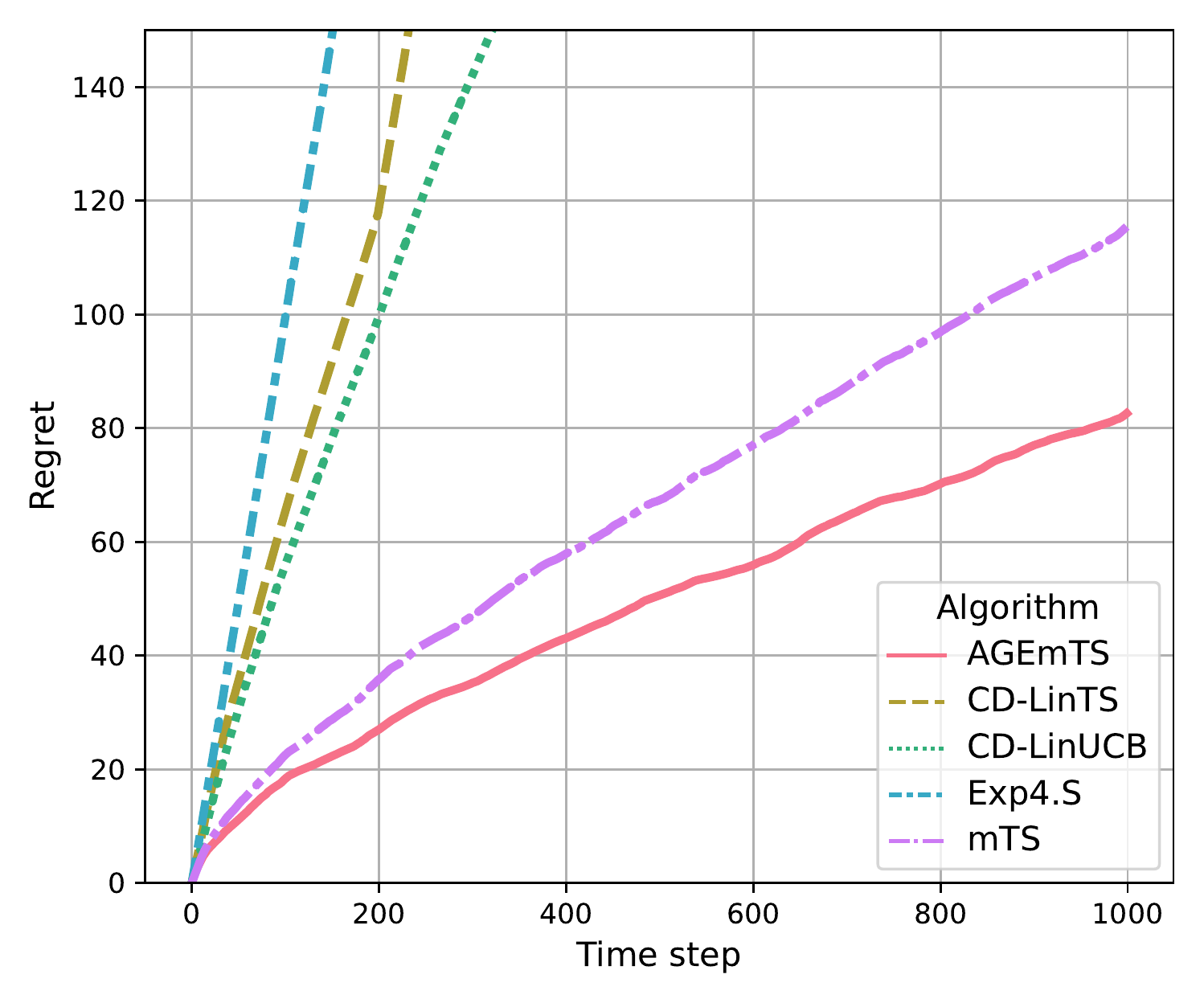}
         \caption{Chain graph with two branches.}
         \label{fig:five over x}
     \end{subfigure}
        \caption{Regret curves for the different transition matrices using the MovieLens 1M data set.}
        \label{fig:RandomMovie}
        
\end{figure}

The results in terms of regret are shown in Figure \ref{fig:RandomMovie}. In all cases, our method performs significantly better than mTs and other algorithms for various transition matrices. While there exists a difference between the transition graphs, it is not as pronounced as in the synthetic experiments. We can explain this by the fact that for fixed $\sigma=0.25$, the reward distributions in MovieLens 1M are relatively distinctive, such that the window is shortened where improved state identification can lead to a better reward. Nevertheless, we observe a significant difference in our synthetic experiments (figure \ref{fig:multistate_synth_examples_regretA_b}) for the fully connected graph, where reward improvements are already significant in the early stages of runs. This is somewhat expected since information-gathering arms in MovieLens 1M generally do not come with significantly worse rewards than the best arms in each state. Thus, information-gathering is not as heavily penalized, resulting in more reward gain in the early stages of interaction compared to the synthetic experiments. To illustrate where most benefits of information-gathering arms can be obtained, we run several additional experiments in the following section.

\subsection{Regions of Benefit}

We run several experiments on a two-state scenario to show the conditions under which the information-gathering arms actively used by AGEmTs achieve reward improvement over mTS. Experiments are carried out for stationary and non-stationary environments using synthetic and MovieLens data. We focus on fully connected transitions graphs with $P(s'|s, \phi^*) = p$ if $s' = s$ and $(1-p)$ otherwise, for $p=0.995$.

\begin{figure}
     \centering
     \begin{subfigure}[b]{0.3\textwidth}
         \centering
         \includegraphics[width=\textwidth]{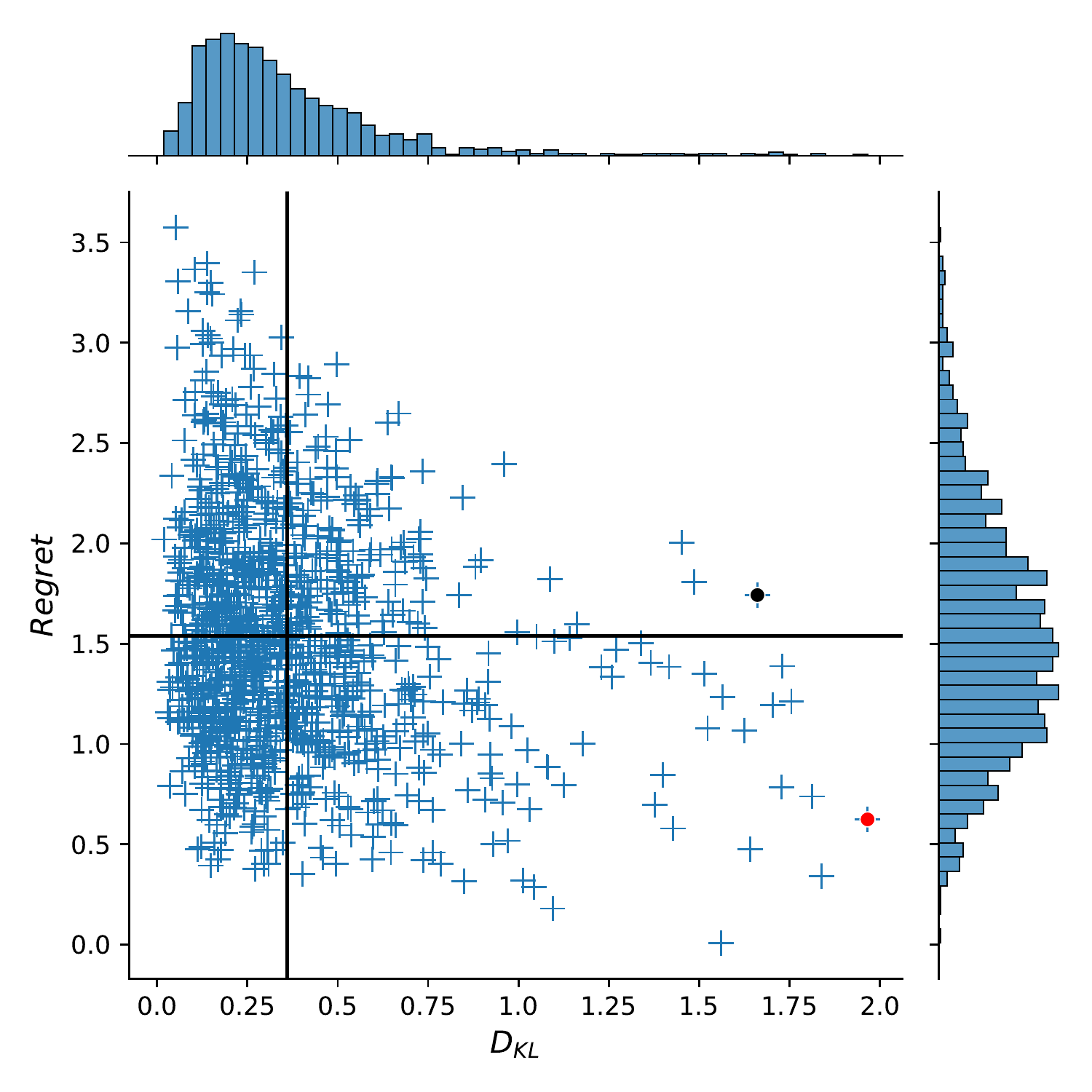}
         \caption{The same fixed $\sigma = 0.25$ for all movies.}
         \label{fig:data_sigma_fixed}
     \end{subfigure}
     \hfill
     \begin{subfigure}[b]{0.3\textwidth}
         \centering
         \includegraphics[width=\textwidth]{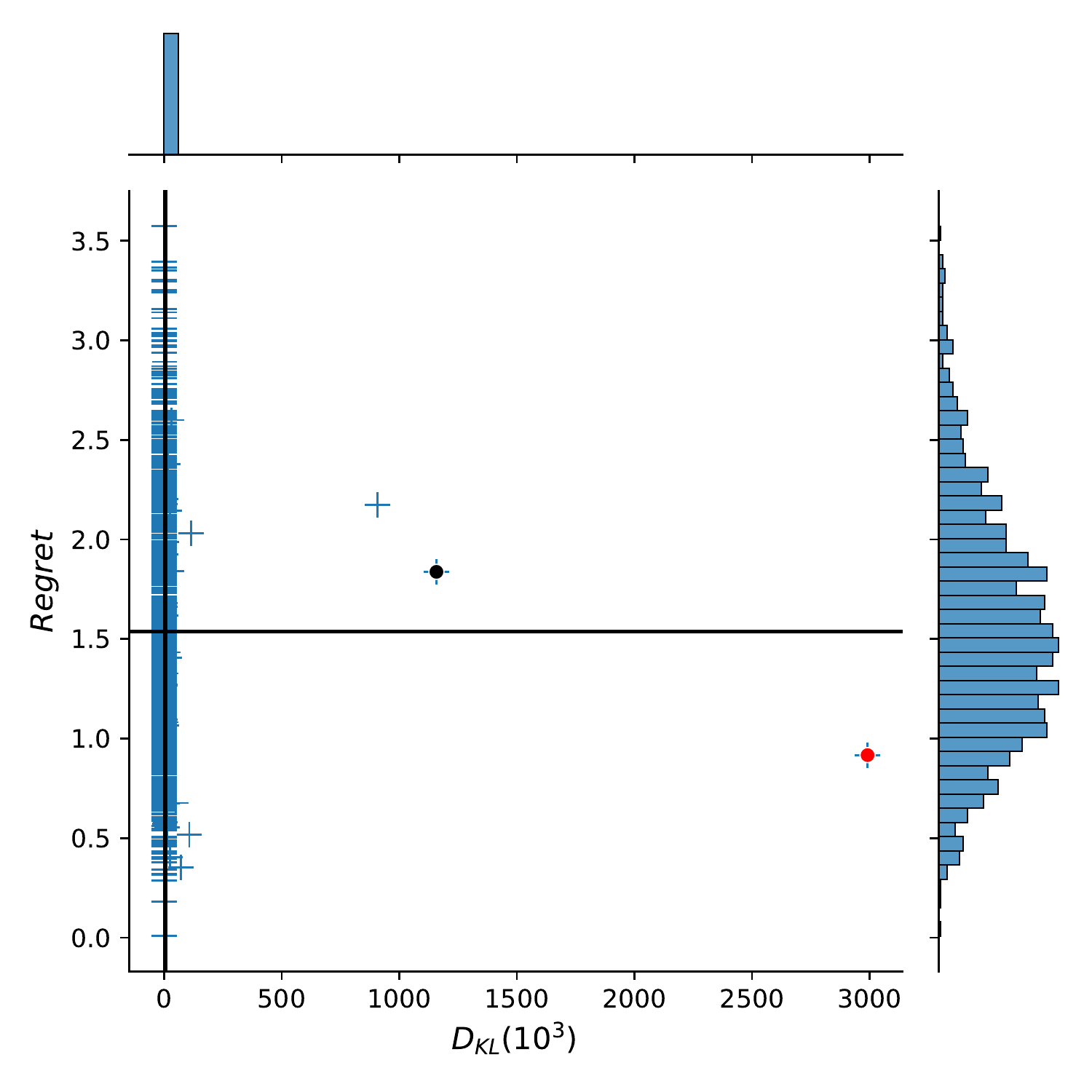}
         \caption{$\sigma$ from ratings of the three nearest neighbors (movies) for each user.}
         \label{fig:data_nearest}
     \end{subfigure}
     \hfill
     \begin{subfigure}[b]{0.3\textwidth}
         \centering
         \includegraphics[width=\textwidth]{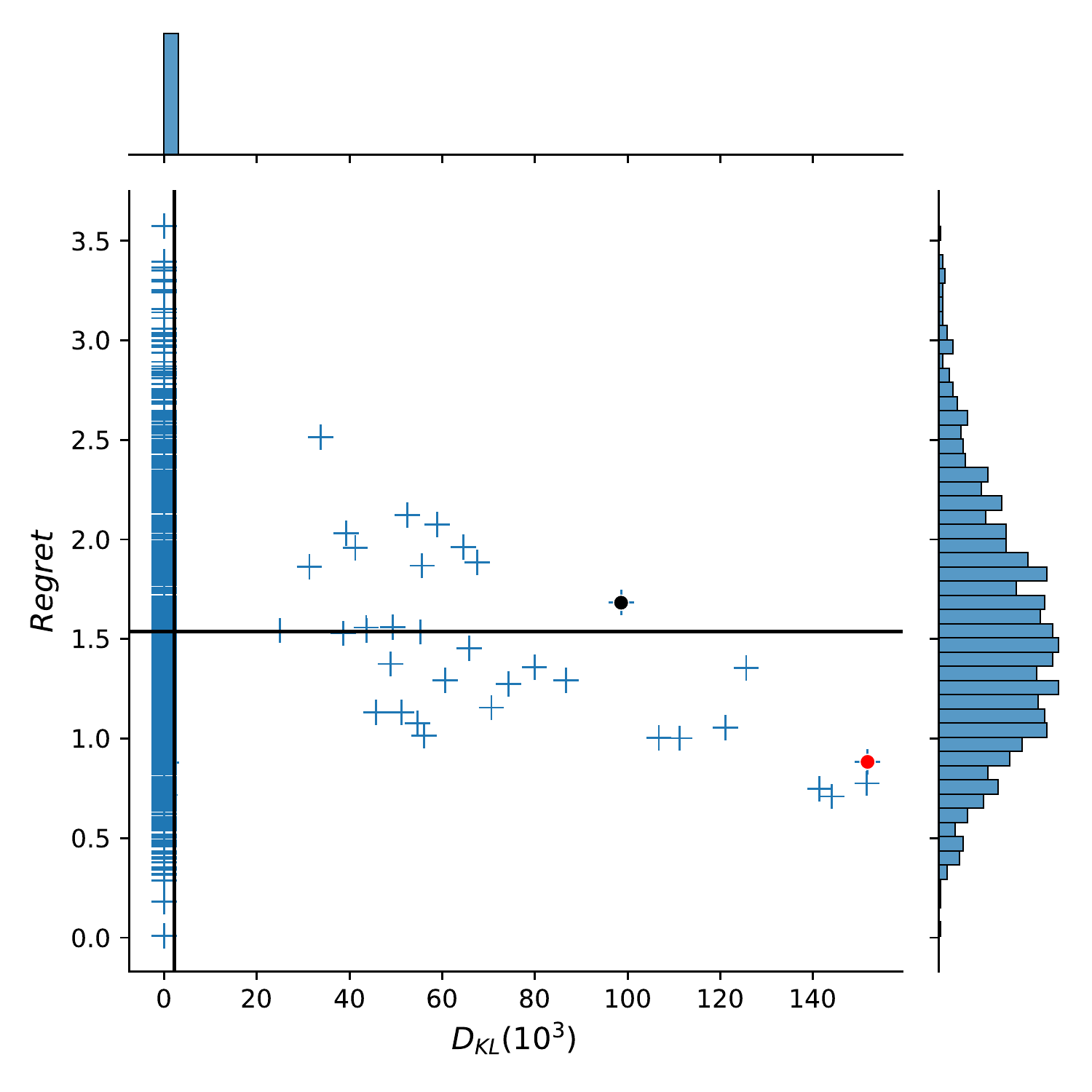}
         \caption{Sampling a random $\sigma$ from $\mathcal{N}(2, 0.8)$.}
         \label{fig:data_random}
     \end{subfigure}
     \caption{Average Regret vs $\bar{D}_{KL}$ for all movies using different methods of variance generation for a random super-user in MovieLens 1M (Crosses). The movie with highest $\bar{D}_{KL}/\Delta R^2$ is shown in red, and movies with above average regret in black. Horizontal and vertical black lines show the average regret and $\bar{D}_{KL}$, respectively.}
     \label{fig:var_example}
\end{figure}

We explore the MovieLens data set, looking for the existence of movies that would be useful for information-gathering about the state. Since the reward distributions stay fixed as they are generated through probabilistic matrix factorization outlined in section \ref{sec:movielens1m}, the source of variation in usefulness for identifying the state is due to different levels of variance in the reward. We investigated three different settings. (i) a constant and equal $\sigma = 0.25$ as used in our experiments on MovieLens, (ii) for each user, we compute the standard deviation of rewards between the three nearest neighbors of a movie given by a user according to $||V_i - V_j||^2_2$. And (iii) we draw the $\sigma$ according to a normal distribution $\sigma \sim \mathcal{N}(2, 0.8)$. Figure \ref{fig:var_example} shows the results for a random super-user. Depending on the method of variance generation, the results are significantly different for MovieLens. Choosing a fixed $\sigma  = 0.25$ results in movies exhibiting very similar $D_{KL}$. Thus, most movies only have limited use for information-gathering. Furthermore, the average $D_{KL}$ is relatively high, allowing mTS to perform well. Nevertheless, some movies have about 4-7 times higher $\bar{D}_{KL}$ than the average movie; we still see a benefit of using these movies for information-gathering, as shown in previous experiments.

When using the three nearest neighbors or sampling, the variance from the normal distribution results in a few movies with very large $D_{KL}$, promising to distinguish states very well. The same is true for the sampled variance, resulting in a few movies with very high $\bar{D}_{KL}$ between states. We ran several experiments using the sampled variance method. Furthermore, to make sampling high $D_{KL}$ arm costly, we require an information-gathering arm to have above mean regret on average. The environment samples $N$ arms randomly each time step. We focus on arms with a high reward with a mean regret below the average of the arm pool (1132 movies), focusing on providing movies with good ratings on average.

\begin{figure}
     \centering
     \begin{subfigure}[b]{0.3\textwidth}
         \centering
         \includegraphics[width=\textwidth]{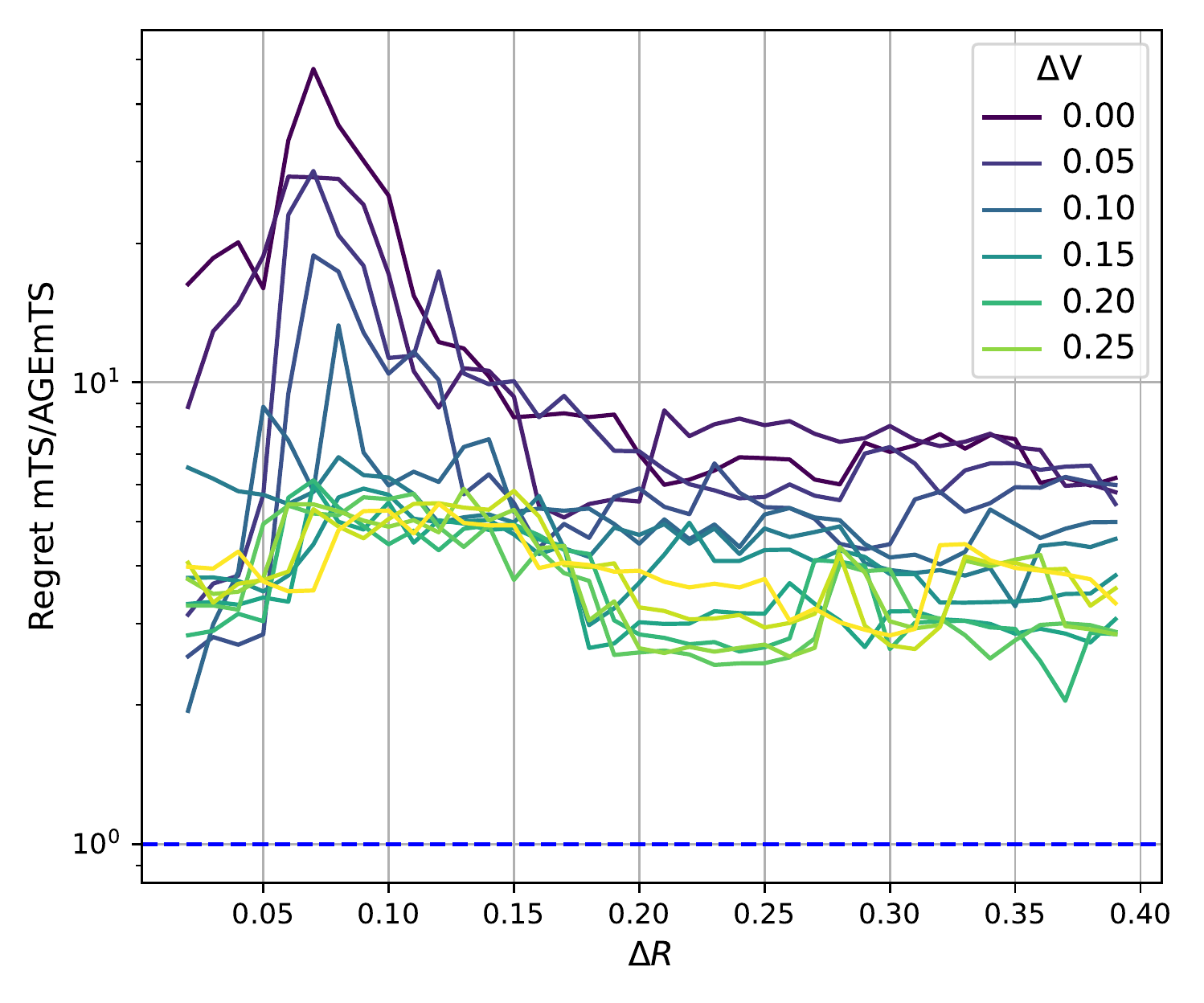}
         \caption{Synthetic experiments ($\log$-scale).}
         \label{fig:twostate_syn_reg}
     \end{subfigure}
     \begin{subfigure}[b]{0.3\textwidth}
         \centering
         \includegraphics[width=\textwidth]{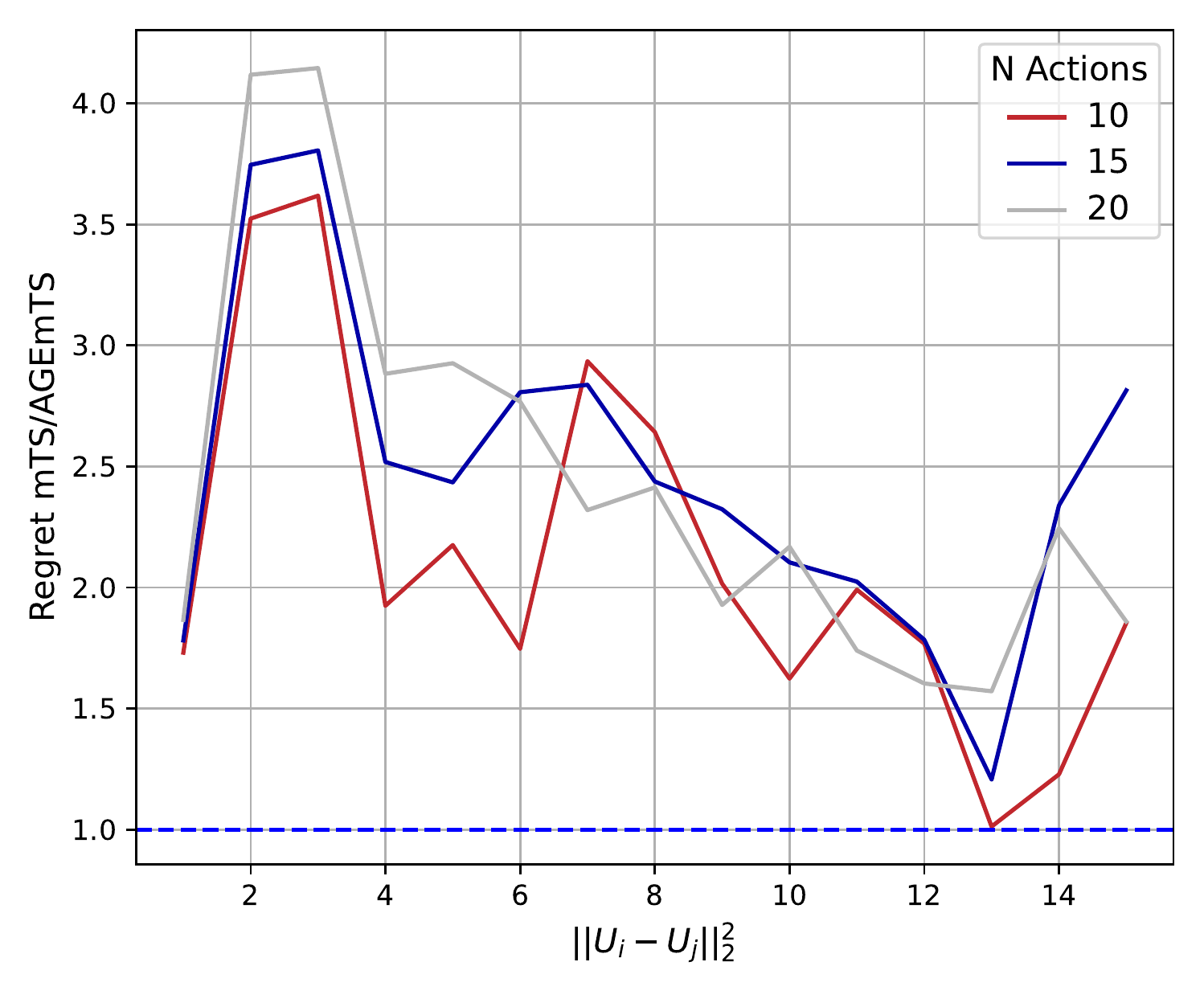}
         \caption{MovieLens 1M ($\log$-scale).}
         \label{fig:twostate_movie_reg}
     \end{subfigure}
     \begin{subfigure}[b]{0.3\textwidth}
         \centering
         \includegraphics[width=\textwidth]{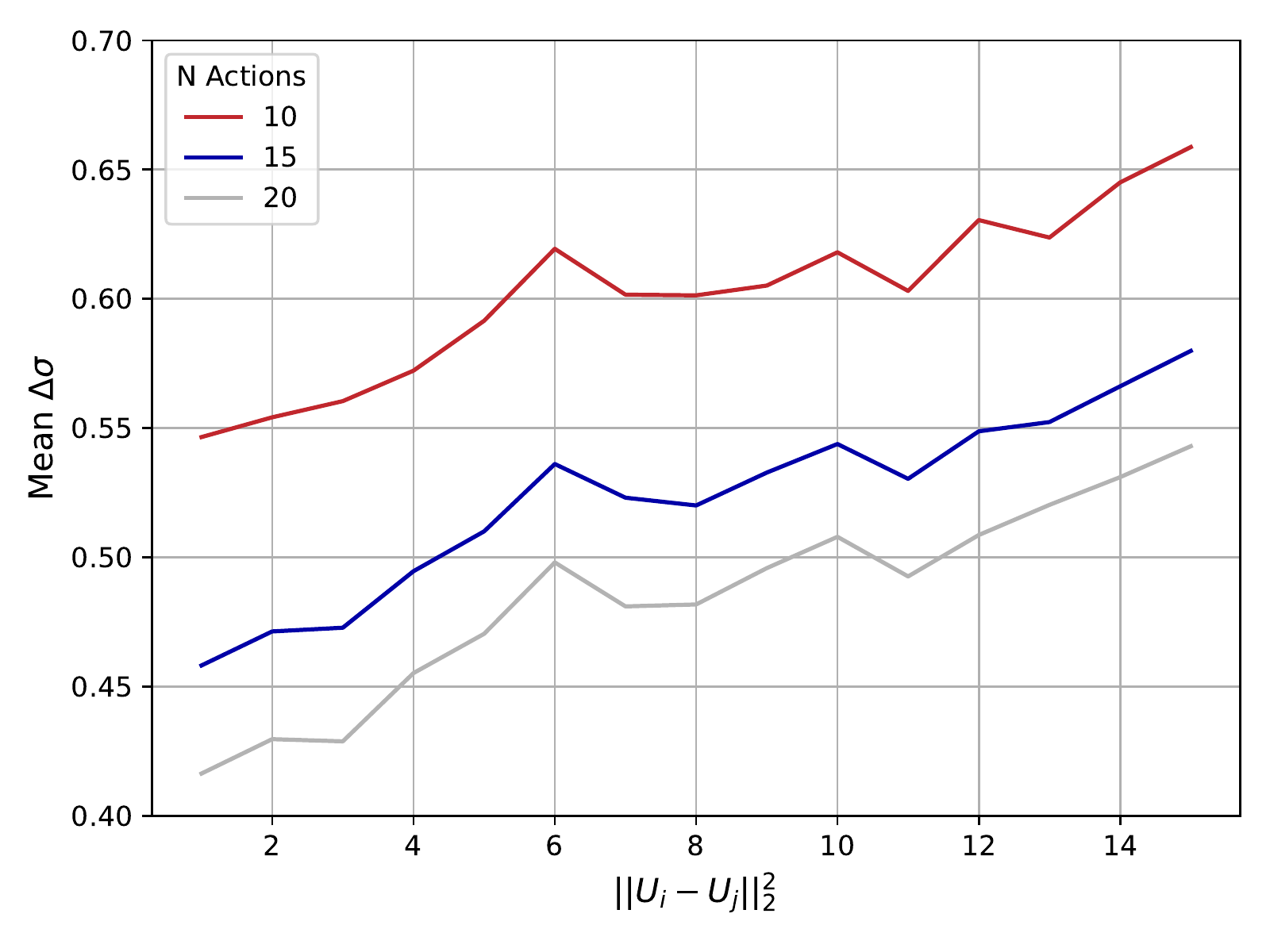}
         \caption{MovieLens 1M. Average $\Delta \sigma$ for different arm set sizes.}
         \label{fig:twostate_movie_var}
     \end{subfigure}
     \caption{Benefit of AGEmTS over mTS in the two-state stationary setting for different similarities of arm-rewards between states.}
     \label{fig:region_stationary}
\end{figure}

Figure \ref{fig:region_stationary} shows the relative regret of mTS compared to AGEmTS as a function of reward similarity of arms between states. The similarity of arm reward distributions between states depends on the difference in mean reward $\Delta R$ and standard deviation $\Delta V$. For the synthetic two-state stationary setting (figure \ref{fig:twostate_syn_reg}), the benefit of information-gathering arms is most significant for hard-to-distinguish reward distributions, close in both mean and variance of the reward. The benefit decreases when reward distributions are too close, where the regret vanishes, or are too dissimilar, becoming easy to distinguish. 

Figure \ref{fig:twostate_movie_reg}) shows the results for MovieLens 1M. We observe a similar phenomenon where the most benefit is achieved for users with similar user vectors (similar reward distributions). We ran the experiments with different sizes of arm sets. With an increasing number of arms, the mean difference between reward variances of arms reduces, thus increasing the benefit of information-gathering arms on average. We note that with an increasing difference in user vectors, the difference in rewards and mean variances increase (figure \ref{fig:twostate_movie_var}). Thus, we do not see a significant difference in regret benefit for a different number of arms anymore. This trend is similar to the synthetic experiments, where the lowest difference in regret between AGEmTS and mTS is observed for high $\Delta R$ and $\Delta \sigma$.      

\begin{figure}
     \centering
     \begin{subfigure}[b]{0.3\textwidth}
         \centering
         \includegraphics[width=\textwidth]{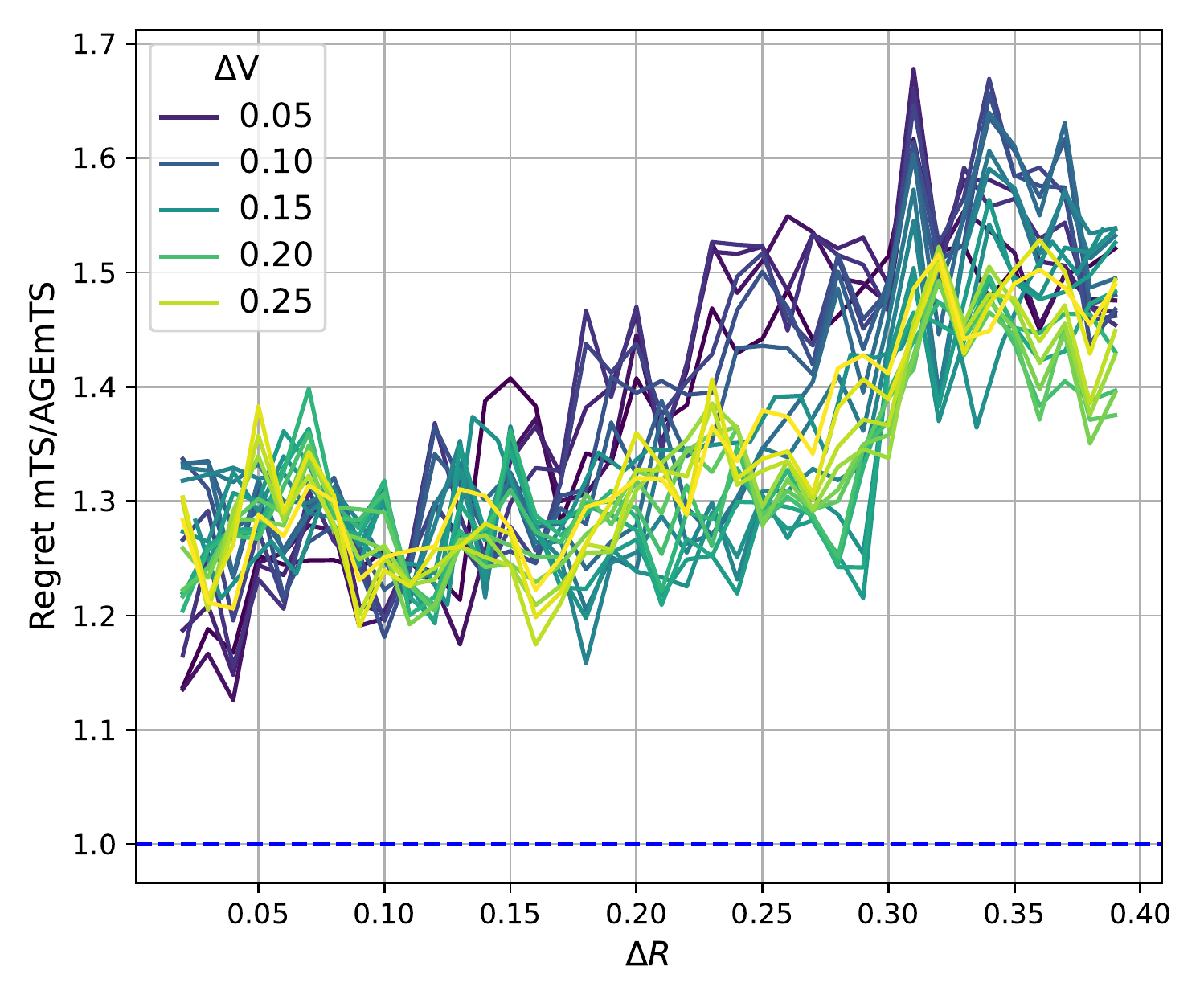}
         \caption{Synthetic experiments.}
         \label{fig:twostate_non_syn_reg}
     \end{subfigure}
     \begin{subfigure}[b]{0.3\textwidth}
         \centering
         \includegraphics[width=\textwidth]{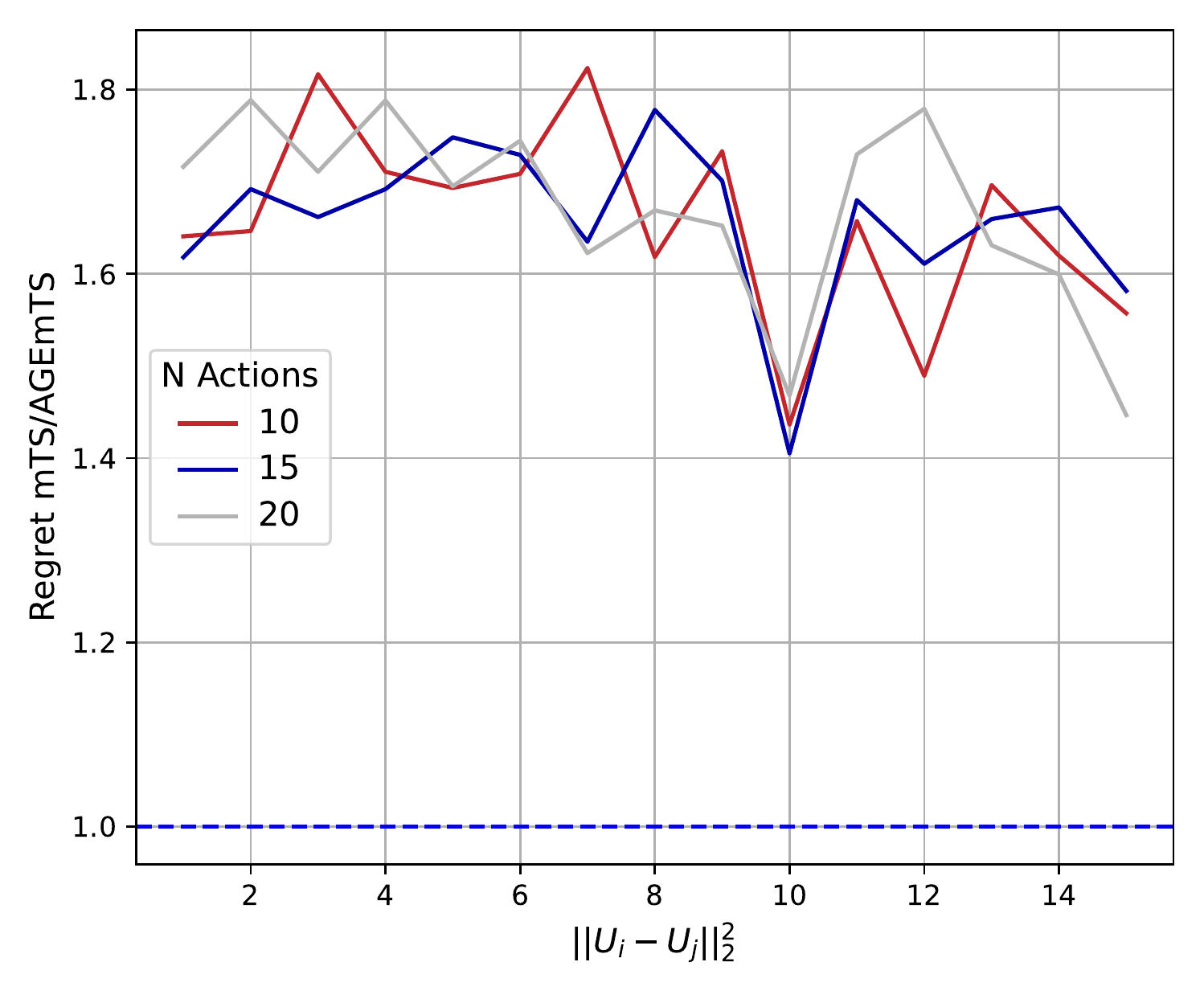}
         \caption{MovieLens 1M}
         \label{fig:twostate_non_movie_reg}
     \end{subfigure}
     \begin{subfigure}[b]{0.3\textwidth}
         \centering
         \includegraphics[width=\textwidth]{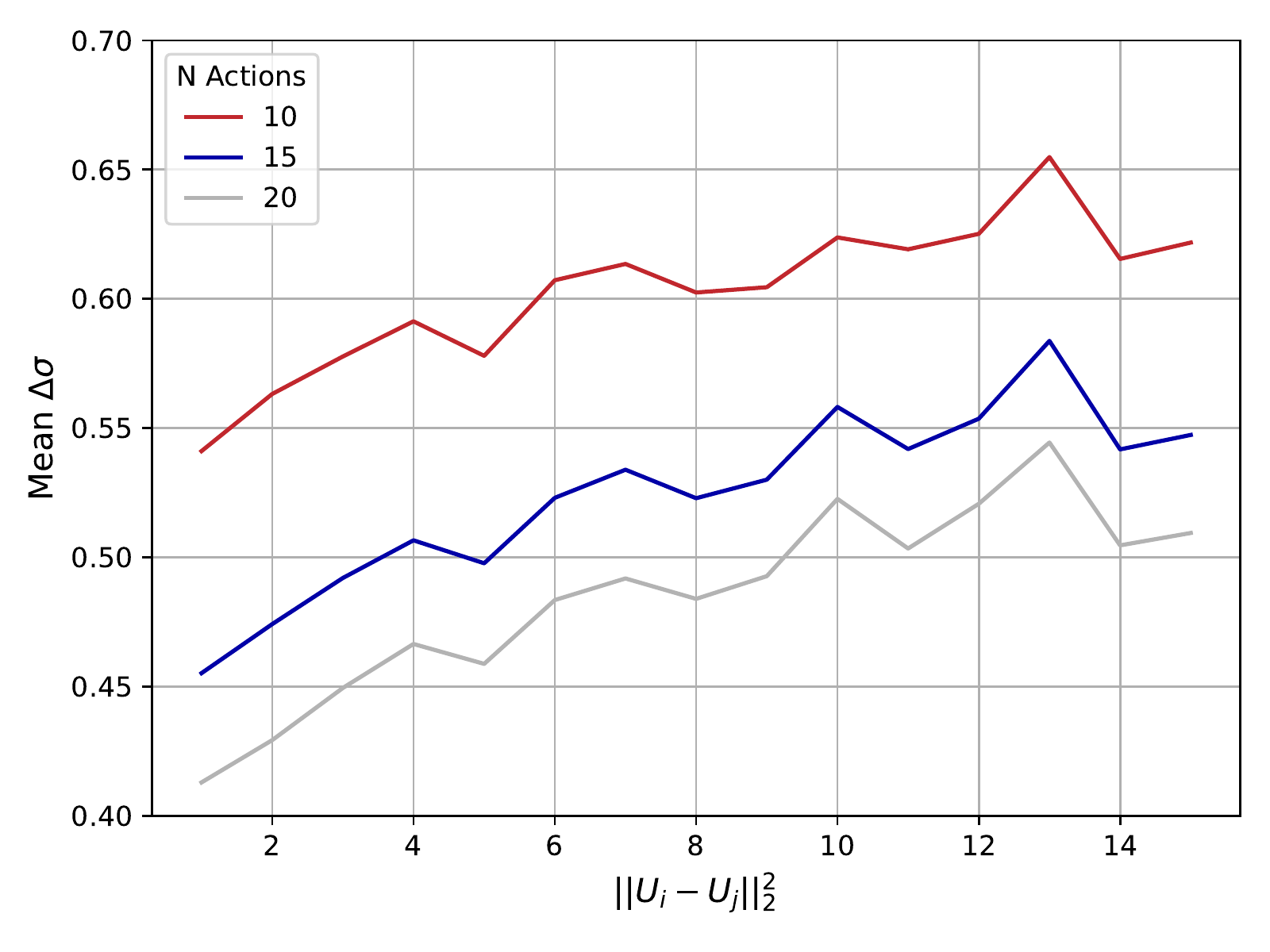}
         \caption{MovieLens 1M. Average $\Delta \sigma$ for different arm set sizes.}
         \label{fig:twostate_non_movie_var}
     \end{subfigure}
     \caption{Benefit of AGEmTS over mTS in the two-state non-stationary setting for different similarities of arm-rewards between states.}
     \label{fig:region_nonstationary}
\end{figure}

Figure \ref{fig:region_nonstationary} show the results for the non-stationary two-state setting. Here, the benefit of information-gathering arms is not as obvious as in the stationary case. For the synthetic experiments, we observe a peak for similar reward distributions, but as the differences in rewards of the best arms increase, we observe a slight reduction followed by an increase in benefit as opposed to a decrease visible in the stationary setting. While it becomes increasingly easier to distinguish between the states as $\Delta R$ increases, mistakes also become more costly. When coupled with non-stationarity, the belief-state using mTS may remain uninformative instead of converging to the true latent state at any point in time. This leads to more mistakes compared to AGEmTS, resulting in higher cumulative regret.

For MovieLens (figure \ref{fig:twostate_non_movie_reg}), the highest benefit is observed for similar user vectors and decreasing benefit with dissimilar $U$. The difference in $\Delta \sigma$ is already high for MovieLens. Thus we are in a regime where states are somewhat easy to distinguish, reducing the overall benefit of information-gathering arms. The observed trend is similar to the synthetic experiments with high $\Delta V$. We do not observe a significant difference in regret between the number of arms. Due to non-stationary and a fully-connected transition graph, the benefit of higher $\Delta \sigma$ (and conversely, information-gathering arms) is limited to the stationary intervals of the state-trajectory. These intervals are, on average, $200$ time steps long, compared to the stationary setting, where the difference in regret is measured over $1000$ time steps.

Overall, we see the most benefit of AGEmTS over mTS in both stationary and non-stationary settings for instances where states are hard to distinguish. Furthermore, as the time between state switches increases, we observe an increasing benefit of information-gathering arms, thus less cumulative regret using AGEmTS over mTS.

\section{Related Work}\label{sec:related}

\textbf{Latent Bandits}. The work that is close to ours is that of \cite{Hong2020}, \cite{Hong2020b} and \cite{Maillard14}. \cite{Maillard14} propose a UCB algorithm for the non-contextual latent bandit problem and carry out a regret analysis. They assume that the mean rewards for each state are known to the agent. \cite{Hong2020} revisit the stationary latent bandit problem developing UCB and TS type algorithms and provide an analysis. \cite{Li2016} extended the latent bandit problem to the contextual case. They consider policies learned through offline data and deployed as experts in the EXP4 algorithm.

Similarly, online policy reuse \citep{Rosman2015}, considers offline learned policies. The agent keeps a belief over the optimally of each policy and chooses accordingly. The most similar setting to ours is the work of ~\cite{Hong2020b}. They develop contextual, uncertainty-aware algorithms and provide a unified analysis of both. To our knowledge, our work is the first to introduce information-gathering in the latent bandit setting.

\textbf{Non-stationary Bandits}. Non-stationary bandits have been studied extensively in the past. The earliest works considered strategies that reweigh the influence of rewards by their recency. Strategies include discounting \citep{Kocis2006, Garivier2011, Galozy2020} or sliding window approaches \citep{Garivier2008}. In other works, the agent monitors the reward distributions and detects change points to adjust its strategy~\citep{Yu2009, Ortner2014, Auer19a, Mellor2013}. All of the methods above, in some way or another, \textit{forget} the past. Thus, they may need to relearn it, even though the agent has encountered the same situation before. Forgetting poses a major issue in environments that change in a structured manner, where the agent can exploit past information for future gain.  

\textbf{Information Directed Sampling}. Information-directed sampling (IDS) has been presented as an alternative approach to commonly used upper-confidence-bound and posterior sampling techniques to solve the exploration-exploitation trade-off in bandit problems~\citep{Russo2014}. The strategy involves minimizing the ratio between squared expected single-period regret and a measure of information gain. IDS has outperformed UCB and TS strategies in environments where knowledge about the reward gained from one arm informs the reward of other arms. A similar approach that is outperformed by IDS uses the so-called knowledge gradient, where information-gathering is guided by maximizing the marginal value of information ~\citep{Powell2010}. We differ from these prior works since, in the presents of known rewards models, we use information-gatering to identify the state of the environment instead of estimating the rewards of other arms.

\section{Conclusion and Future Work} \label{sec:conclusion}
We investigate the use of information-gathering arms (i.e., arms that offer lower immediate reward but provide long-term benefits in terms of state discrimination capability) in the latent bandit setting, where the agent's goal is to identify the current state and choose actions to maximize cumulative reward. We have developed an algorithm for this setting that shows superior performance over several state-of-the-art algorithms. We demonstrate this approach's advantages in various synthetic environments and on real-world data.

Our synthetic experiments show the importance of selecting the information-gathering arm at the right time to gain the most benefit. Since state transitions are generally stochastic, it requires a nontrivial balance between investing (regret) resources to obtain additional information that may pay off in the long run. Gathering information before the state change will provide limited gain since enough time may have passed to uncover the current state reliably. On the other hand, choosing to gather information too late after a transition has occurred will lead to 
inability to recuperate the incurred regret due to lack of time before the next transition. 
We show how the interactions between reward distribution and transition matrix influence the potential gain that can be achieved using information-gathering arms. In particular, for reducible Markov chains, significant gains in future states can be achieved by knowing the current state.

There are several avenues for future work. In this paper, we select a globally best information-gathering arm, but there is no guarantee that this arm is the best choice for all states. It would be appropriate to find arms that best resolve particular state confusion, considering potential future states. 

We primarily focus on the setting where the true reward and transition models are known. In practice, we might only have a prior over possible reward and transition models available, requiring the agent to learn new models over time. Here, information-gathering not only helps to uncover the true state but may be used to learn better models over the environment faster.

\bibliographystyle{unsrtnat}
\bibliography{references}  

\begin{thebibliography}{31}
\providecommand{\natexlab}[1]{#1}
\providecommand{\url}[1]{\texttt{#1}}
\expandafter\ifx\csname urlstyle\endcsname\relax
  \providecommand{\doi}[1]{doi: #1}\else
  \providecommand{\doi}{doi: \begingroup \urlstyle{rm}\Url}\fi

\bibitem[Villar et~al.(2015)Villar, Bowden, and Wason]{Villar2015}
S.~S. Villar, J.~Bowden, and J.~Wason.
\newblock {{M}ulti-armed {B}andit {M}odels for the {O}ptimal {D}esign of
  {C}linical {T}rials: {B}enefits and {C}hallenges}.
\newblock \emph{Stat Sci}, 30\penalty0 (2):\penalty0 199--215, 2015.
\newblock URL \url{https://doi.org/10.1214/14-STS504}.

\bibitem[Bastani and Bayati(2020)]{Bastani2020}
Hamsa Bastani and Mohsen Bayati.
\newblock Online decision making with high-dimensional covariates.
\newblock \emph{Operations Research}, 68\penalty0 (1):\penalty0 276--294, 2020.
\newblock \doi{10.1287/opre.2019.1902}.
\newblock URL \url{https://doi.org/10.1287/opre.2019.1902}.

\bibitem[Shen et~al.(2015)Shen, Wang, Jiang, and Zha]{Shen2015}
Weiwei Shen, Jun Wang, Yu-Gang Jiang, and Hongyuan Zha.
\newblock Portfolio choices with orthogonal bandit learning.
\newblock In \emph{International Conference on Artificial Intelligence},
  IJCAI'15, page 974–980. AAAI Press, 2015.
\newblock ISBN 9781577357384.

\bibitem[Huo and Fu(2017)]{Huo2017}
Xiaoguang Huo and Feng Fu.
\newblock Risk-aware multi-armed bandit problem with application to portfolio
  selection.
\newblock \emph{Royal Society Open Science}, 4\penalty0 (11):\penalty0 171377,
  2017.
\newblock \doi{10.1098/rsos.171377}.
\newblock URL
  \url{https://royalsocietypublishing.org/doi/abs/10.1098/rsos.171377}.

\bibitem[Boldrini et~al.(2018)Boldrini, De~Nardis, Caso, Le, Fiorina, and
  Di~Benedetto]{Boldrini2018}
Stefano Boldrini, Luca De~Nardis, Giuseppe Caso, Mai Le, Jocelyn Fiorina, and
  Maria-Gabriella Di~Benedetto.
\newblock mumab: A multi-armed bandit model for wireless network selection.
\newblock \emph{Algorithms}, 11\penalty0 (2):\penalty0 13, Jan 2018.
\newblock ISSN 1999-4893.
\newblock \doi{10.3390/a11020013}.
\newblock URL \url{http://dx.doi.org/10.3390/a11020013}.

\bibitem[{Kerkouche} et~al.(2018){Kerkouche}, {Alami}, {Féraud}, {Varsier},
  and {Maillé}]{Kerkouche2018}
R.~{Kerkouche}, R.~{Alami}, R.~{Féraud}, N.~{Varsier}, and P.~{Maillé}.
\newblock Node-based optimization of lora transmissions with multi-armed bandit
  algorithms.
\newblock In \emph{2018 25th International Conference on Telecommunications
  (ICT)}, pages 521--526, 2018.
\newblock \doi{10.1109/ICT.2018.8464949}.

\bibitem[Wen et~al.(2017)Wen, Kveton, Valko, and Vaswani]{Wen2017}
Zheng Wen, Branislav Kveton, Michal Valko, and Sharan Vaswani.
\newblock Online influence maximization under independent cascade model with
  semi-bandit feedback.
\newblock In I.~Guyon, U.~V. Luxburg, S.~Bengio, H.~Wallach, R.~Fergus,
  S.~Vishwanathan, and R.~Garnett, editors, \emph{Advances in Neural
  Information Processing Systems}, volume~30, pages 3022--3032. Curran
  Associates, Inc., 2017.
\newblock URL
  \url{https://proceedings.neurips.cc/paper/2017/file/7137debd45ae4d0ab9aa953017286b20-Paper.pdf}.

\bibitem[Schwartz et~al.(2017)Schwartz, Bradlow, and Fader]{Schwartz2017}
Eric Schwartz, Eric Bradlow, and Peter Fader.
\newblock Customer acquisition via display advertising using multi-armed bandit
  experiments.
\newblock \emph{Marketing Science}, 36, 04 2017.
\newblock \doi{10.1287/mksc.2016.1023}.

\bibitem[{Wang} et~al.(2019){Wang}, {Zeng}, {Zhou}, {Li}, {Iyengar}, {Shwartz},
  and {Grabarnik}]{Wang2019}
Q.~{Wang}, C.~{Zeng}, W.~{Zhou}, T.~{Li}, S.~S. {Iyengar}, L.~{Shwartz}, and
  G.~Y. {Grabarnik}.
\newblock Online interactive collaborative filtering using multi-armed bandit
  with dependent arms.
\newblock \emph{IEEE Transactions on Knowledge and Data Engineering},
  31\penalty0 (8):\penalty0 1569--1580, 2019.
\newblock \doi{10.1109/TKDE.2018.2866041}.

\bibitem[Baltrunas et~al.(2015)Baltrunas, Church, Karatzoglou, and
  Oliver]{Baltrunas2015}
Linas Baltrunas, Karen Church, Alexandros Karatzoglou, and Nuria Oliver.
\newblock Frappe: Understanding the usage and perception of mobile app
  recommendations in-the-wild.
\newblock \emph{CoRR}, abs/1505.03014, 2015.
\newblock URL \url{http://arxiv.org/abs/1505.03014}.

\bibitem[Hong et~al.(2020{\natexlab{a}})Hong, Kveton, Zaheer, Chow, Ahmed,
  Ghavamzadeh, and Boutilier]{Hong2020}
Joey Hong, Branislav Kveton, Manzil Zaheer, Yinlam Chow, Amr Ahmed, Mohammad
  Ghavamzadeh, and Craig Boutilier.
\newblock Non-stationary latent bandits.
\newblock \emph{CoRR}, abs/2012.00386, 2020{\natexlab{a}}.
\newblock URL \url{https://arxiv.org/abs/2012.00386}.

\bibitem[Maillard and Mannor(2014)]{Maillard14}
Odalric-Ambrym Maillard and Shie Mannor.
\newblock Latent bandits.
\newblock In Eric~P. Xing and Tony Jebara, editors, \emph{Proceedings of the
  31st International Conference on Machine Learning}, volume~32 of
  \emph{Proceedings of Machine Learning Research}, pages 136--144, Bejing,
  China, 22--24 Jun 2014. PMLR.
\newblock URL \url{https://proceedings.mlr.press/v32/maillard14.html}.

\bibitem[Russo and Van~Roy(2014)]{Russo2014}
Daniel Russo and Benjamin Van~Roy.
\newblock Learning to optimize via information-directed sampling.
\newblock In Z.~Ghahramani, M.~Welling, C.~Cortes, N.~Lawrence, and K.Q.
  Weinberger, editors, \emph{Advances in Neural Information Processing
  Systems}, volume~27. Curran Associates, Inc., 2014.
\newblock URL
  \url{https://proceedings.neurips.cc/paper/2014/file/301ad0e3bd5cb1627a2044908a42fdc2-Paper.pdf}.

\bibitem[Auer(2003)]{Auer2003b}
Peter Auer.
\newblock Using confidence bounds for exploitation-exploration trade-offs.
\newblock \emph{J. Mach. Learn. Res.}, 3:\penalty0 397–422, March 2003.
\newblock ISSN 1532-4435.
\newblock URL \url{https://dl.acm.org/doi/10.5555/944919.944941}.

\bibitem[Agrawal and Goyal(2013)]{agrawal2014thompson}
Shipra Agrawal and Navin Goyal.
\newblock Thompson sampling for contextual bandits with linear payoffs.
\newblock volume~28 of \emph{Machine Learning Research}, pages 127--135,
  Atlanta, Georgia, USA, 17--19 Jun 2013. PMLR.
\newblock URL \url{http://proceedings.mlr.press/v28/agrawal13.html}.

\bibitem[Abeille and Lazaric(2017)]{Abeille2017}
Marc Abeille and Alessandro Lazaric.
\newblock {Linear Thompson Sampling Revisited}.
\newblock In Aarti Singh and Jerry Zhu, editors, \emph{Proceedings of the 20th
  International Conference on Artificial Intelligence and Statistics},
  volume~54 of \emph{Proceedings of Machine Learning Research}, pages 176--184.
  PMLR, 20--22 Apr 2017.
\newblock URL \url{https://proceedings.mlr.press/v54/abeille17a.html}.

\bibitem[Hong et~al.(2020{\natexlab{b}})Hong, Kveton, Zaheer, Chow, Ahmed, and
  Boutilier]{Hong2020b}
Joey Hong, Branislav Kveton, Manzil Zaheer, Yinlam Chow, Amr Ahmed, and Craig
  Boutilier.
\newblock Latent bandits revisited.
\newblock In H.~Larochelle, M.~Ranzato, R.~Hadsell, M.F. Balcan, and H.~Lin,
  editors, \emph{Advances in Neural Information Processing Systems}, volume~33,
  pages 13423--13433. Curran Associates, Inc., 2020{\natexlab{b}}.
\newblock URL
  \url{https://proceedings.neurips.cc/paper/2020/file/9b7c8d13e4b2f08895fb7bcead930b46-Paper.pdf}.

\bibitem[Auer et~al.(2003)Auer, Cesa-Bianchi, Freund, and Schapire]{Auer2003}
Peter Auer, Nicol\`{o} Cesa-Bianchi, Yoav Freund, and Robert~E. Schapire.
\newblock The nonstochastic multiarmed bandit problem.
\newblock \emph{SIAM J. Comput.}, 32\penalty0 (1):\penalty0 48–77, January
  2003.
\newblock ISSN 0097-5397.
\newblock \doi{10.1137/S0097539701398375}.
\newblock URL \url{https://doi.org/10.1137/S0097539701398375}.

\bibitem[Salakhutdinov and Mnih(2007)]{Salakhutdinov2008}
Ruslan Salakhutdinov and Andriy Mnih.
\newblock Probabilistic matrix factorization.
\newblock In \emph{Proceedings of the 20th International Conference on Neural
  Information Processing Systems}, NIPS'07, page 1257–1264, Red Hook, NY,
  USA, 2007. Curran Associates Inc.
\newblock ISBN 9781605603520.

\bibitem[Wu et~al.(2018)Wu, Iyer, and Wang]{Wu2018}
Qingyun Wu, Naveen Iyer, and Hongning Wang.
\newblock Learning contextual bandits in a non-stationary environment.
\newblock In \emph{The 41st International ACM SIGIR Conference on Research \&
  Development in Information Retrieval}, SIGIR '18, page 495–504, New York,
  NY, USA, 2018. Association for Computing Machinery.
\newblock ISBN 9781450356572.
\newblock \doi{10.1145/3209978.3210051}.
\newblock URL \url{https://doi.org/10.1145/3209978.3210051}.

\bibitem[Zhou and Brunskill(2016)]{Li2016}
Li~Zhou and Emma Brunskill.
\newblock Latent contextual bandits and their application to personalized
  recommendations for new users.
\newblock In \emph{Proceedings of the Twenty-Fifth International Joint
  Conference on Artificial Intelligence}, IJCAI'16, page 3646–3653. AAAI
  Press, 2016.
\newblock ISBN 9781577357704.

\bibitem[Rosman et~al.(2016)Rosman, Hawasly, and Ramamoorthy]{Rosman2015}
Benjamin Rosman, Majd Hawasly, and Subramanian Ramamoorthy.
\newblock Bayesian policy reuse, 2016.
\newblock URL \url{https://doi.org/10.1007/s10994-016-5547-y}.

\bibitem[Kocsis and Szepesvári(2006)]{Kocis2006}
Levente Kocsis and Csaba Szepesvári.
\newblock Discounted ucb.
\newblock \emph{2nd PASCAL Challenges Workshop}, 2006.
\newblock URL \url{https://www.lri.fr/~sebag/Slides/Venice/Kocsis.pdf}.

\bibitem[Garivier and Moulines(2011)]{Garivier2011}
Aur{\'e}lien Garivier and Eric Moulines.
\newblock On upper-confidence bound policies for switching bandit problems.
\newblock In Jyrki Kivinen, Csaba Szepesv{\'a}ri, Esko Ukkonen, and Thomas
  Zeugmann, editors, \emph{Algorithmic Learning Theory}, pages 174--188,
  Berlin, Heidelberg, 2011. Springer Berlin Heidelberg.
\newblock ISBN 978-3-642-24412-4.

\bibitem[Galozy et~al.(2020)Galozy, Nowaczyk, and Ohlsson]{Galozy2020}
Alexander Galozy, Slawomir Nowaczyk, and Mattias Ohlsson.
\newblock A new bandit setting balancing information from state evolution and
  corrupted context, 2020.
\newblock URL \url{https://arxiv.org/abs/2011.07989}.

\bibitem[Garivier and Moulines()]{Garivier2008}
Aurélien Garivier and Eric Moulines.
\newblock On upper-confidence bound policies for non-stationary bandit
  problems.
\newblock URL \url{https://arxiv.org/abs/0805.3415}.

\bibitem[Yu and Mannor(2009)]{Yu2009}
Jia~Yuan Yu and Shie Mannor.
\newblock Piecewise-stationary bandit problems with side observations.
\newblock In \emph{Proceedings of the 26th Annual International Conference on
  Machine Learning}, ICML '09, page 1177–1184, New York, NY, USA, 2009.
  Association for Computing Machinery.
\newblock ISBN 9781605585161.
\newblock \doi{10.1145/1553374.1553524}.
\newblock URL \url{https://doi.org/10.1145/1553374.1553524}.

\bibitem[Ortner et~al.(2014)Ortner, Ryabko, Auer, and Munos]{Ortner2014}
Ronald Ortner, Daniil Ryabko, Peter Auer, and Rémi Munos.
\newblock Regret bounds for restless markov bandits.
\newblock \emph{Theoretical Computer Science}, 558:\penalty0 62--76, 2014.
\newblock ISSN 0304-3975.
\newblock \doi{10.1016/j.tcs.2014.09.026}.
\newblock URL
  \url{https://www.sciencedirect.com/science/article/pii/S030439751400704X}.
\newblock Algorithmic Learning Theory.

\bibitem[Auer et~al.(2019)Auer, Gajane, and Ortner]{Auer19a}
Peter Auer, Pratik Gajane, and Ronald Ortner.
\newblock Adaptively tracking the best bandit arm with an unknown number of
  distribution changes.
\newblock In Alina Beygelzimer and Daniel Hsu, editors, \emph{Proceedings of
  the Thirty-Second Conference on Learning Theory}, volume~99 of
  \emph{Proceedings of Machine Learning Research}, pages 138--158, Phoenix,
  USA, 25--28 Jun 2019. PMLR.
\newblock URL \url{http://proceedings.mlr.press/v99/auer19a.html}.

\bibitem[Mellor and Shapiro(2013)]{Mellor2013}
Joseph Mellor and Jonathan Shapiro.
\newblock Thompson sampling in switching environments with bayesian online
  change detection.
\newblock In Carlos~M. Carvalho and Pradeep Ravikumar, editors,
  \emph{Proceedings of the Sixteenth International Conference on Artificial
  Intelligence and Statistics}, volume~31 of \emph{Proceedings of Machine
  Learning Research}, pages 442--450, Scottsdale, Arizona, USA, 29 Apr--01 May
  2013. PMLR.
\newblock URL \url{https://proceedings.mlr.press/v31/mellor13a.html}.

\bibitem[Powell(2011)]{Powell2010}
Warren~B. Powell.
\newblock \emph{The Knowledge Gradient for Optimal Learning}.
\newblock John Wiley \& Sons, Ltd, 2011.
\newblock ISBN 9780470400531.
\newblock \doi{https://doi.org/10.1002/9780470400531.eorms0444}.
\newblock URL
  \url{https://onlinelibrary.wiley.com/doi/abs/10.1002/9780470400531.eorms0444}.

\end{thebibliography}






\end{document}